\documentclass[onecolumn,10pt]{asme2ej}

\usepackage{graphicx} 
\usepackage{hyperref}   
\hypersetup{
	colorlinks=true,
	linkcolor=blue,
	citecolor=blue,
	urlcolor=blue,
}
\usepackage[square,numbers]{natbib}
\usepackage{url}
\usepackage{hyperref}
\usepackage{amsmath}
\usepackage{amssymb}
\usepackage{tikz}
\usetikzlibrary{shapes,arrows,backgrounds,shadows,petri,calc}

\title{Robustness investigation of cross-validation based 
quality measures for model assessment
}

\author{Thomas Most
    \affiliation{
	Ansys Germany GmbH\\
	Weimar, Germany
    }
		\vspace{2mm}
    \affiliation{ Bauhaus-Universität Weimar\\
		Weimar, Germany\\
    }
}
\author{Lars Gräning
    \affiliation{ Ansys Germany GmbH\\
	Weimar, Germany\\
    }
}
\author{Sebastian Wolff
        \affiliation{Ansys Austria GmbH\\
	Vienna, Austria\\
     }
}

\begin{document}
\maketitle    

\begin{abstract}
{\it In this paper the accuracy and robustness of quality measures for the assessment of machine learning models are investigated. The prediction quality of a machine learning model is evaluated model-independent based on a cross-validation approach, where the approximation error is estimated for unknown data. The presented measures quantify the amount of explained variation in the model prediction. The reliability of these measures is assessed by means of several numerical examples, where an additional data set for the verification of the estimated prediction error is available. Furthermore, the confidence bounds of the presented quality measures are estimated and local quality measures are derived from the prediction residuals obtained by the cross-validation approach.
}
\end{abstract}

\section{Introduction}
Nowadays, the application of mathematical surrogate models plays an important role in engineering design.
Starting with classical Design of Experiment schemes and classical polynomial response surface models 
\cite{Myers2002}, \cite{Montgomery2003},
meanwhile a wide range of surrogate models has been developed such as 
Kriging \cite{Krige1951}, Moving Least Squares \cite{Lancaster1981}, Radial Basis Functions \cite{Park1993} and Support Vector Machines \cite{Smola2004}.
Recently, artificial neural networks \cite{Hagan1996} have been extended to more sophisticated Deep Learning models \cite{goodfellow2016deep} which can be applied on a very wide range of engineering fields \cite{Herrmann2024}.
A good overview of current applications of surrogate models in global optimization is given in \cite{Ye2019} and recent developments in surrogate-assisted global sensitivity analysis can be found in \cite{Cheng2020}.
Investigations on the accuracy of machine learning models for uncertainty quantification are published in \cite{Bucher_2008_PEM}, \cite{Moustapha2022}.
Further reviews on engineering applications are available in \cite{Yondo2019}, \cite{Westermann2019} and \cite{Zhang2023}.

Generally, the application of surrogate models will introduce an additional model error in the prediction. 
Dependent on the application, the assessment of the approximation quality and the verification of the surrogate model with unknown data is very important as discussed in \cite{Queipo2005}, \cite{Forrester2008}
the assessment of the prediction quality for unknown data is necessary. A quite common approach for this purpose is the well-known cross validation \cite{Browne_2000}.
Further methods on model assessment are discussed in \cite{Kleijnen_2000},\cite{Molinaro_2005},\cite{Bischl_2012} where mainly re-sampling methods are considered.
A different approach is Bayesian model assessment \cite{Beck_2004},\cite{Park_2010},\cite{Most_2011_CompStruct} where the model evidence due to the model parameter uncertainty is evaluated.

In our study we consider quality measures based on cross-validation due to the straight-forward implementation and clear interpretation of the results as discussed recently in
\cite{Cheng2020},\cite{Bucher_2018_MOQ},\cite{Niehoff2024},\cite{Escribano_2024_pymetamodels}.
Based on the cross-validation procedure the approximation errors of unknown data points can be estimated. In \cite{Most_2011_WOST} a variance-based quality measure, the Coefficient of Prognosis (CoP) was introduced based on this principle. With help of this measure a model independent assessment and selection is possible which was realized in the Metamodel of Optimal Prognosis (MOP) in \cite{Most_2011_WOST} and extended for deep-learning models in \cite{Most_2022_NAFEMS_DACH1}.

In this paper, the robustness and stability of these quality measures by using different cross validation procedures are investigated. Based on the prediction residuals, the confidence bounds of the CoP are estimated and verified by means of several numerical examples. Additional to the global quality measures, a local model independent error estimator is introduced, which can be utilized for local model improvement by additional samples.
Finally, we recommend an extension of the CoP for non-scalar outputs, which is investigated by a further example.

\section{Quality measures for the model assessment}
\subsection{Measuring the goodness of fit}
Let us assume a simulation model with a certain number of scalar outputs. Each of these outputs can be represented as a black-box function of a given number of inputs
\begin{equation}
y(\mathbf{x})=f(x_1,x_2,\ldots,x_m).
\end{equation}
If these output functions are approximated by a mathematical surrogate model, we obtain an approximation of the true function
\begin{equation}
\hat y(\mathbf{x})=\hat f(x_1,x_2,\ldots,x_m).
\end{equation}
If the approximation model is build or trained based on a given number of support points $n$, we can calculate the residuals for each of the support points and estimate different error measures to quantify the goodness of fit
\begin{equation}
\epsilon_i = y(\mathbf{x}_i)- \hat y(\mathbf{x}_i) = y_i - {\hat y}_i.
\label{res_cod}
\end{equation}
One well known measure is the root mean square error (RMSE) 
\begin{equation}
RMSE = \displaystyle \sqrt{\frac{1}{n}\sum_{i=1}^n ( y_i - {\hat y}_i)^2},
\end{equation}
which has the same unit as the output itself and can be interpreted as the standard deviation of the approximation error. Another well-known measure is the unitless Coefficient of Determination (CoD), which measures the ratio of the explained vs. the original variation of the investigated response. In \cite{Kvalseth1985} different formulations for CoD are discussed. The two most common formulations are
\begin{equation}
CoD_1=1-\frac{SS_E}{SS_T},\quad CoD_2=\frac{SS_R}{SS_T},
\label{CoD_eq1}
\end{equation}
where the sum of squared errors $SS_E$ quantifies the unexplained variation, the  explained sum of squares $SS_R$ quantifies the explained variation and the sum of total squares $SS_T$ is equivalent to the total variation of the response
\begin{equation}
SS_E=\sum_{i=1}^n(y_i-{\hat y}_i)^2, \quad SS_R=\sum_{i=1}^n({\hat y}_i-\mu_{Y})^2, \quad SS_T=\sum_{i=1}^n(y_i-\mu_{Y})^2, \quad \mu_{Y}=\frac{1}{n}\sum_{i=1}^n y_i.
\end{equation}
Only for a linear least-squares model, the two formulations in equation \ref{CoD_eq1} agree and the following equation is valid 
\begin{equation}
SS_T = SS_E + SS_R,\quad 0 \leq CoD_{1,2} \leq 1.
\end{equation}
The application of the CoD for non-linear models is possible but requires special attention as discussed in \cite{Kvalseth1985}.

\begin{figure}[th]
\includegraphics[width=0.49\textwidth]{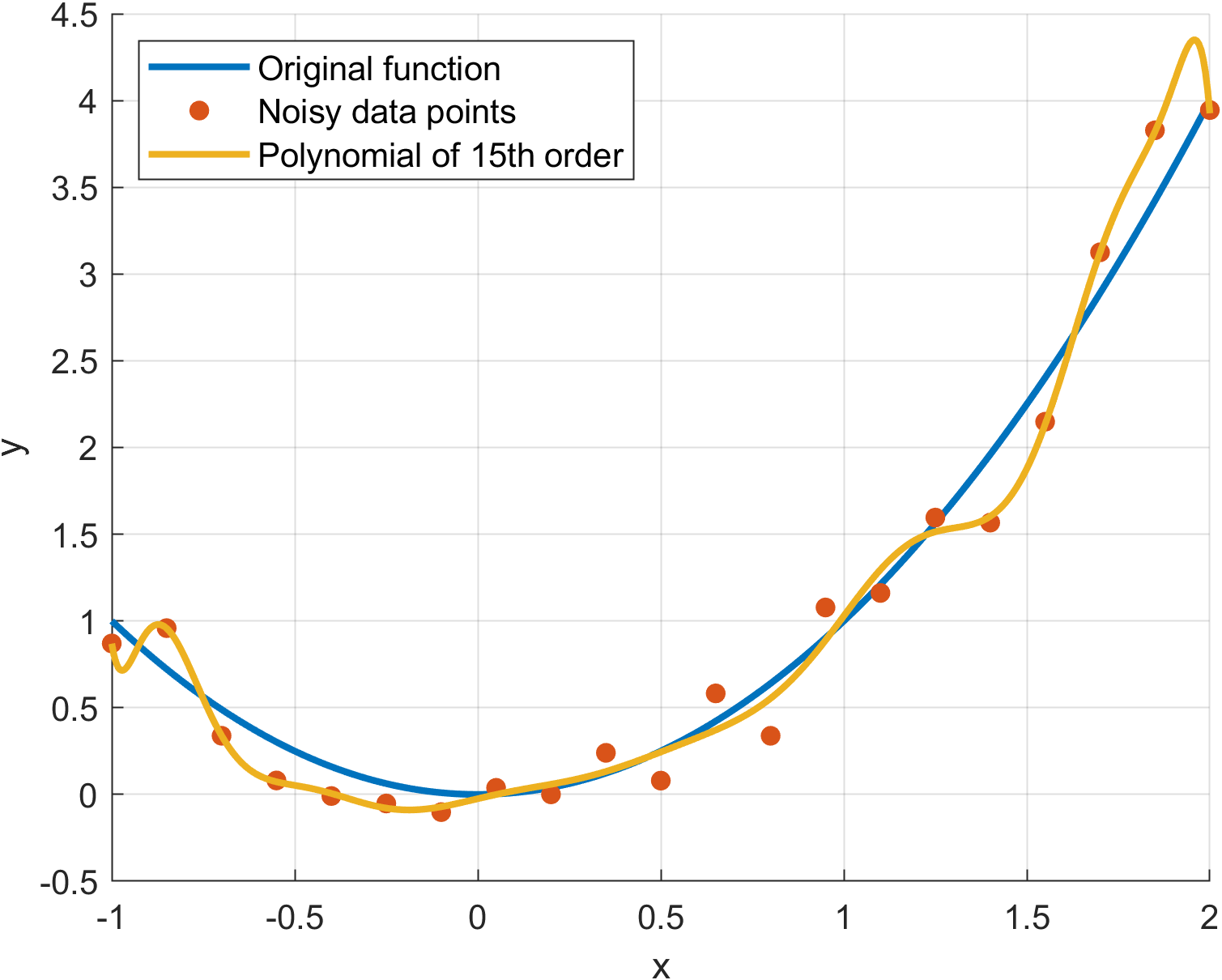}
\hfill
\includegraphics[width=0.49\textwidth]{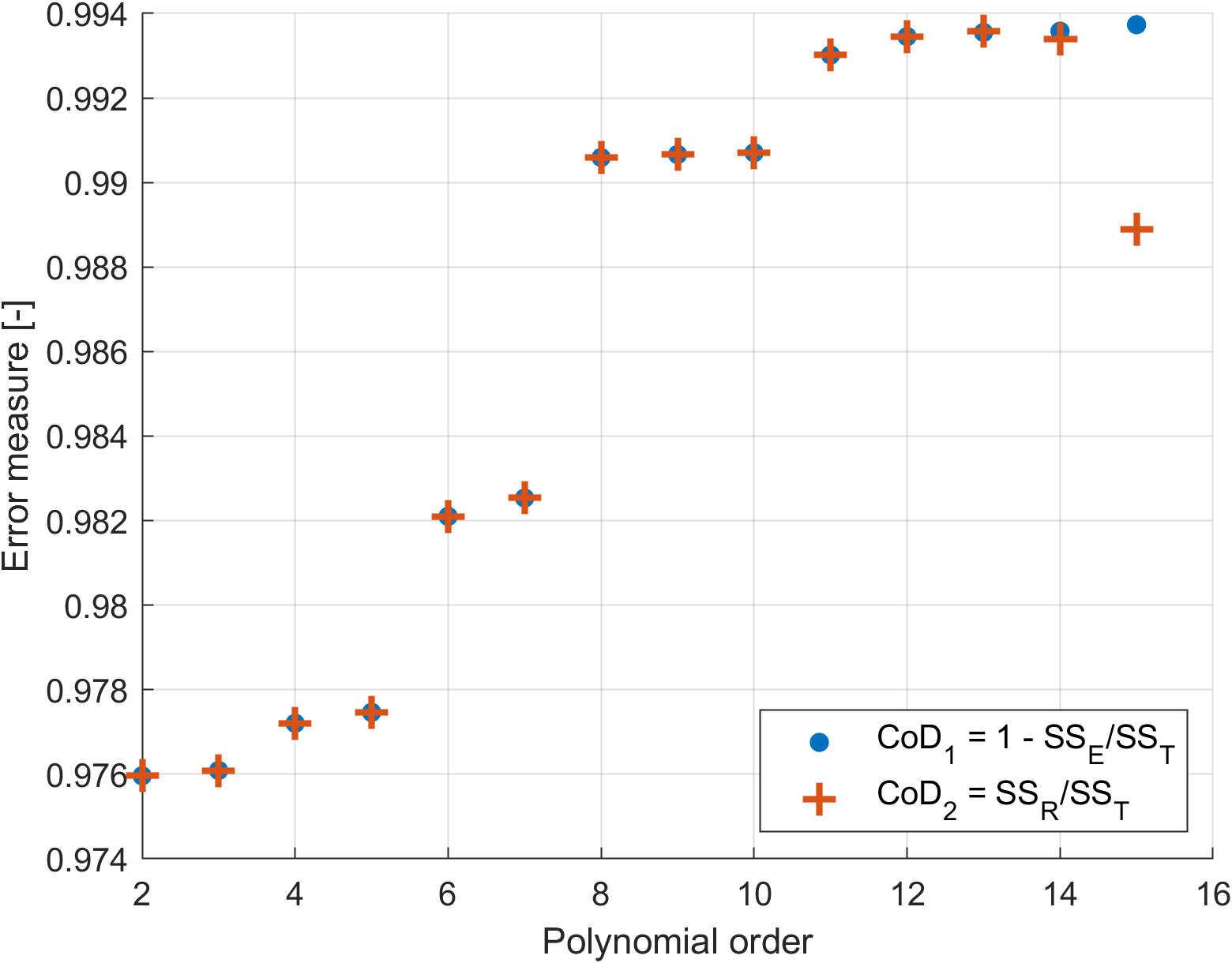}
\caption{Approximation of noisy data points of a one-dimensional quadratic function with a polynomial model with increasing order}
\label{cod_fig}
\end{figure}
Figure~\ref{cod_fig} shows an illustrative example, where a quadratic function is approximated with a linear regression model with increasing polynomial order. In this example the synthetic data points contain a small amount of random noise. The figure indicates, that a high-order polynomial will tend to fit through the noisy data points  and the corresponding $SS_E$ decreases. With increasing approximation order, the non-linearity of the polynomial model and thus the difference between the two formulations of the CoD will increase. The explained sum of squares $SS_R$ could exceed the total sum of squares $SS_T$ and the second formulation of the CoD could lead to values larger than one.

The first formulation of the CoD could be directly formulated in terms of the squared RMSE as follows
\begin{equation}
CoD=1-\frac{SS_E}{SS_T}= 1 - \frac{n^2}{SST} \cdot RMSE^2.
\label{CoD_eq2}
\end{equation}
This formulation is always smaller or equal one and could be interpreted as the scaled error variance of the approximation model. Only if the sum of squared errors $SS_E$ is larger as the total sum of squares $SS_T$,
the formulation in equation \ref{CoD_eq2} could be negative. This is not the case for the most approximation models mentioned in the introduction as long a constant baseline  is included in the model, which is the case in linear regression \cite{Montgomery2003}, Moving Least Squares \cite{Lancaster1981} and Ordinary Kriging \cite{Forrester2008}.

Since the formulation in equation~\ref{CoD_eq2} is directly related to the RMSE, we use this measure in the following paper equivalently to the RSME in order to quantify the deviation between the support point values used for the training and the approximation values at these points. Unfortunately, this will not give us any information of the prediction quality of the surrogate model for unknown data points. Therefore, we extend this measure in the following section.

\subsection{Measuring the prognosis quality}
\begin{figure}[th]
\center
	\includegraphics[width=0.48\textwidth]{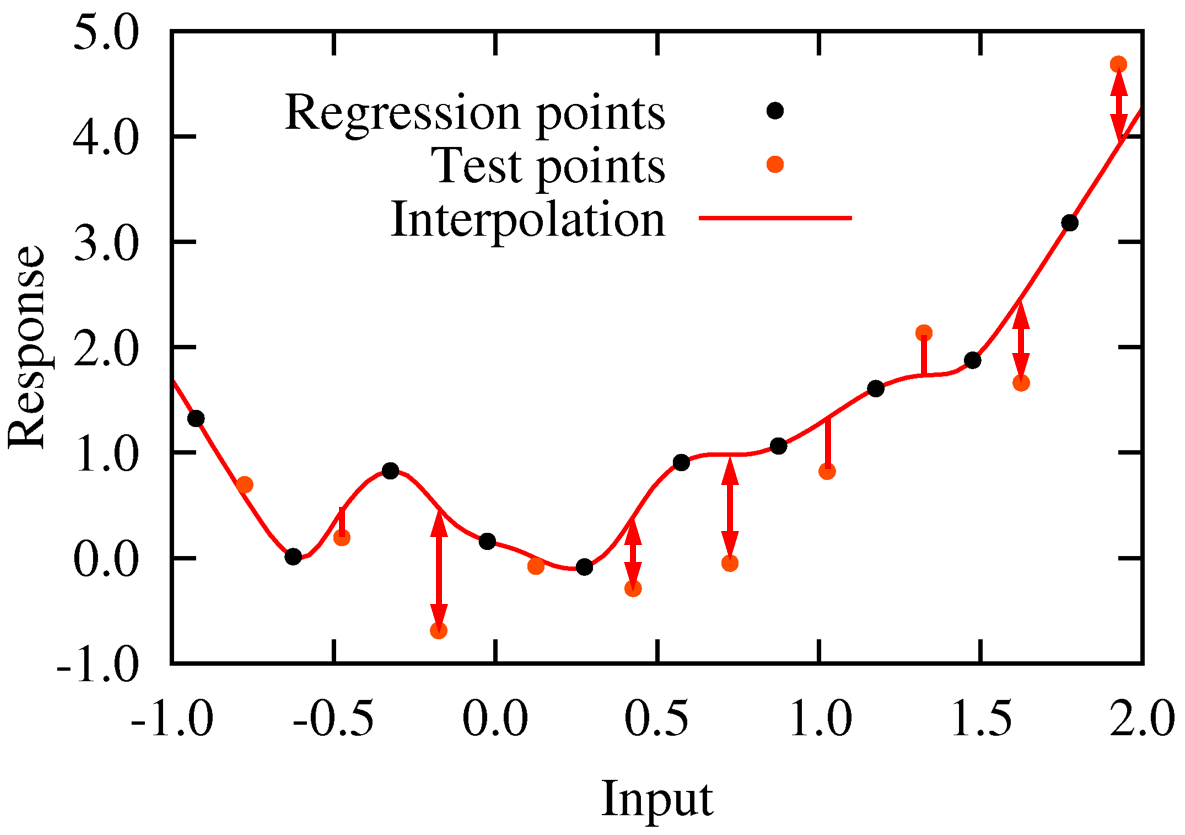}
	\hfill
	\includegraphics[width=0.48\textwidth]{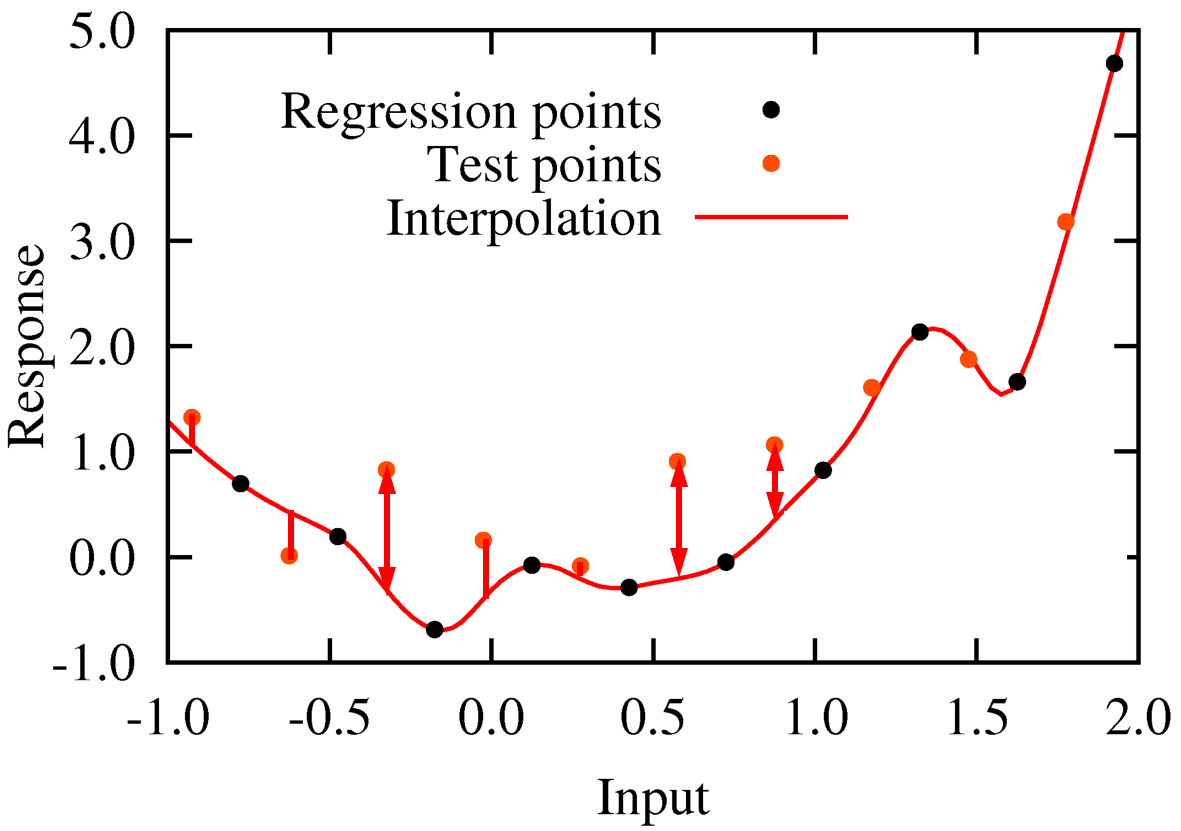}
\caption{Basic cross-validation procedure by splitting the data set in two subsets: Using set one for training and set two for prediction (left) and set two for training and set one for prediction (right)}
\label{cv_fig}
\end{figure}
In order to estimate the prediction error of a mathematical surrogate model, we can split the data set in two data sets of same size and use set number one for the training and set number two for the estimation of the prediction errors. In a second step this procedure is applied by using data set two for the training and data set one for the estimation. This procedure as shown in figure \ref{cv_fig} is called cross-validation and is explained in more detail in \cite{Forrester2008}. More generally, we can subdivide the original data set in $q$ subsets of almost equal size, where the points in each subset should be selected in that way that they cover the investigated space of the input variables almost uniformly. Thus, each of the $n$ support points are mapped to one subset
\begin{equation}
\zeta:\{1,\ldots,n\} \rightarrow \{1,\ldots,q\}.
\end{equation}
Once, the $q$ individual cross validation models have been trained, we use the approximation values to evaluate the prediction residuals for each of the available data points
\begin{equation}
{\hat y}^{cv}(\mathbf{x}_i)={\hat f}_{\sim \zeta(i)}(\mathbf{x}_i),
\end{equation}
where ${\hat f}_{\sim \zeta(i)}(.)$ is the approximation model built by using all cross validation subsets except the one set belonging to the support point $i$.
Usually, 5-10 subsets are used within the cross validation procedure to obtain stable estimators \cite{Forrester2008}. This procedure is called k-fold cross validation. 
From this prediction the corresponding residuals of the cross-validation prediction errors can be estimated as
\begin{equation}
\epsilon_i^{cv} = y(\mathbf{x}_i)- {\hat y}^{cv}(\mathbf{x}_i) = y_i - {\hat y}_i^{cv}.
\label{residuals_cv}
\end{equation}
Based on the prediction residuals we can estimate the root mean squared error 
\begin{equation}
RMSE^{cv} = \sqrt{\frac{1}{n}\sum_{i=1}^n ( y_i - {\hat y}_i^{cv})^2},
\label{rmse_pred_eq}
\end{equation}
and the Coefficient of Prognosis \cite{Most_2011_WOST}
\begin{equation}
CoP=1-\frac{SS_E^{cv}}{SS_T}, \quad SS_E^{cv}=\sum_{i=1}^n(y_i-{\hat y}_i^{cv})^2.
\label{cop_eq}
\end{equation}
The CoP quantifies the explained variation in the support data points similarly to the CoD, but the 
prediction errors estimated with the cross-validation procedure are considered instead of the pure fitting residuals. 

\begin{figure}[th]
\includegraphics[width=0.49\textwidth]{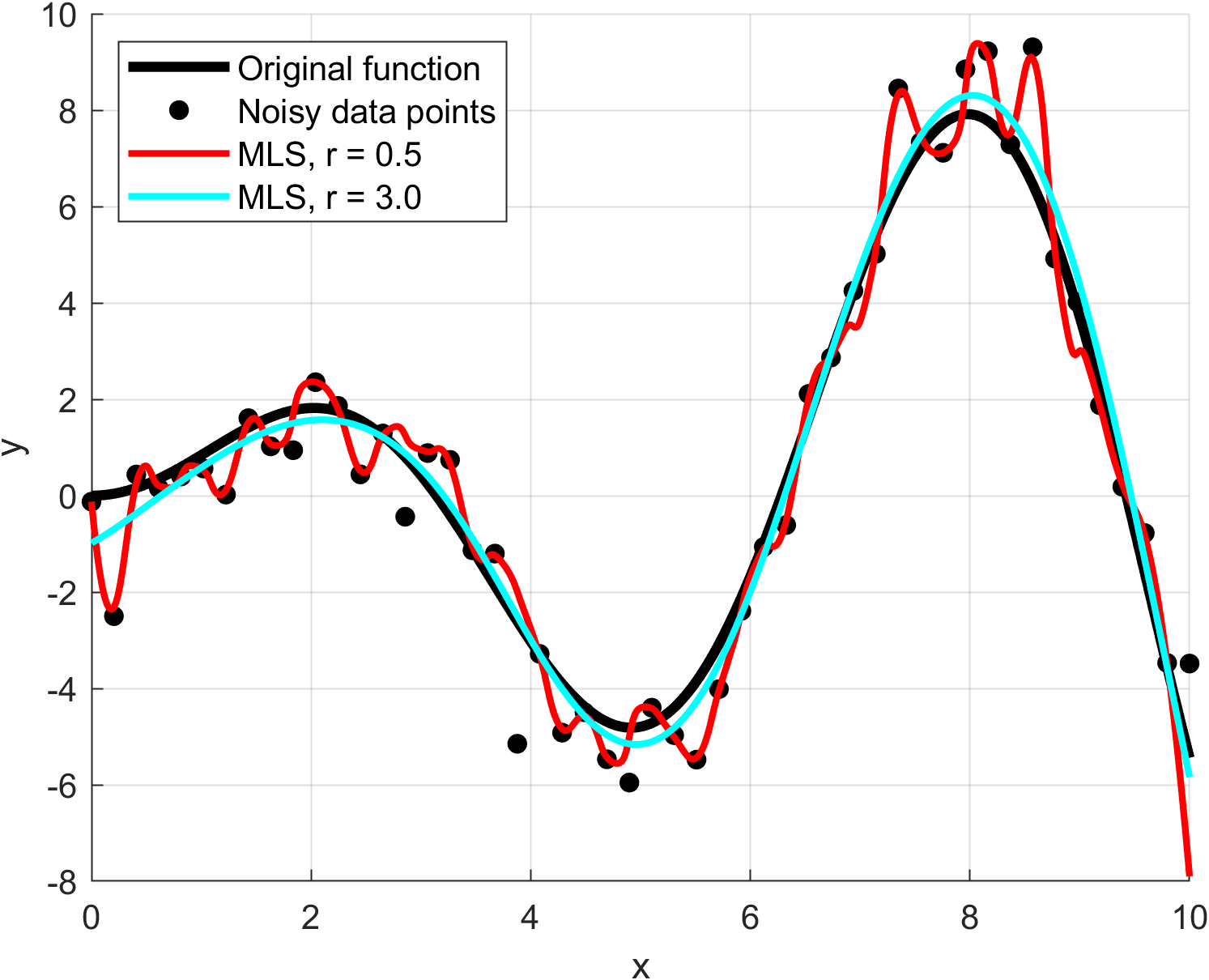}
\hfill
\includegraphics[width=0.49\textwidth]{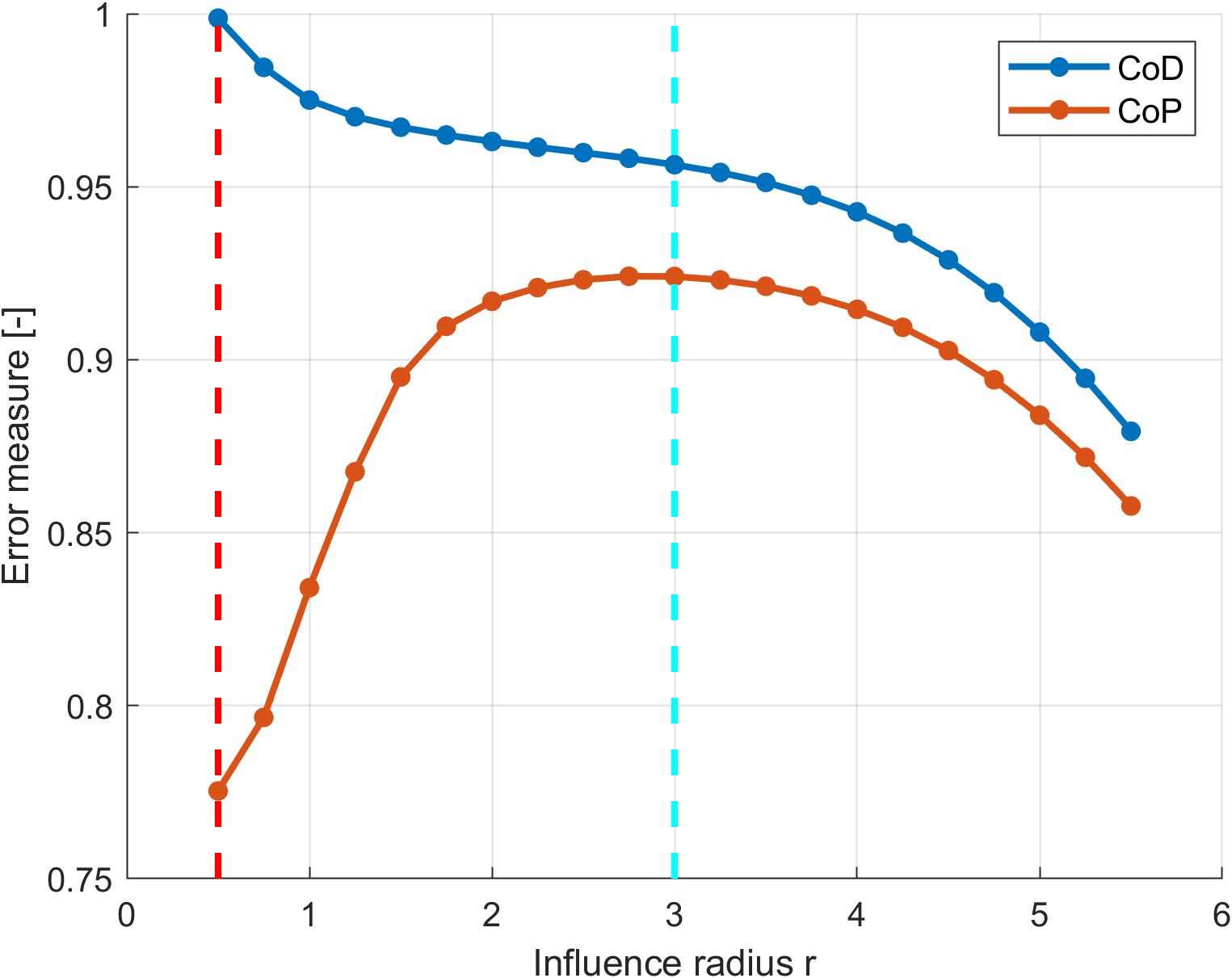}
\caption{Moving Least Squares approximation of noisy data points of a non-linear function for different values of the influence radius and corresponding difference between CoD and CoP}
\label{cop_fig} 
\end{figure}
In figure~\ref{cop_fig} an illustrative example of an one-dimensional non-linear function is given to demonstrate the differences between these two measures.
The CoD and CoP are evaluated using equation~\ref{CoD_eq2} and \ref{cop_eq} with 5 subsets for the cross-validation procedure. As approximation model a Moving Least Squares (MLS) approximation 
\cite{Lancaster1981} with quadratic basis is applied. Further details about the implementation can be found in \cite{Most_2011_WOST}.
In figure~\ref{cop_fig} the CoD and CoP are shown depending on the influence radius $r$ of the MLS approximation. If the radius is chosen very small, the approximation model will tend to fit well true the noisy support points. As a result, the CoD is close to one. This case is called over-fitting. With increasing influence radius, the approximation model becomes smoother and filters the noise in the supports more efficiently. If the radius is chosen to large, the model will tend to the quadratic basis function and the approximation  becomes inadequate. The CoP will indicate a poor model quality if the radius is chosen to small in contrast to the CoD. For a large radius both measures approach to each other and both indicate a poor approximation. The radius with the maximum CoP is the optimal choice for this example and will result in a suitable approximation model as indicated in figure~\ref{cop_fig}. This simple example shows, that a model evaluation, comparison and possible selection based on the CoP would be much more suitable in order to get the best prediction quality for a given support data set. For models with high flexibility, it could support the appropriate choice of the model parameters in order to prevent  over-fitting for noisy data.

The numerical implementation of the k-fold cross validation procedure is straight-forward for the most classical surrogate models. Our implementation, which is available in the Ansys optiSLang software package \cite{optislang2023}, considers linear regression, Moving Least Squares, Radial Basis Functions and Kriging with up to 10.000 data points and requires just a small amount of additional numerical effort compared to the model hyper-parameter search. 
More challenging is the implementation for complex deep learning models, since a re-training for each data subset will not always converge to the same global functions and might stuck in different local optima. To overcome this issue, we developed a specific, regularized training procedure based on a hybrid approach for the tuning of the optimal network architecture and the evaluation of the prognosis measure. Further details on this approach could be found in \cite{Most_2022_NAFEMS_DACH1},\cite{Abdulhkim_2022_Patent}.

Some mathematical surrogate models provide also closed form solutions for leave-one-out (LOO) cross validation, where each data set belongs just to a single sample. This so-called leave-one-out (LOO) cross validation is very attractive from the computational point of view. However, in our examples we will show, that the LOO cross validation may be to optimistic as an error estimator and the k-fold cross validation gives more reliable results.

\begin{figure}[th]
\center
	\includegraphics[width=0.48\textwidth]{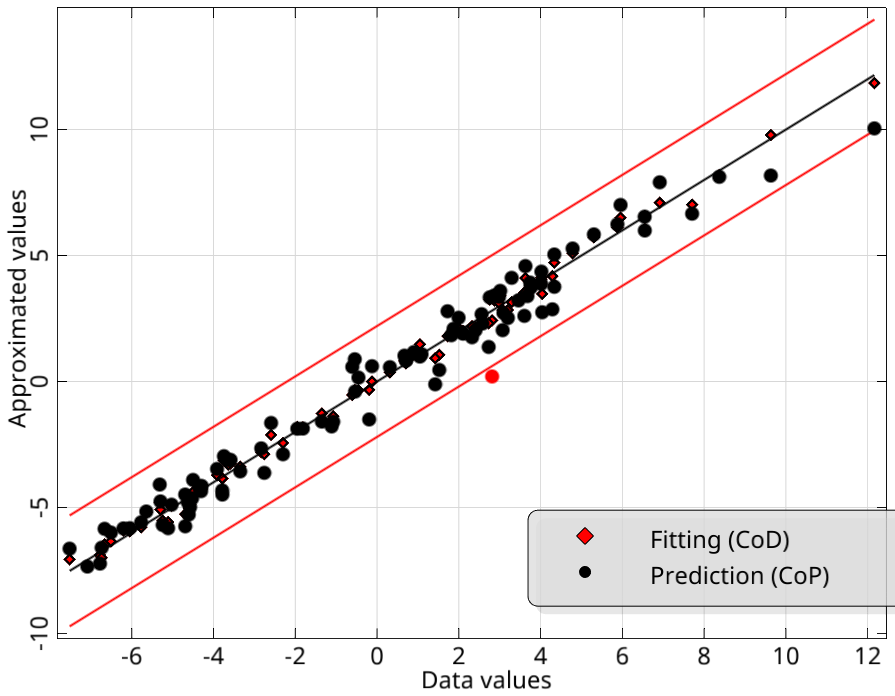}
	\hfill
	\includegraphics[width=0.48\textwidth]{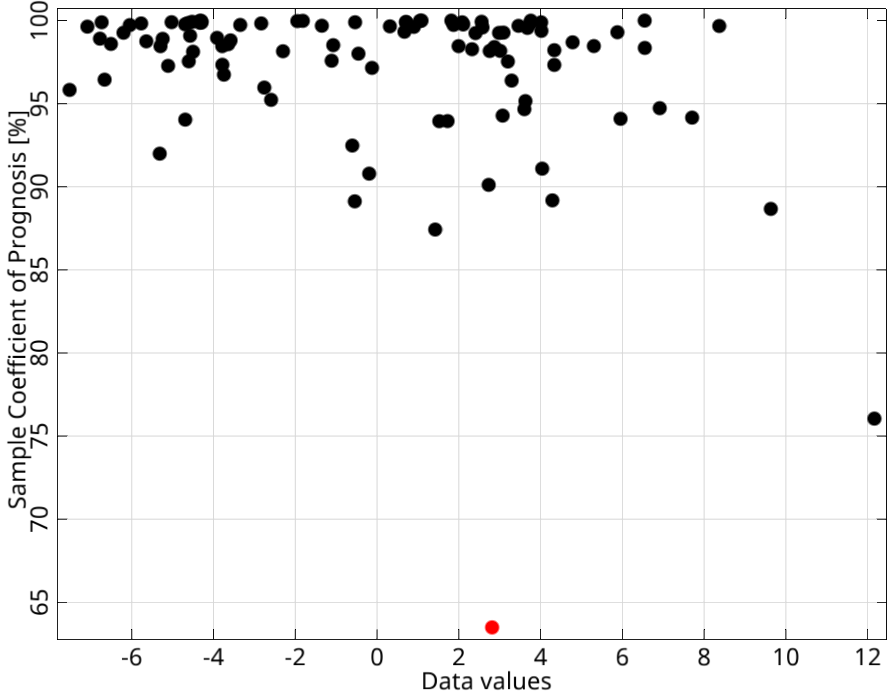}
\caption{Residual plot with the fitting and prediction residuals (left) and sample CoP, which quantifies the contribution of each sample to the CoP (right). One possible outlier is indicated in red which is outside the range of $\pm 3\times RMSE^{cv}$.}
\label{res_fig}
\end{figure}
The residuals of the goodness of fit in equation \ref{res_cod} and of the cross-validation residuals can be displayed in a so-called residual plot as shown in figure \ref{res_fig}. If a large deviation of the residuals from the fit and the prediction can be observed, we can assume that the applied surrogate model tends to over-fitting. 
The estimated RMSE in equation \ref{rmse_pred_eq} can be used to identify possible outliers. Since the RMSE can be understood as the standard deviation of the approximation error, we can assume a boundary of about $\pm 3\times RMSE^{cv}$ to check for outliers. In the residual plot in 
figure \ref{res_fig}, this is indicated as the two red lines.
In order to get an estimate, how the residuals of an individual support point $\mathbf{x}_i$ contribute to the CoP, we can further formulate the sample CoP as follows
\begin{equation}
CoP_{\mathbf{x}_i} =1-\frac{(y_i-{\hat y}_i^{cv})^2}{SS_T},
\end{equation}
which is shown additionally in figure \ref{res_fig}. The figure clearly indicates, that the sample CoP may help to detect outliers more clearly by using a different scaling. The mean value of all individual sample CoPs is consequently the global CoP value introduced in equation \ref{cop_eq}.

\subsection{Local measures of the prognosis quality}
\begin{figure}[th]
\center
	\includegraphics[width=0.48\textwidth]{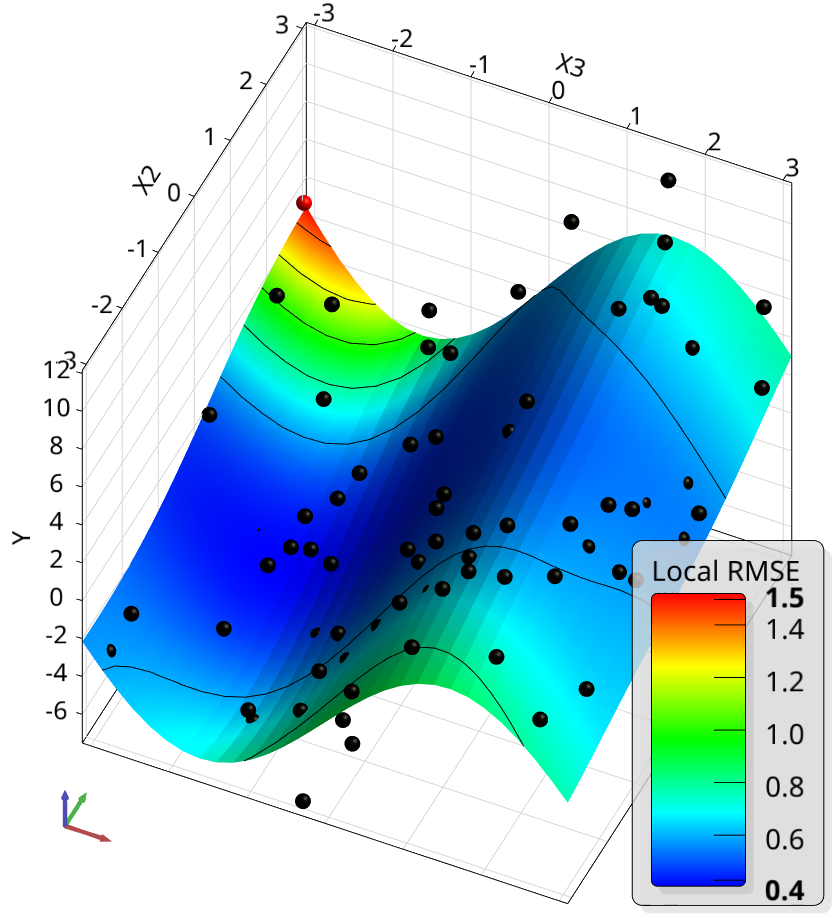}
	\hfill
	\includegraphics[width=0.48\textwidth]{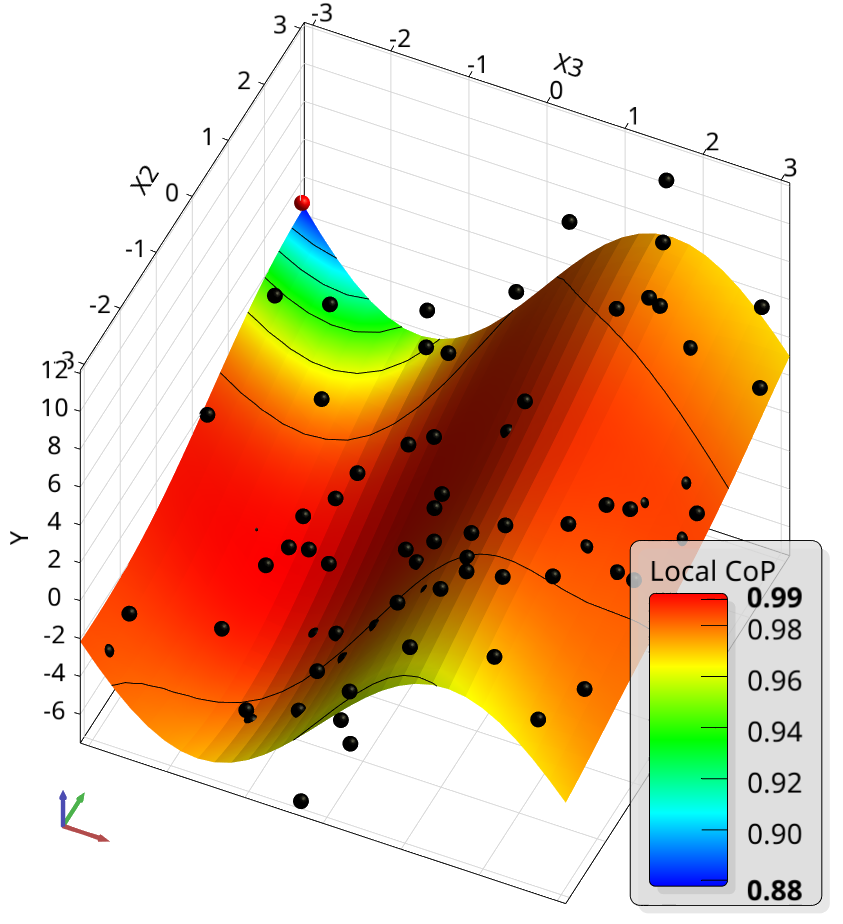}
\caption{Estimated local root mean squared error (left) and the local Coefficient of Prognosis (right) as subspace plot in a 5D input space. }
\label{local_cop_fig}
\end{figure}

Based on the individual residuals of each support point we can formulate a continuous function of the local prediction error for an arbitrary point in the input space. By using a local averaging scheme similar to the Moving Least Squares approximation \cite{Lancaster1981}
the locally weighted RMSE and the local CoP read as follows
\begin{equation}
RMSE^{cv}(\mathbf{x})=\sqrt{\frac{\sum_{i=1}^n w_i(\mathbf{x}) (y_i-{\hat y}^{cv}_i)^2}{\sum_{i=1}^n w_i (\mathbf{x})}},
\label{rmse_mop_local}
\end{equation}

\vspace{-5mm}
\begin{equation}
CoP(\mathbf{x})=1-\frac{\sum_{i=1}^n w_i(\mathbf{x}) (y_i-{\hat y}^{cv}_i)^2}{\sum_{i=1}^n w_i (\mathbf{x})}\cdot \frac{n}{SS_T}= 1-\frac{n\cdot (RMSE^{cv}(\mathbf{x}))^2}{SS_T},
\end{equation}
where $w_i(\mathbf{x})$ is chosen as an exponential, isotropic weighting function, which is scaled with respect to the number of necessary averaging points.
In figure \ref{local_cop_fig} the estimated local prediction errors are shown for the residuals from figure \ref{res_fig}. The figure indicates, that in the region of the identified outlier the approximation quality is worst.

The presented local prediction errors can be easily utilized in an adaption scheme such as the expected improvement criterion according to \cite{jones1998efficient}. The advantage of this error estimator is its independence w.r.t. the approximation model. Thus it can be applied for simple polynomial models in the same manner as for more sophisticated deep learning networks. This estimator has been applied in the Adaptive Metamodel of Optimal Prognosis (AMOP) \cite{optislang2023} in the Ansys optiSLang software package. 
With help of the local RMSE the prediction uncertainty of an investigated surrogate model can be interpreted as a normally distributed random process, where the mean corresponds to the model approximation itself and the standard deviation to the estimated local RMSE.

\subsection{Quantification of the input sensitivity}
\label{saltelli}

If we interpret the scalar output of a simulation model as well as the $m$ input parameters as the random numbers $Y$ and  $X_i$,
the first order sensitivity index can be formulated for the input $X_i$ according to \cite{Sobol1993} as follows
\begin{equation}
S_i=\frac{V_{X_i}(E_{\mathbf{X}_{\sim i}}(Y|X_i))}{V(Y)},
\label{first}
\end{equation}
where $V(Y)$ is the unconditional variance of the model output $Y$ and 
$V_{X_i}(E_{\mathbf{X}_{\sim i}}(Y|X_i))$ is called the {\it variance of  conditional expectation}
with $\mathbf{X}_{\sim i}$ denoting the matrix of all inputs without $X_i$.
Thus, $V_{X_i}(E_{\mathbf{X}_{\sim i}}(Y|X_i))$ measures the first order effect of $X_i$ on the model output.
The first order sensitivity indices quantify only the contribution of a single input to the output variance, but not the interaction with other inputs. In order to quantify higher order interaction terms, the total effect sensitivity indices 
have been introduced in \cite{Homma1996} as follows
\begin{equation}
S_{Ti}=1-\frac{V_{\mathbf{X}_{\sim i}}(E_{X_i}(Y|\mathbf{X}_{\sim i}))}{V(Y)},
\label{total}
\end{equation}
where $V_{\mathbf{X}_{\sim i}}(E_{X_i}(Y|\mathbf{X}_{\sim i}))$ measures the first order effect of $\mathbf{X}_{\sim i}$
on the model output which does not contain any effect corresponding to $X_i$.

In our implementation we use the first order and total effect indices to quantify the importance of the 
inputs with respect to the approximated model output $\hat y (\mathbf{x})$. Since the approximation model is not representing the 
full variance of the original response $y(\mathbf{x})$, we use the estimated CoP to quantify this proportion
\begin{equation}
\hat S_i^{cv}= CoP \cdot \hat S_i, \quad \hat S_{Ti}^{cv} = CoP \cdot \hat S_{Ti}, 
\end{equation}
where the $CoP$ is estimated using equation \ref{cop_eq} and $\hat S_i$ and $\hat S_{Ti}$ are the estimated first order and total effect indices using the surrogate model approximation instead of the original model, whereby the input variables are assumed to be uniformely distributed within the given sampling bounds. Further details on the computation of the estimators of $\hat S_i$ and $\hat S_{Ti}$ can be found in \cite{Most_2012_REC1},\cite{Saltelli_2008_Book}. 
For correlated inputs a modification of the estimators is necessary which is explained in detail in \cite{Most_2012_REC1}.

\subsection{Estimation of confidence bounds using bootstrapping}

\begin{figure}[h]
\begin{center}
\begin{tikzpicture}[node distance = 3.5cm, auto,
        IS/.style={blue, very thick},
        axis/.style={very thick, ->, >=stealth', line join=miter},
        important line/.style={thick}, dashed line/.style={dashed, thin},
        every node/.style={color=black},
        dot/.style={circle,fill=black,minimum size=4pt,inner sep=0pt,
            outer sep=-1pt},
]
	\tikzstyle{observation} = [rectangle, draw, fill=black!20,  node distance=0.7cm,
    	text width=4em, text centered, minimum height=2.5em]
	\tikzstyle{bootstrap} = [rectangle, draw, fill=black!20,  node distance=0.7cm,
    	text width=9em, text centered, minimum height=2.5em]
	\tikzstyle{line} = [draw, -latex']
    \tikzstyle{annot} = [text width=15em, text centered,minimum height=2.5em]
	\tiny
    \node[annot] (obs) {\small Observations};
    \node[annot,right of=obs, node distance=2.5cm] (choice) {};
    \node[annot,right of=choice, node distance=2.5cm] (boot) {\small Bootstrap set $j$};
    \node[annot,below of=choice, node distance=3.15cm] (choice2) {\small Random choice};
    \node[annot,below of=choice, node distance=3.5cm] (choice3) {\small with replacement};
	\node [observation, below of=obs,node distance=0.5cm] (x1) {\small $\epsilon_1^{cv}$};
	\node [observation, below of=x1] (x2) {\small $\epsilon_2^{cv}$};
	\node [observation, below of=x2] (x3) {\small $\epsilon_3^{cv}$};
	\node [observation, below of=x3] (x4) {\small $\epsilon_4^{cv}$};
	\node [observation, below of=x4] (x5) {\small $\epsilon_5^{cv}$};
	\node [bootstrap, below of=boot, node distance = 0.5cm] (b1) {\small $\epsilon^*_{1,j} = \epsilon_3^{cv}$};
	\node [bootstrap, below of=b1] (b2) {\small $\epsilon^*_{2,j} = \epsilon_1^{cv}$};
	\node [bootstrap, below of=b2] (b3) {\small $\epsilon^*_{3,j} = \epsilon_5^{cv}$};
	\node [bootstrap, below of=b3] (b4) {\small $\epsilon^*_{4,j} = \epsilon_3^{cv}$};
	\node [bootstrap, below of=b4] (b5) {\small $\epsilon^*_{5,j} = \epsilon_2^{cv}$};

    \path [line] (b1) -- (x3);
    \path [line] (b2) -- (x1);
    \path [line] (b3) -- (x5);
    \path [line] (b4) -- (x3);
    \path [line] (b5) -- (x2);

\end{tikzpicture}
\end{center}
\vspace{-5mm}
\caption{Principle of non-parametric bootstrap method: generation of a bootstrap sample set from the original residual set by random choice with replacement}
\label{boot_principle}
\end{figure}
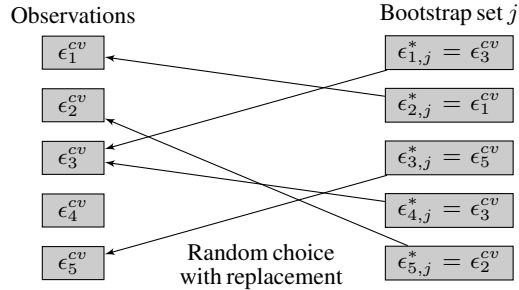

Once the cross-validation residuals and the prediction quality estimators have been evaluated, one may need further information on the confidence bounds of these estimators. For this purpose, we apply the bootstrapping method introduced in \cite{Efron1992}. In this method the statistical properties of an estimator are obtained by sampling from an approximate distribution which can be the empirical distribution of the observed data or a parametrized form of this distribution. In our study we use the most common approach, the non-parametric bootstrapping, where the sampling is done directly from the empirical distribution of the original observations. This method assumes independent and identically distributed observations and constructs a number of re-samples from the original samples. In \cite{Most_2010_CompGeo} this procedure is discussed in detail for the estimation of statistical moments of material properties.

\begin{figure}[h]
\center
	\includegraphics[width=0.48\textwidth]{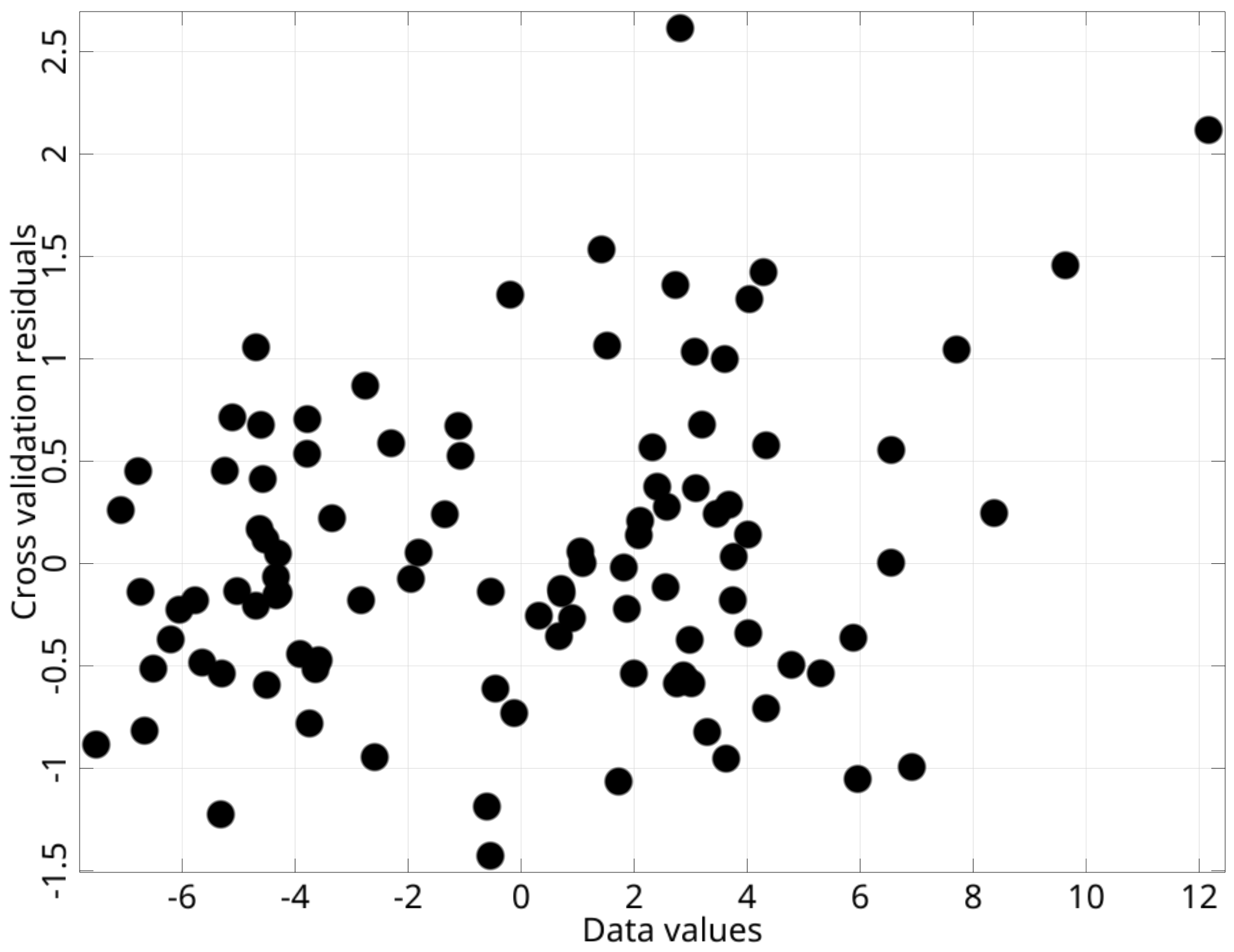}
	\hfill
	\includegraphics[width=0.48\textwidth]{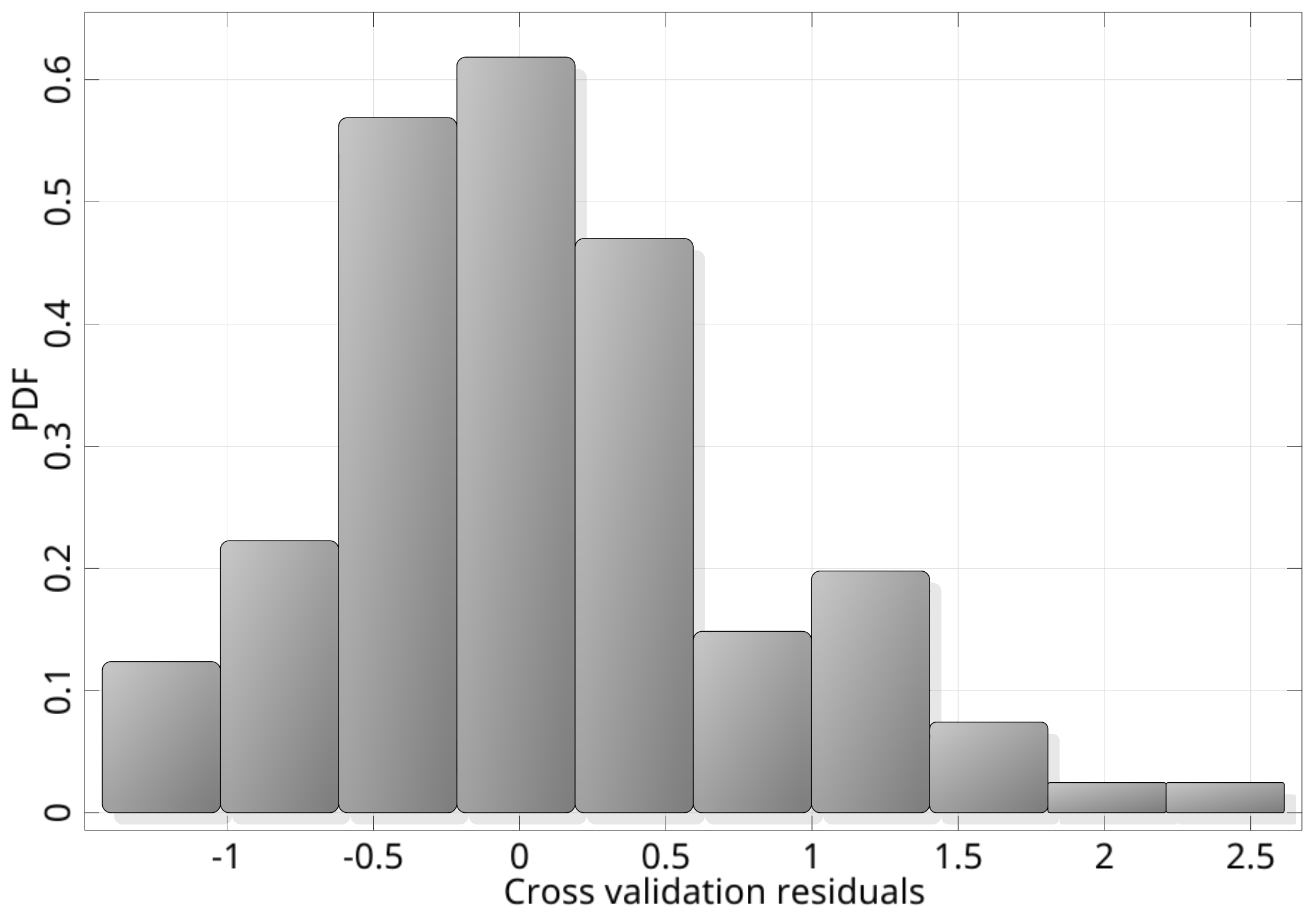}\\
	\vspace{3mm}
	\includegraphics[width=0.48\textwidth]{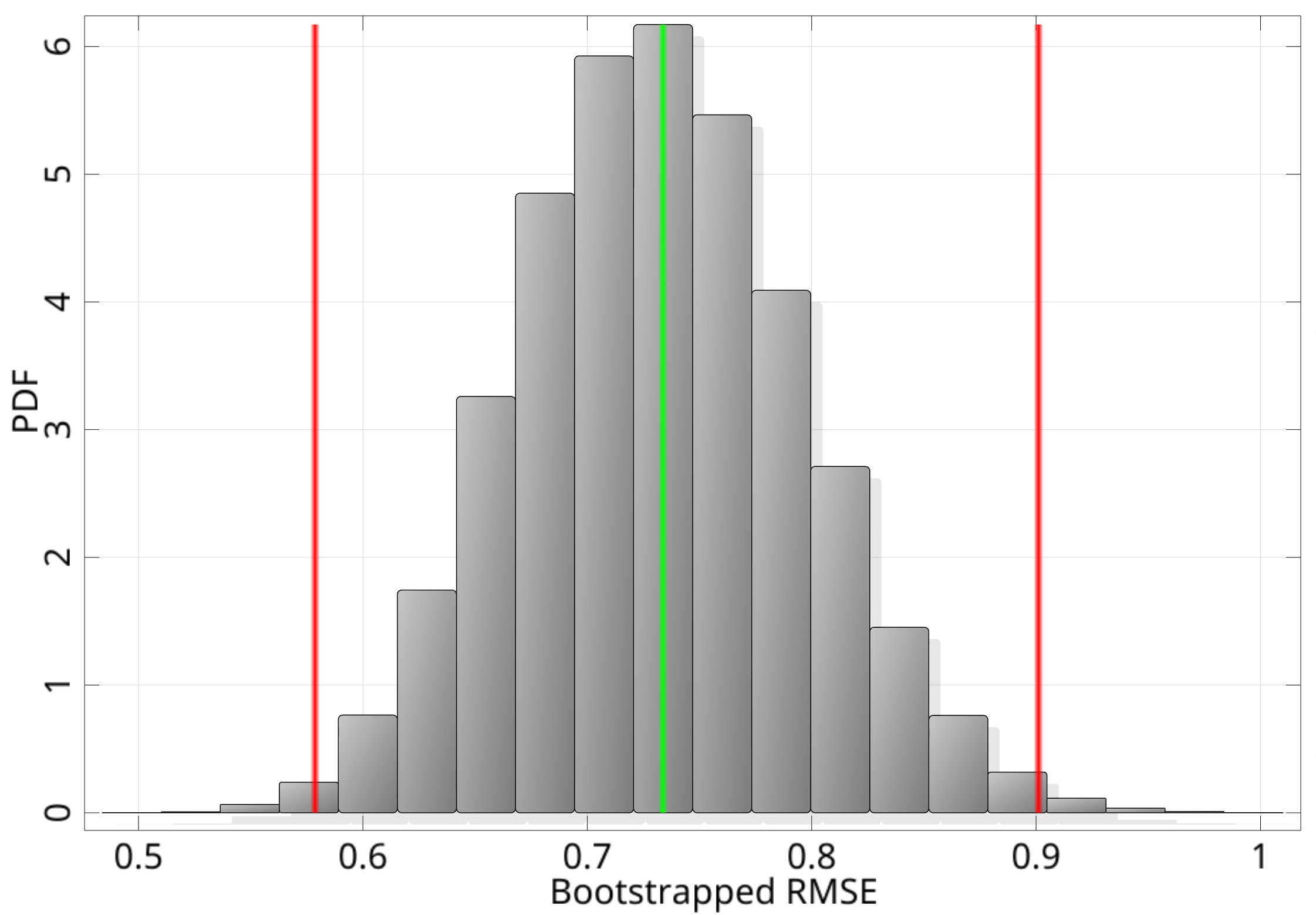}
	\hfill
	\includegraphics[width=0.48\textwidth]{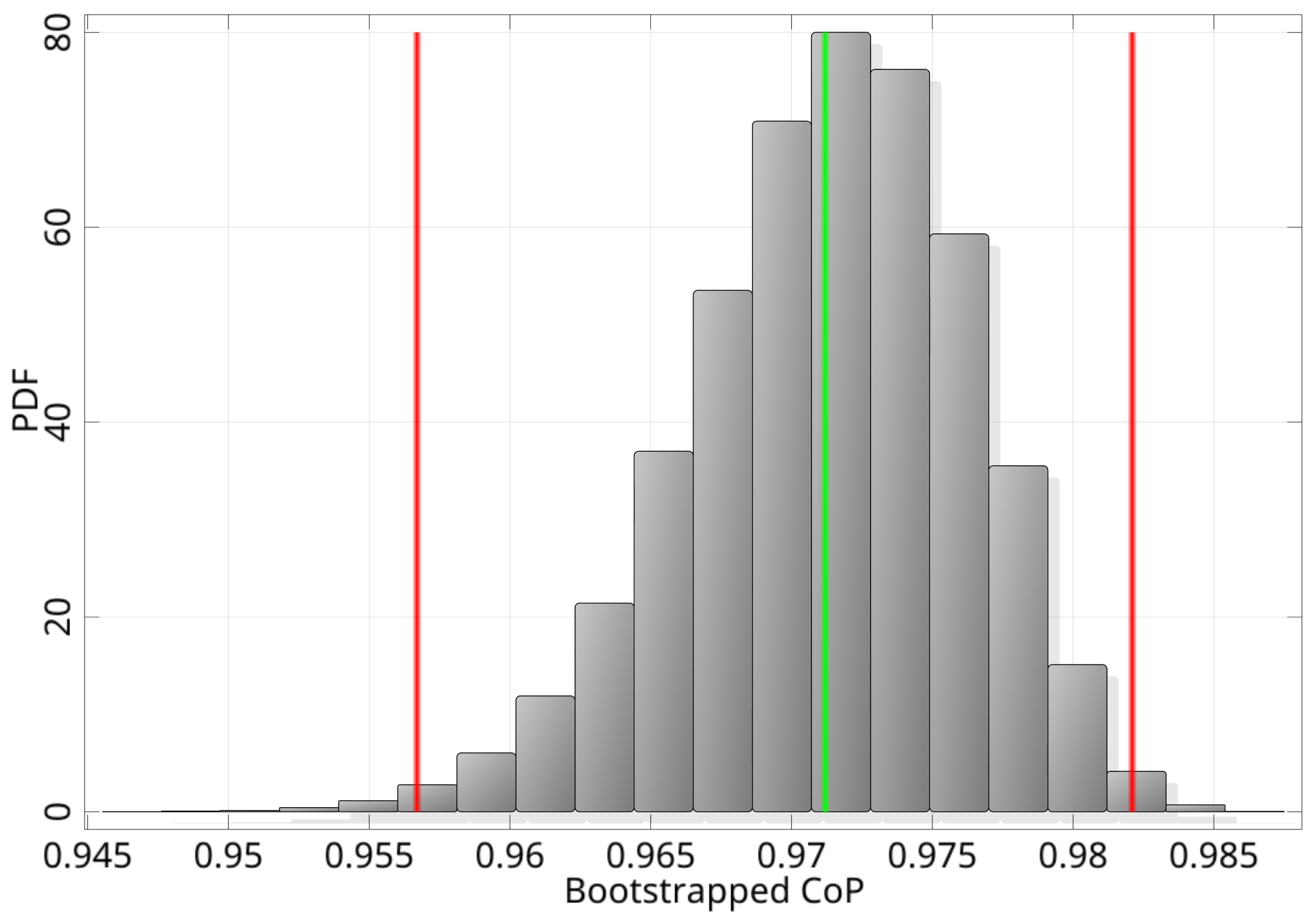}\\
\caption{Cross validation residuals of 100 support points: distribution and histogram (top) and bootstrapped RMSE and CoP (bottom) with deterministic estimates (green) and 99\% confidence interval (red) }
\label{residuals_fig}
\end{figure}

In our study we assume the cross validation residuals of the approximation function in equation \ref{residuals_cv} as independent observations of an unknown random number.
From this original set of observations $\epsilon_1^{cv}, \epsilon_2^{cv}, \ldots, \epsilon_n^{cv}$ a bootstrap sample set $\mathbf{B}_j = \epsilon^*_{1,j}, \epsilon^*_{2,j}, \ldots, \epsilon^*_{n,j}$ 
with $n$ samples is chosen by random sampling with replacement from the observation data set as illustrated in figure~\ref{boot_principle}.
In this set each observation $\epsilon_i^{cv}$ may appear once, more than once or not at all.
This procedure is repeated with a large number of repetitions and the presented quality measures are estimated for each bootstrap sample set $\mathbf{B}_j$ as follows
\begin{equation}
RMSE_{\mathbf{B}_j} = \sqrt{\frac{1}{n}\sum_{i=1}^n \left(\epsilon^*_{i,j}\right)^2},\quad CoP_{\mathbf{B}_j}=1-\frac{\sum_{i=1}^n \left(\epsilon^*_{i,j}\right)^2}{SS_T} .
\end{equation}
From the individual results of each bootstrap set $\mathbf{B}_j$, the statistical properties of the RMSE and CoP estimates can be evaluated.
In figure \ref{residuals_fig} the 100 cross validation residuals of the previous example plots are shown. The anthill plot indicates an almost independent relation between the data values and the residuals. However, the histogram is non-symmetric and indicates a skewed distribution. For these residuals the bootstrap resampling is applied using $10^5$ repetitions and the statistical measures are evaluated for each of the bootstrap samples. In figure \ref{residuals_fig} the histograms of the corresponding RMSE and CoP are shown including the 99\% confidence intervals, which can be directly estimated from the bootstrap samples. The figure indicates an almost symmetric distribution of the RMSE, which would fit to a normal distribution very well. The distribution of the CoP is non-symetric and skewed, which means that the mean value and the standard deviation might be not sufficient to characterize the confidence interval. Therefore, we calculate the confidence interval of each quality measure in the following examples directly from the bootstrap samples without assuming any distribution.

The benefit in bootstrapping the residuals directly instead of building up new surrogate models for each bootstrap set is clearly the reduction of the numerical effort. Once the cross validation residuals are obtained for a given support point set, the bootstrapping and the evaluation of the CoP distribution can be performed very cheap. However, the estimator will not cover the case, that the support points do not have a suitable distribution. Nevertheless, the confidence estimates from this procedure are quite helpful to assess the quality estimators as shown in the numerical examples.

\subsection{Extension to non-scalar outputs}

\begin{figure}[h]
\center
	\includegraphics[width=0.7\textwidth]{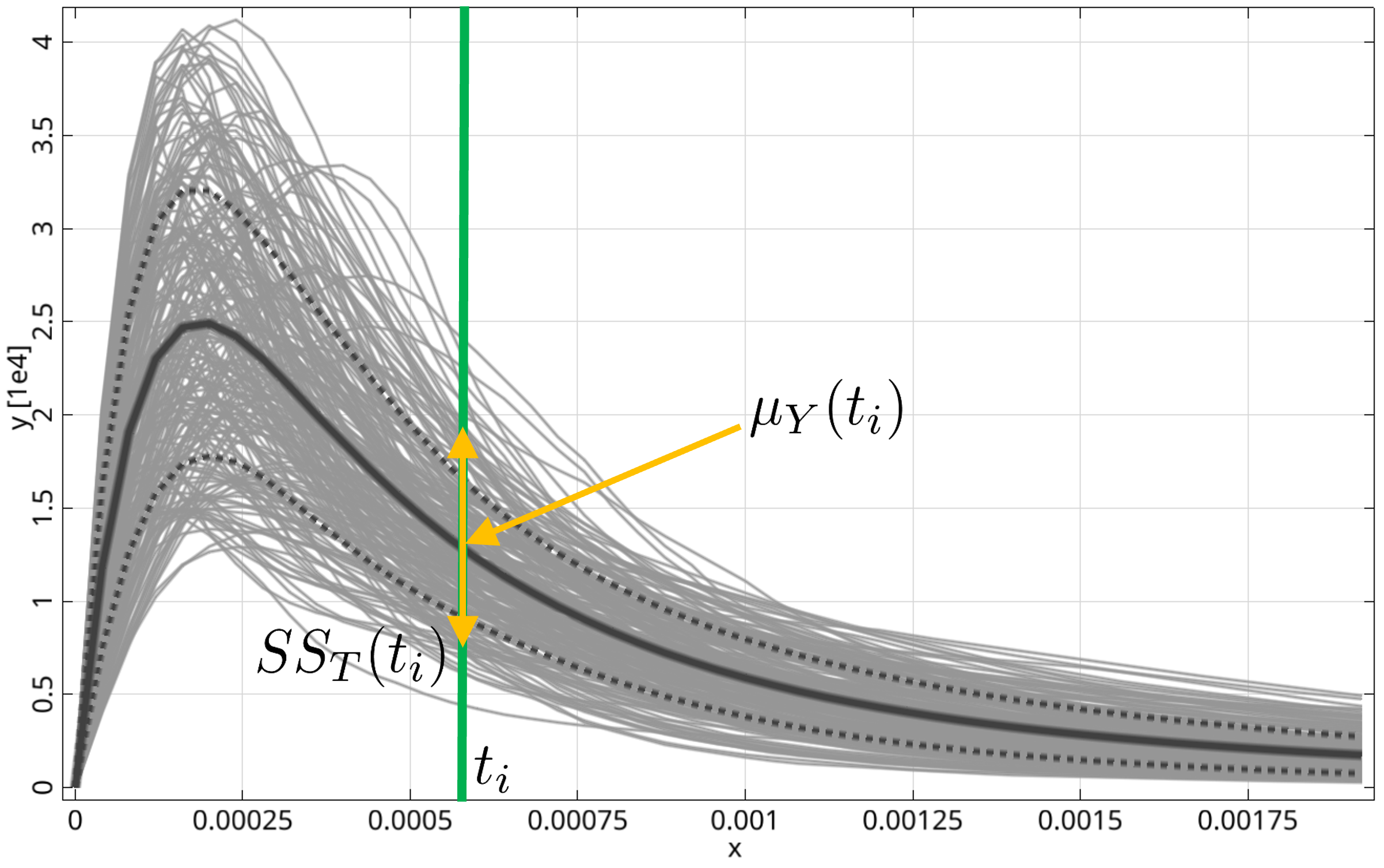}\\
 	\caption{Samples of a time-series and indicated mean and sum of total squares $SS_T$ at a certain discretization point $t_i$}
	\label{signa_samples}
\end{figure}
The non-scalar outputs of a simulation model can be described as a function of the vector of input parameters $\mathbf{x}$ and a discretization vector $\mathbf{t}$
\begin{equation}
y(\mathbf{x},\mathbf{t})=f(\mathbf{x}, \mathbf{t}).
\end{equation}
This discretization could be defined by a specific time step of a time-series output, a spatial coordinate of the stress or strain field outputs of a finite element model or a combination of spatial and time discretization.
Let us assume, that the discretization vector maps a spatial output object to a single scalar output as shown in figure for a one-dimensional time-series output.
If the mapping of the discretization is unique for every sample, we can formulate the residuals of the simulation and the approximation model as follows
\begin{equation}
\epsilon_j(t_i) = y(\mathbf{x}_j,t_i)- \hat y(\mathbf{x}_j,t_i) = y_j(t_i) - {\hat y}_j(t_i).
\label{res_cod2}
\end{equation}
With this formulation we can introduce the spatial sum of errors and sum of squares accordingly to the scalar outputs
\begin{equation}
SS_E(t_i)=\sum_{j=1}^n(y_j(t_i)-{\hat y}_j(t_i))^2, \quad SS_T(t_i)=\sum_{j=1}^n(y_j(t_i)-\mu_{Y}(t_i))^2, \quad \mu_{Y}(t_i)=\frac{1}{n}\sum_{j=1}^n y_j(t_i).
\end{equation}
The discretized formulations for $SS_E$ and $SS_T$ could be used to calculate the CoD according to equation~\ref{CoD_eq2}, which would normalize the output residuals at each discretization point individually
\begin{equation}
CoD(t_i)=1-\frac{SS_E(t_i)}{SS_T(t_i)}.
\end{equation}
If a stress field or a time-series has small variations in a certain region this normalization might be difficult to interpret since it might indicate low CoD values for similar $SS_E$ estimates just due to the different normalization with $SS_T(t_i)$.
If the spatial or time-series output is assumed to be a stationary random process the stationary CoD might be a more suitable measure for this type of applications
\begin{equation}
CoD^{stat}(t_i)=1-\frac{SS_E(t_i)}{SS_T^{stat}},
\label{CoD_stat}
\end{equation}
where the stationary sum of squares could be assumed as the stationary variance of the whole non-scalar output considering $n_d$ discretization points
\begin{equation}
SS_{T}^{stat} = \frac{1}{n_d}\sum_{i=1}^{n_d}  \sum_{j=1}^n (y_j(t_i)-\mu_{Y}(t_i))^2 = \frac{1}{n_d}\sum_{i=1}^{n_d} SS_T(t_i).
\end{equation}
Similar to the stationary CoD, we can define the stationary CoP as follows
\begin{equation}
CoP^{stat}(t_i)=1-\frac{SS_E^{cv}(t_i)}{SS_T^{stat}}, \quad SS_E^{cv}(t_i)=\sum_{j=1}^n(y_j(t_i)-{\hat y}_j^{cv}(t_i))^2.
\label{CoP_stat}
\end{equation}
The cross-validation procedure and the calculation of the residuals is straight-forward from the mathematical viewpoint, similar as for scalar outputs. However, the residuals require a unique mapping to a reference discretization.
Nevertheless, the analysis of finite element meshes with high resolution requires an efficient implementation  of the cross-validation procedure and especially the sensitivity estimation.
In \cite{Wolff_2016_RDO},\cite{Bayer_2018_Signal},\cite{Cremanns_2021_Diss},\cite{Most_2024_NAFEMS} further details on different approximation models and discretization types for non-scalar outputs are discussed.




\section{Benchmarks and applications}
\subsection{Analytical benchmark function}

\begin{figure}[th]
\center
	\includegraphics[width=0.48\textwidth]{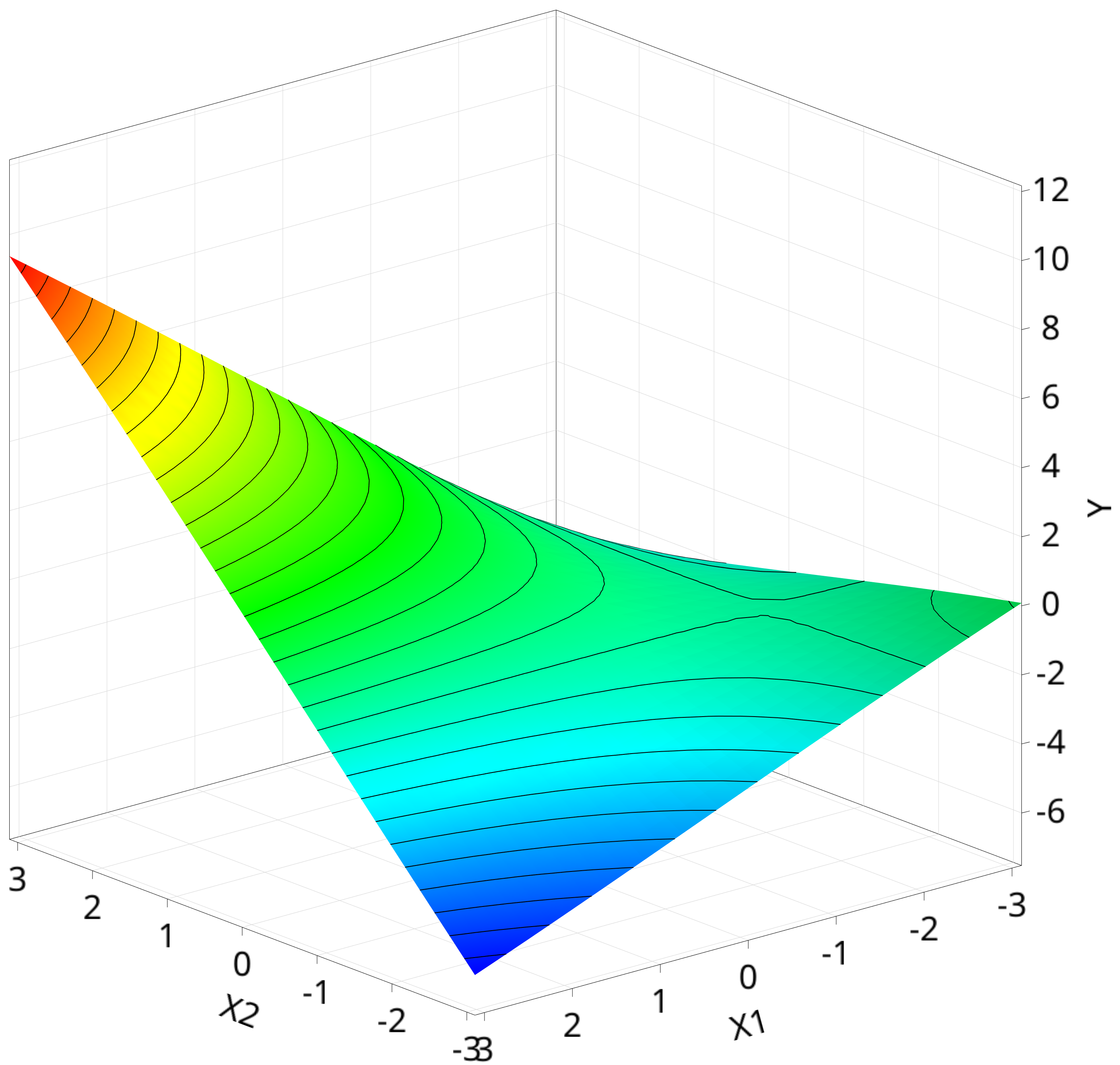}
	\hfill
	\includegraphics[width=0.48\textwidth]{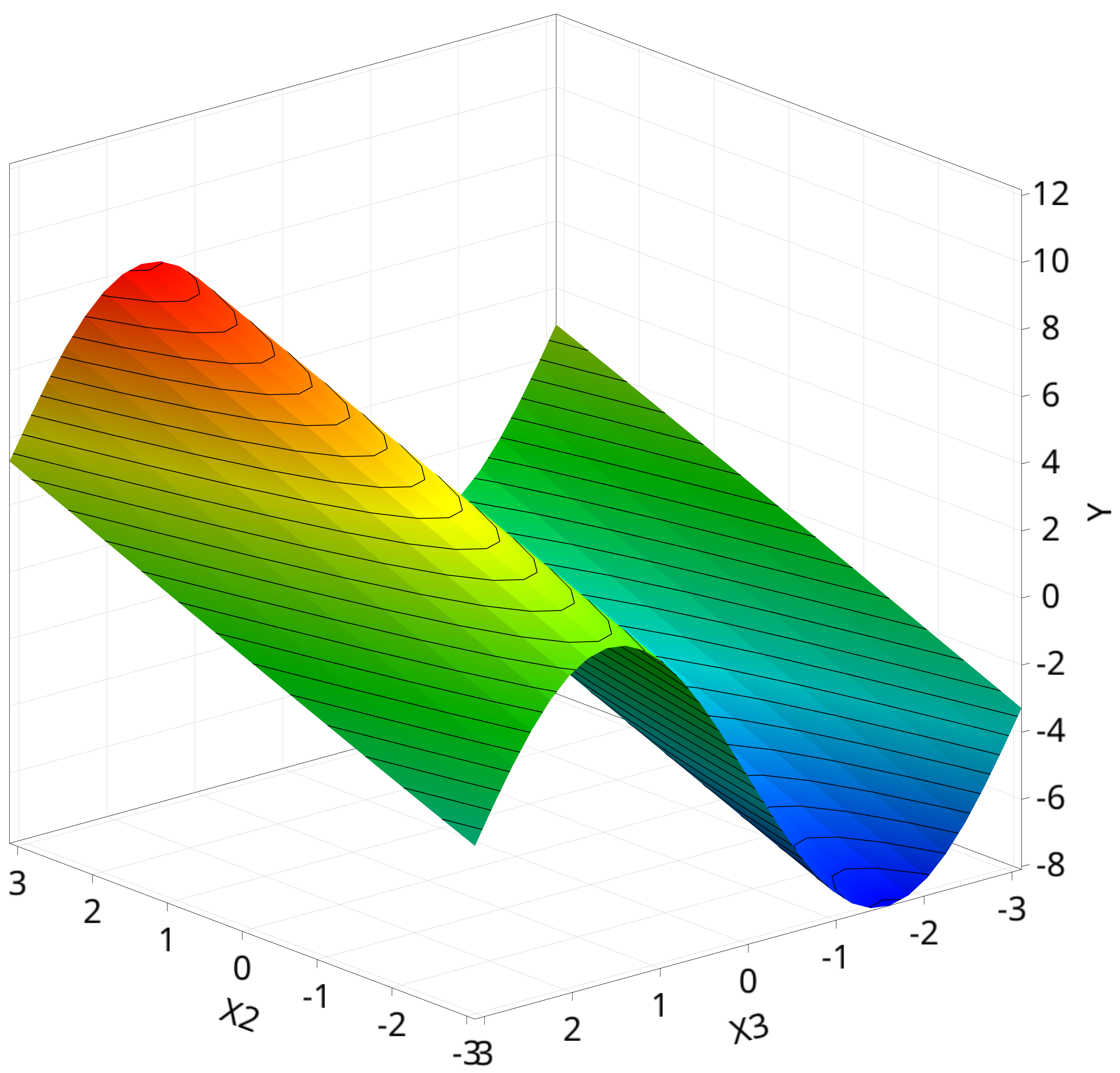}
\caption{Analytical 5D benchmark function plotted in the 2D subspaces spanned by $x_1$-$x_2$ and $x_2$-$x_3$ }
\label{coupled_fig}
\end{figure}

In a first example, we investigate an analytical benchmark function with 5 inputs
\begin{equation}
\begin{aligned}
y(\mathbf{x})=&0.5\cdot x_1+x_2+0.5\cdot x_1 x_2 + 5.0\cdot \sin(x_3) +0.2\cdot x_4 + 0.1\cdot x_5, \quad -\pi \leq x_i \leq \pi,
\end{aligned}
\label{coupled_eq}
\end{equation}
 which is shown in figure~\ref{coupled_fig} in different two-dimensional subspaces of the inputs while keeping the remaining inputs constant at the mean values.
This benchmark function was introduced in \cite{Most_2011_WOST} and consists of additive linear and non-linear terms as well as one coupling term. Furthermore, the inputs $x_4$ and $x_5$ have minor importance. 
We investigate this example 
by generating 50, 100 and 200 support points within the input bounds by using an improved Latin-Hypercube Sampling (LHS) according to \cite{Huntington1998}. An isotropic Kriging approximation model according to \cite{Forrester2008} is trained by using these support points and a k-fold and LOO cross-validation is performed to estimate the prediction errors. Unimportant inputs are removed automatically from the approximation model using the Metamodel of Optimal Prognosis approach \cite{Most_2011_WOST}.
500 additional test samples are generated by an independent LHS and are evaluated with the benchmark function. These samples are used to compared the estimated prediction errors from the cross-validation procedure with the errors in unknown data.
For this purpose the prediction sum of squares $SS_E^{cv}$ is evaluated for the cross-validation residuals and for the additional test data according to equation \ref{cop_eq}.

\begin{figure}[th]
\center
	\includegraphics[width=0.49\textwidth]{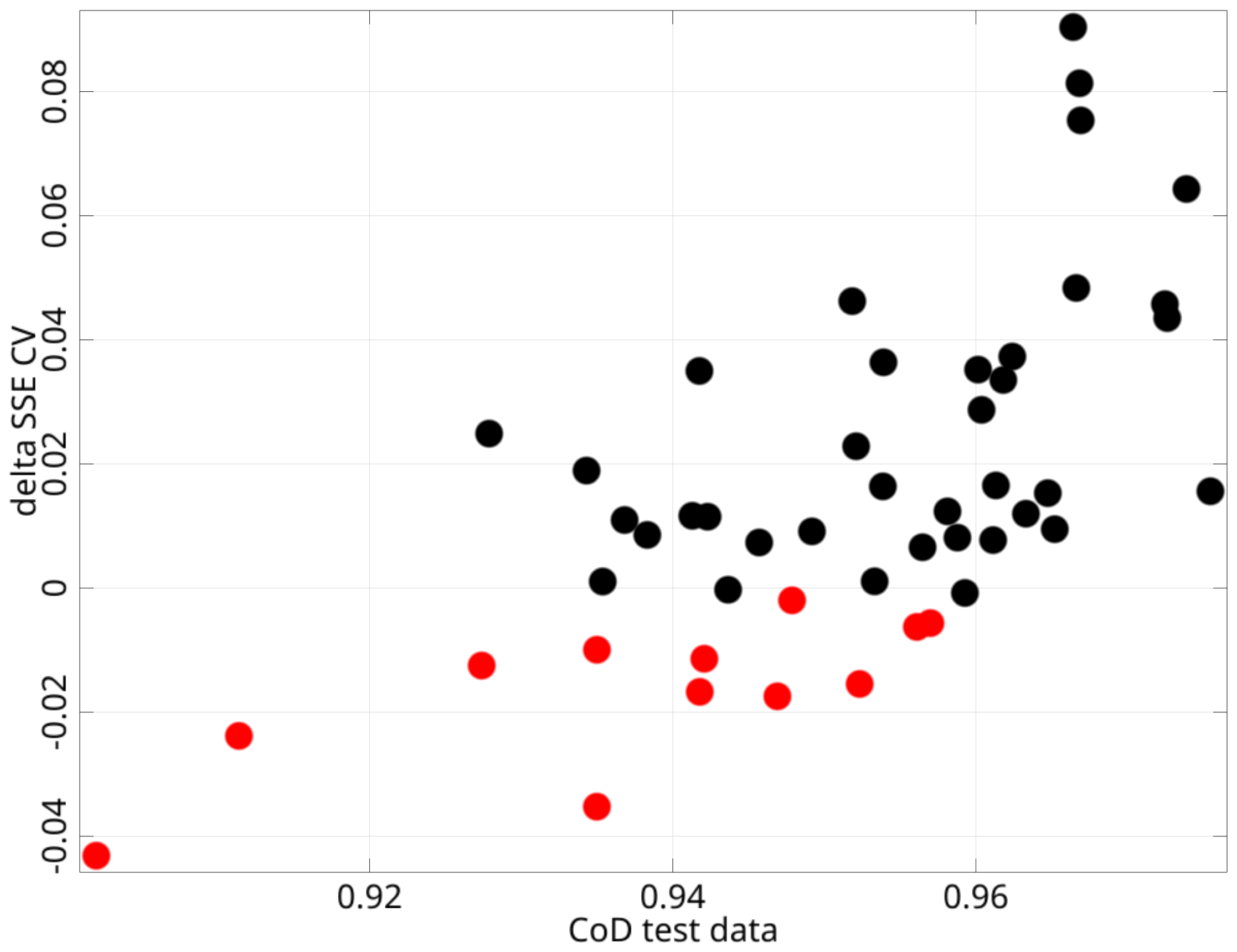}
	\hfill
	\includegraphics[width=0.49\textwidth]{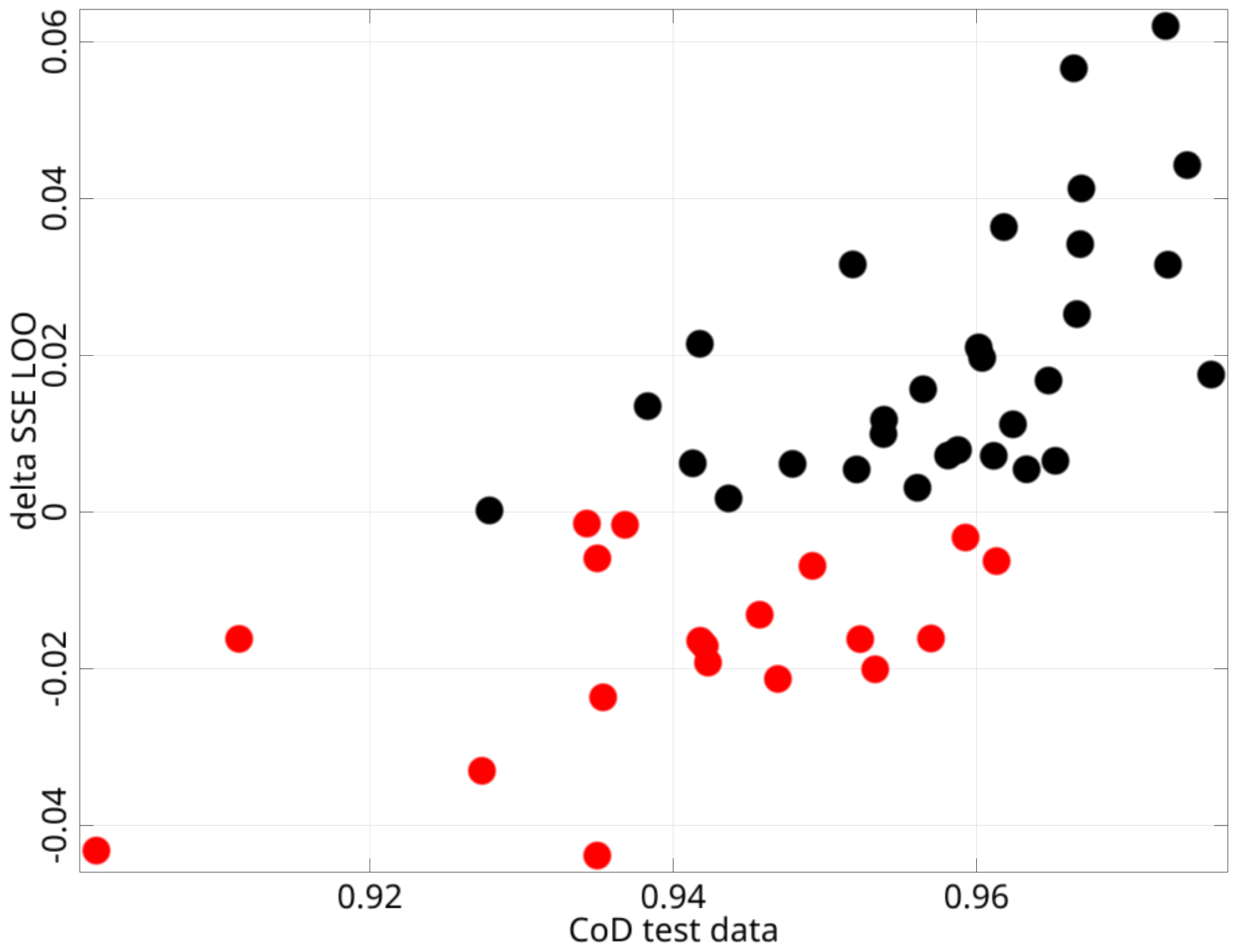}\\
 	\caption{Statistical evaluation of the prediction errors of the analytical test function by using 50 support points and 500 test points with k-fold-cross-validation (left) and  LOO-cross-validation (right)}
	\label{coupled_function_50}
	
	\vspace{1cm}
	\includegraphics[width=0.49\textwidth]{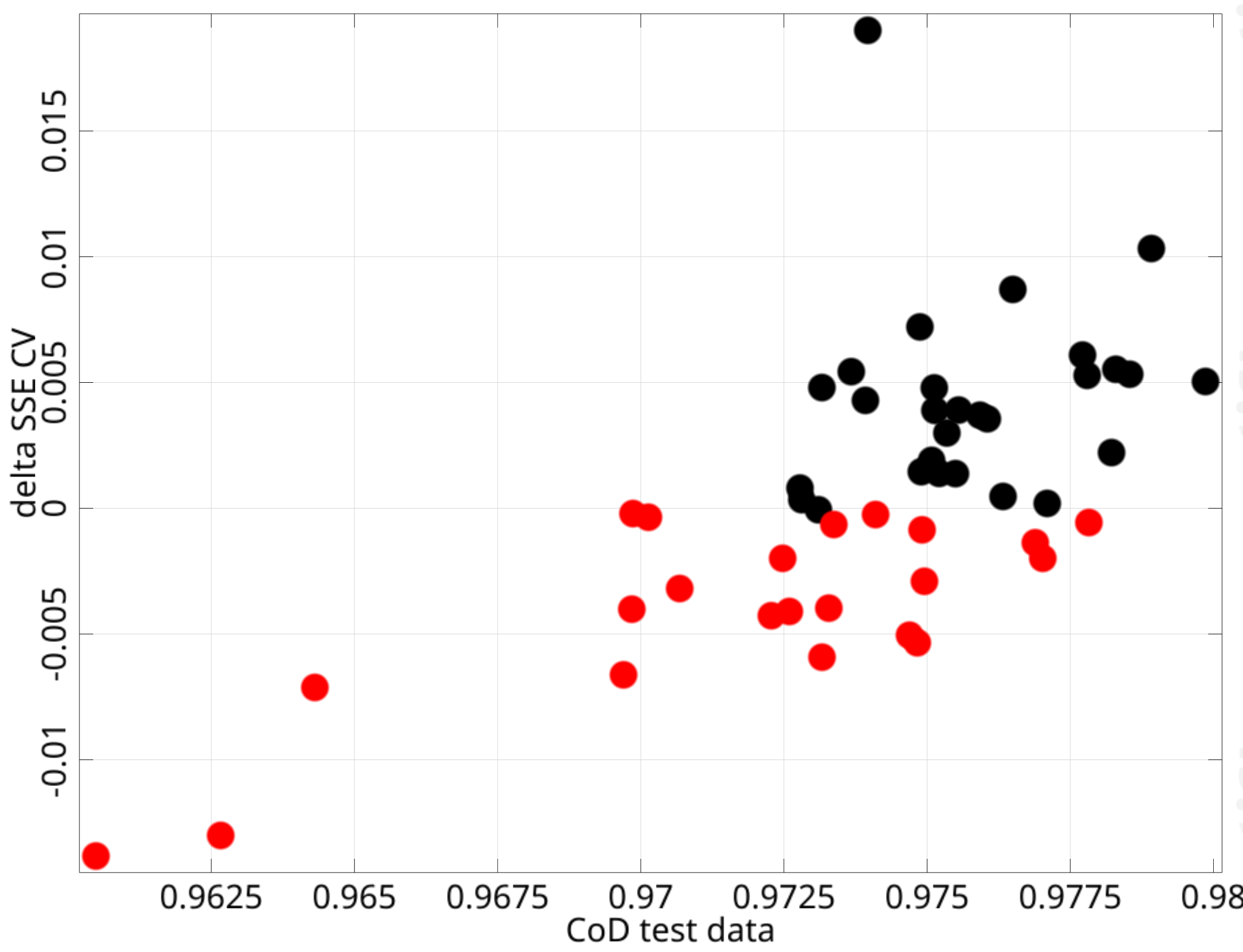}
	\hfill
	\includegraphics[width=0.49\textwidth]{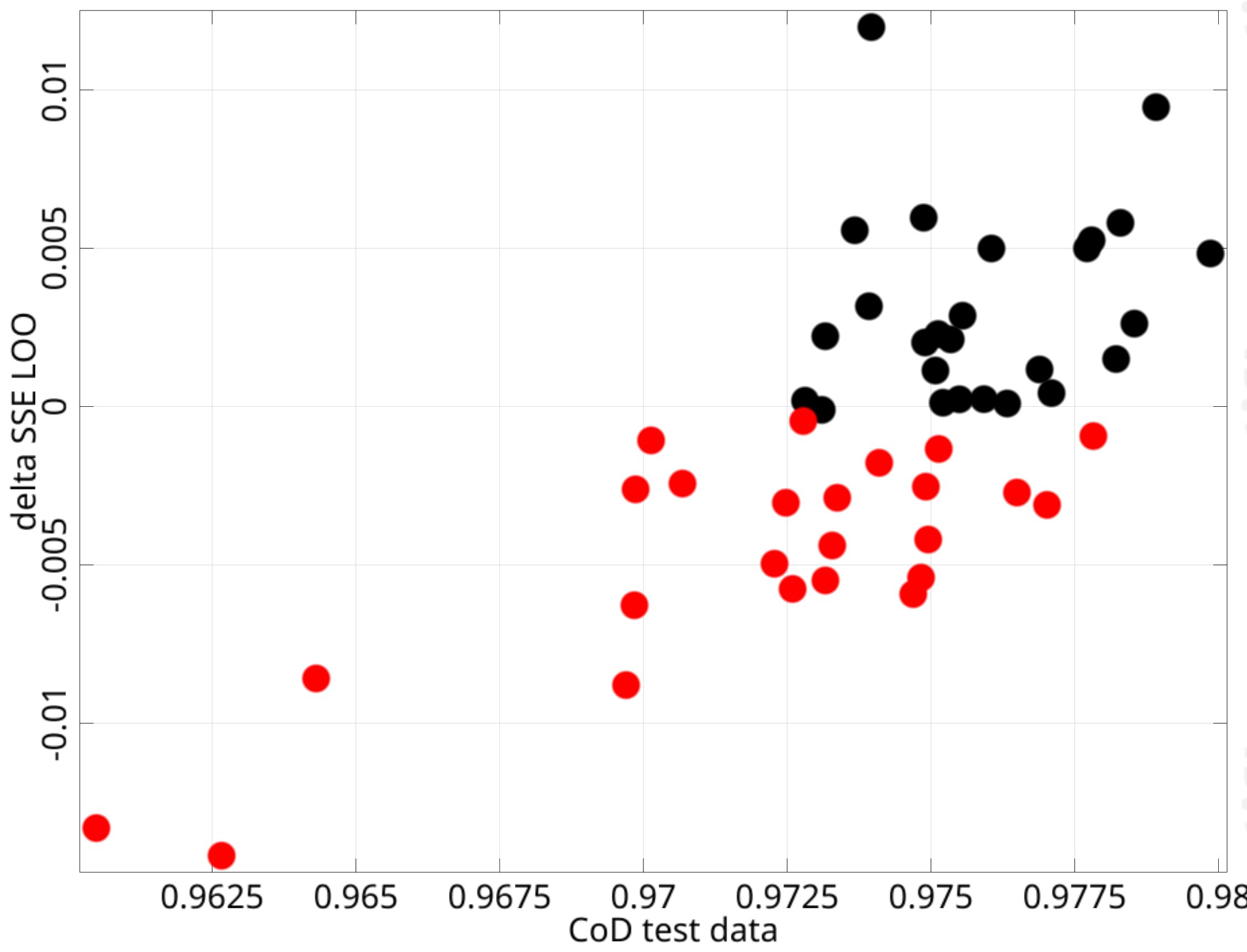}
 	\caption{Statistical evaluation of the prediction errors of the analytical test function by using 200 support points and 500 test points with k-fold-cross-validation (left) and  LOO-cross-validation (right)}
	\label{coupled_function_200}
\end{figure}

In order to quantify the statistical scatter of the prediction error estimates, we generate 50 independent data sets for the support points and perform the model training and error estimation and compare these estimates with the prediction error of a fixed test data set. Since the $SS_T$ itself varies for each support point set, we compare not directly the estimated CoP from the cross-validation with the CoD of the test data. Instead we scale the $SS_E^{cv}$ from the cross-validation with the $SS_T$ of the test data as follows
\begin{equation}
\Delta SS_E^{cv}=\frac{\frac{1}{n} SS_E^{cv} - \frac{1}{n_t} SS_E^{test}}{\frac{1}{n_t} SS_T^{test}},
\end{equation}
where the normalization with the number of supports $n$ and the number of test data points $n_t$ is necessary due to the different number of samples in both sets. In figure \ref{coupled_function_50} and \ref{coupled_function_200} the obtained $\Delta SS_E^{cv}$ are shown for the 50 investigated runs by using LOO as well as k-fold-cross-validation in the prediction quality estimation. The figure indicates, that in case of LOO the number of runs, where the $SS_E$ is over-estimated, is similar as the number of cases where the $SS_E$ is under-estimated. If the k-fold-cross-validation is used with 50 support points, the estimated $SS_E$ is mostly larger as verified by the test data, which is indicated by $\Delta SS_E^{cv} > 0$. This means that the CoP estimate is in the most cases more conservative and does not over-estimate the prediction quality of the investigated surrogate model.
If the number of support points is increased, the deviation between the LOO and k-fold-cross-validation quality estimates reduces.

\begin{figure}[th]
\center
	\includegraphics[width=0.8\textwidth]{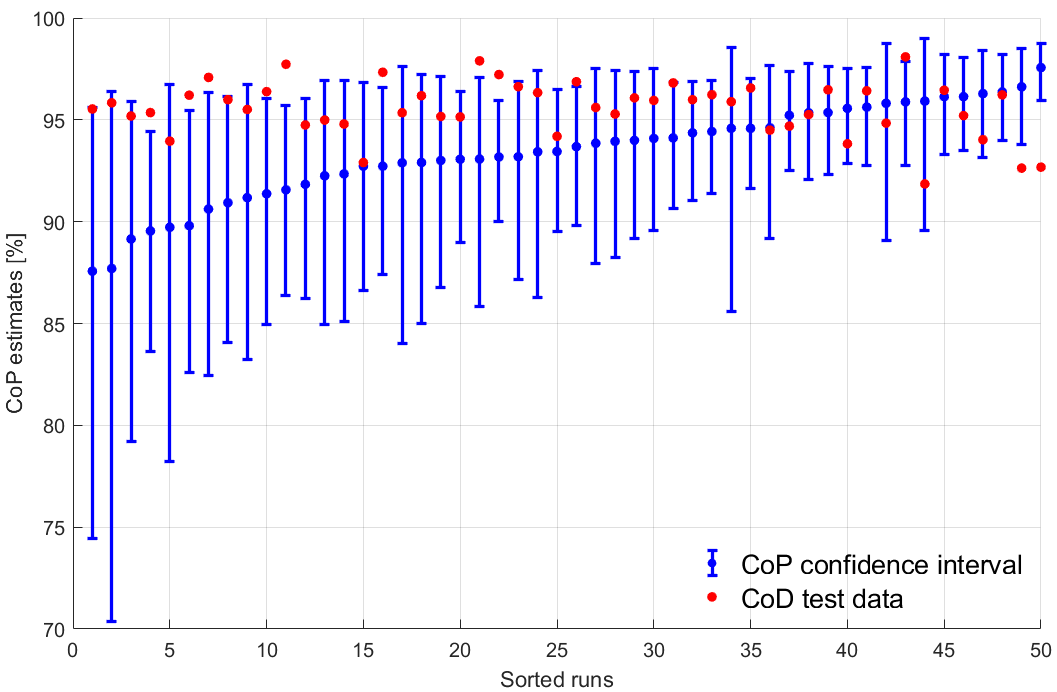}\\
	\includegraphics[width=0.8\textwidth]{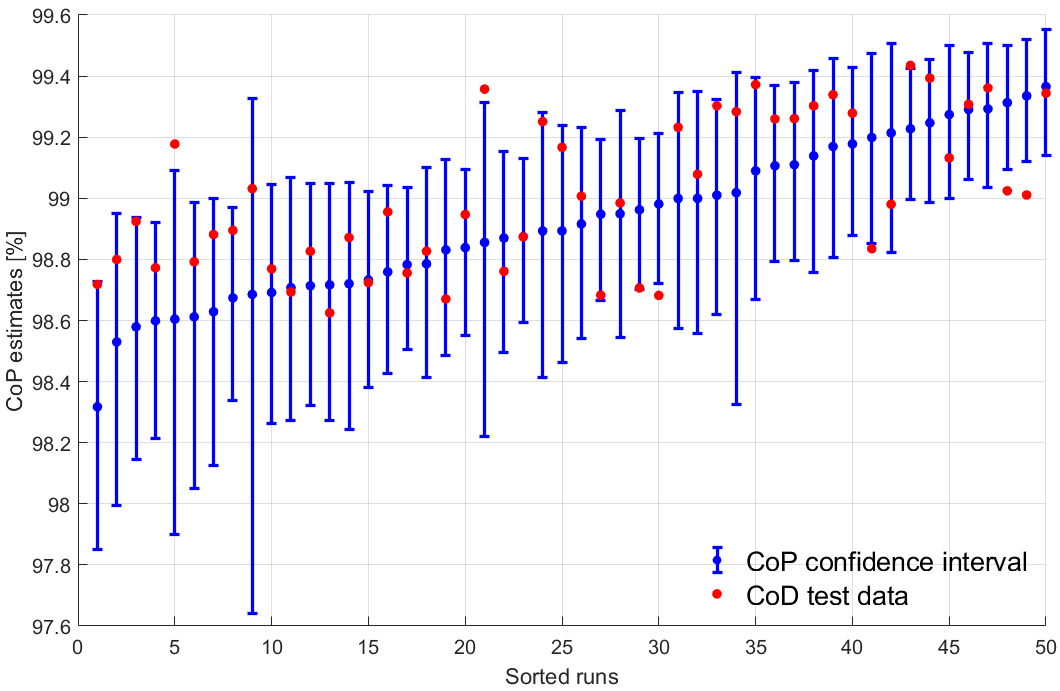}
 	\caption{CoP estimates and confidence bounds of the analytical test function by using k-fold-cross-validation compared to the CoD of the test data for 50 support points (top) and 200 support points (bottom)}
	\label{coupled_function_boot}
\end{figure}

\begin{figure}[th]
\center
	\includegraphics[width=0.49\textwidth]{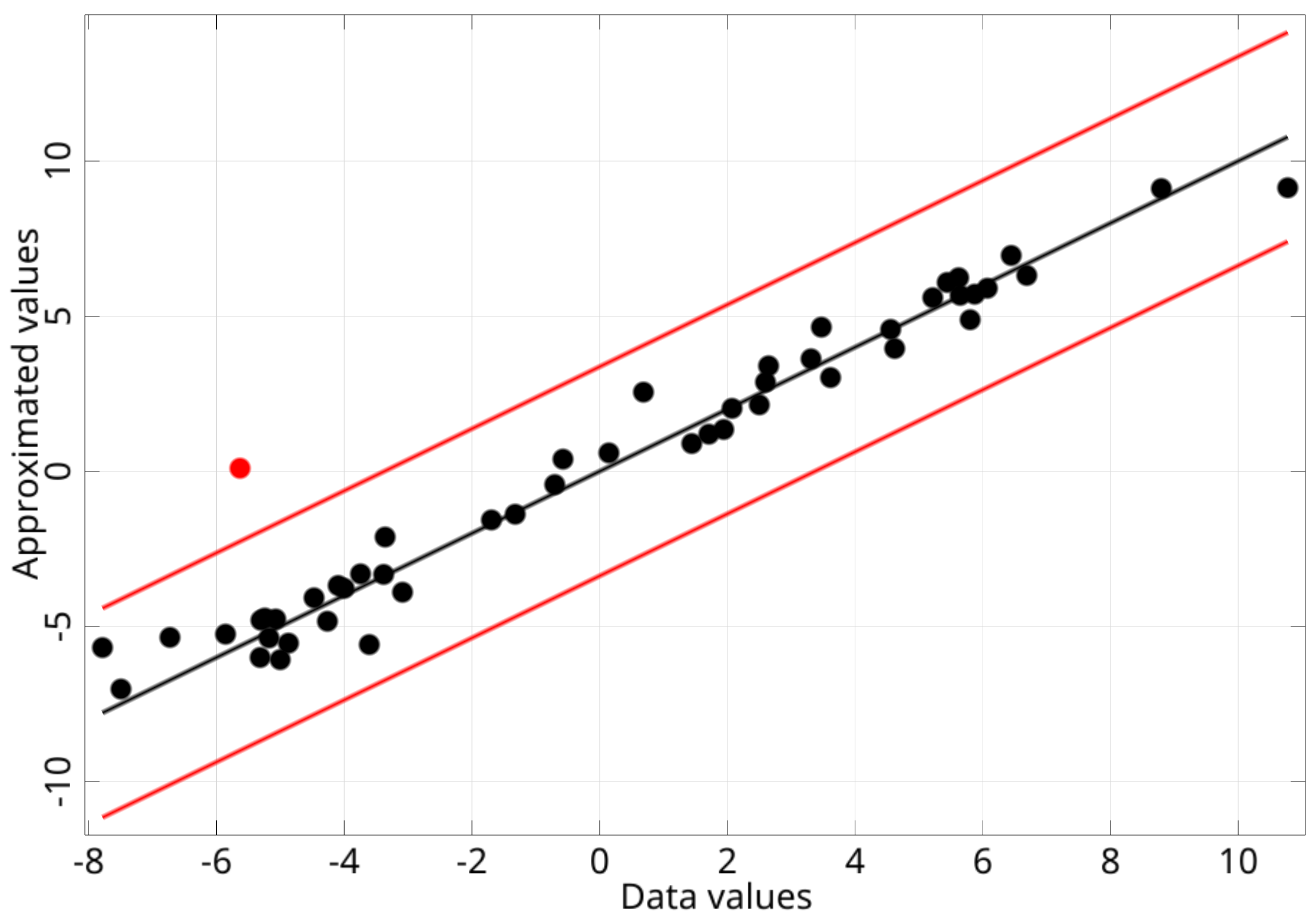}
	\hfill
	\includegraphics[width=0.49\textwidth]{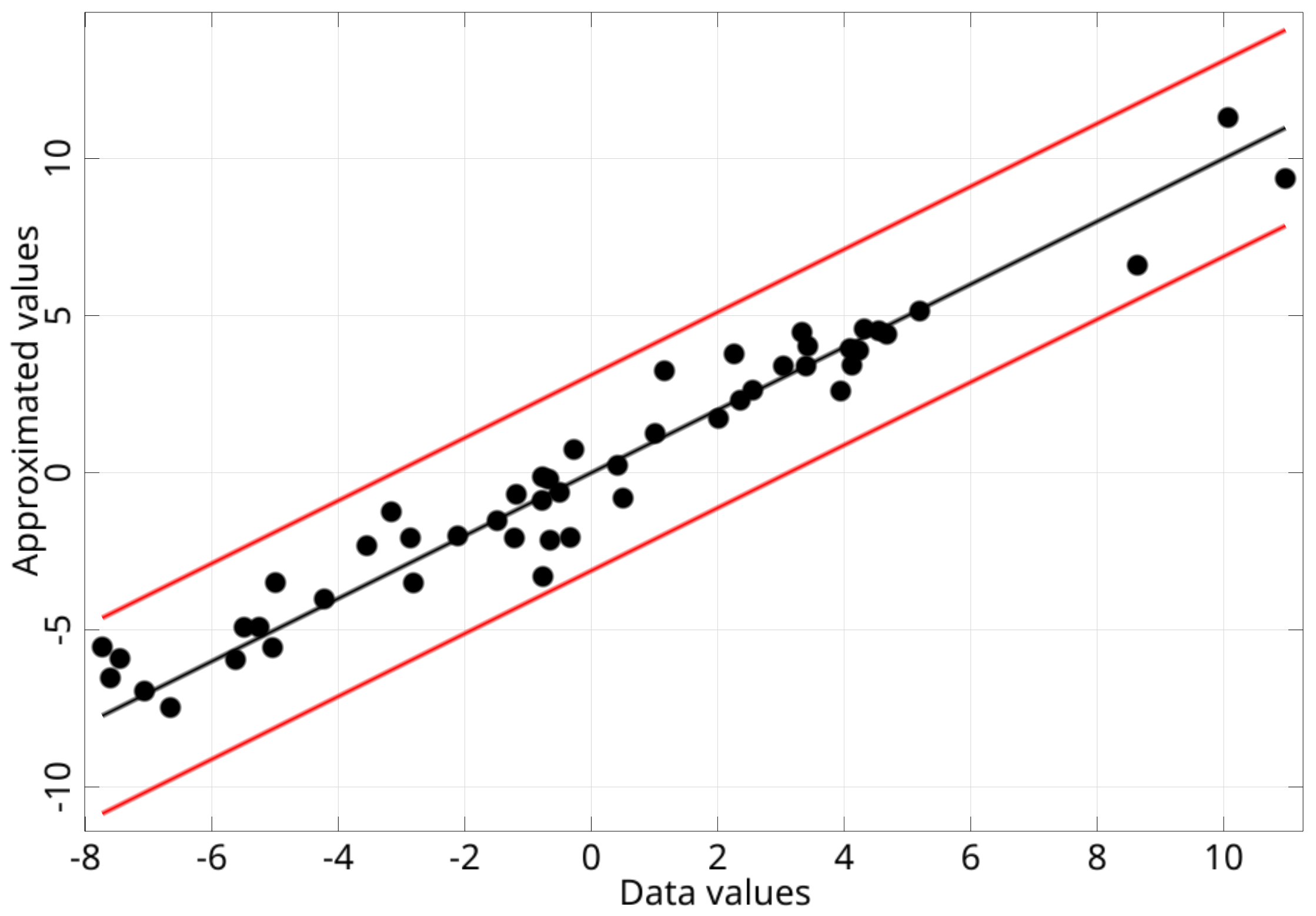}\\
	\includegraphics[width=0.49\textwidth]{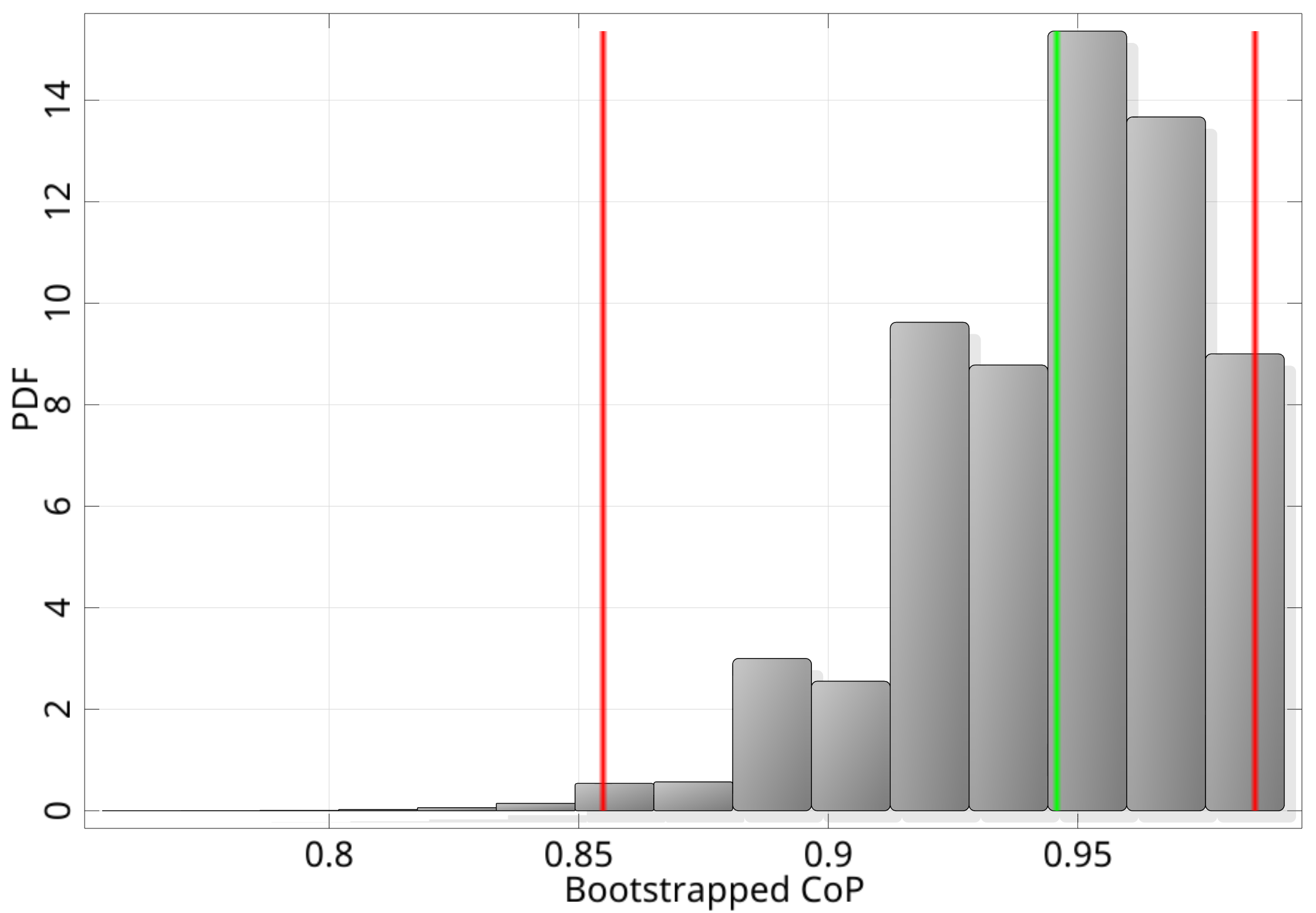}
	\hfill
	\includegraphics[width=0.49\textwidth]{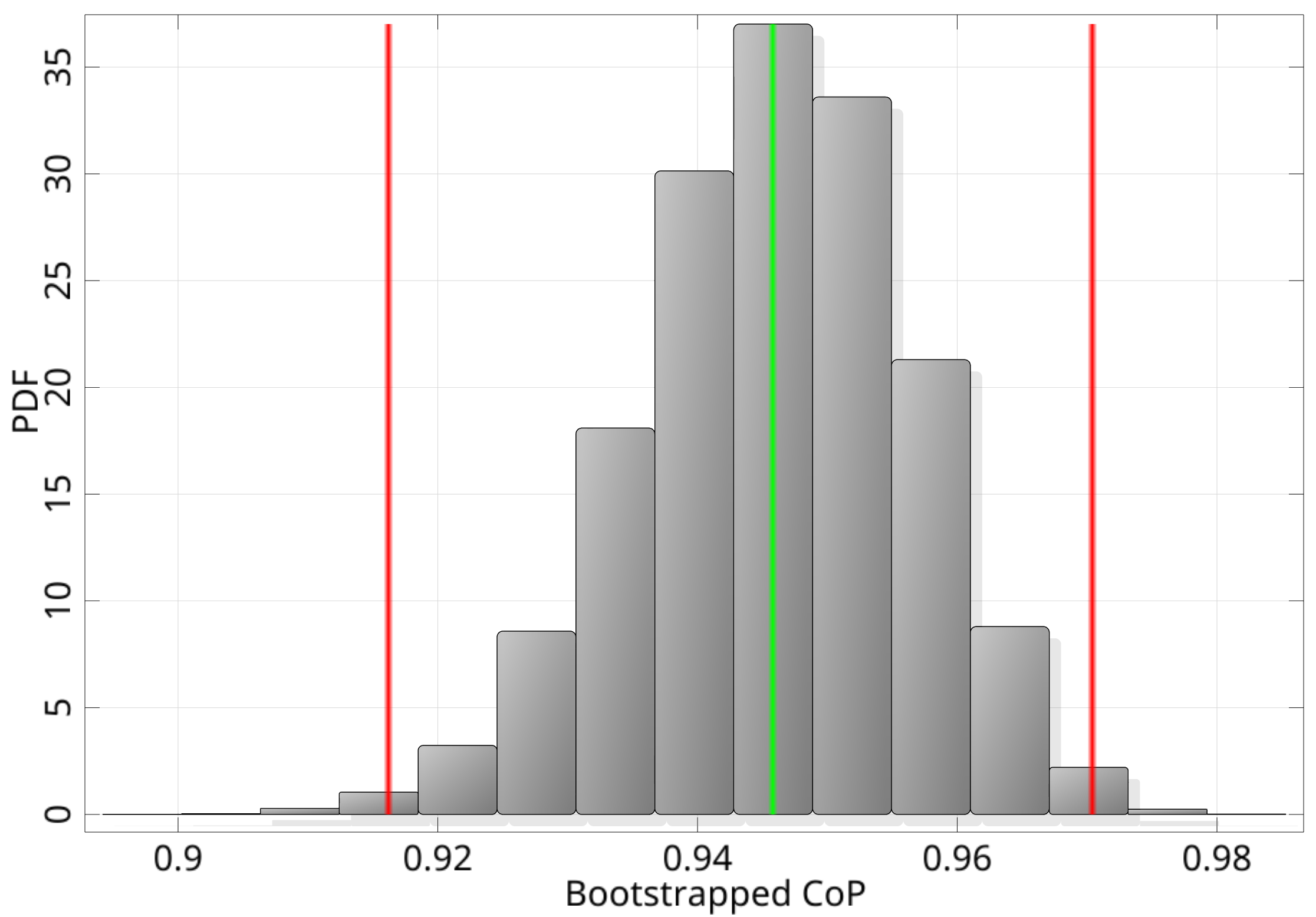}
 	\caption{Residual plots and bootstrapped CoP's of the analytical test function of sorted run number 34 (left) and run number 35 (right) by using 50 support points}
	\label{coupled_residuals}
\end{figure}

Additionally, we investigate the confidence of the estimated CoP compared to the corresponding CoD of the test data. The confidence interval of the CoP is estimated directly from the k-fold cross validation residuals for each run using the bootstrap approach with $10^5$ repetitions. In figure \ref{coupled_function_boot} the CoP estimates with 99\% confidence bounds are shown for the 50 investigated runs.
The figure indicates, that the confidence interval of the CoP covers the verified CoD in almost all cases. If only 50 support points are used, the CoP estimate is generally more conservative as for of 200 support points.
If we look deeper into the results, we can observe, that for several runs the CoP estimate is similar but the confidence bounds differ significantly. For the 50 support points this is the case for the sorted run numbers 34 and 35. In figure \ref{coupled_residuals} the residual plots and the histogram of the bootstrapped CoP of both cases are shown. The figure indicates, that for run 34 with the larger confidence interval, one significant outlier can be observed while the remaining residuals are smaller. In run number 35 the residuals indicate no significant outlier but have larger variation as in run number 34. This means, that in case of possible outliers the confidence interval of the CoP should be larger and a narrow estimate of the CoP is not possible.

\subsection{Noisy benchmark function} 
In the second example we extend the analytical function with additional linear, non-linear and noise terms. The function for 20 inputs reads
\begin{equation}
\begin{aligned}
y(\mathbf{x})=&0.5\cdot x_1+x_2+0.5\cdot x_1 x_2 + 5.0\cdot \sin(x_3) +0.5\cdot x_4 +0.5\cdot x_4^2+0.1\cdot x_5\\ 
&+ \sum_{i=6}^{20} 0.01\cdot x_i +0.5\cdot \mathcal{N}(0,1), \quad -\pi \leq x_i \leq \pi,
\end{aligned}
\label{noisy}
\end{equation}
where $\mathcal{N}(0,1)$ is a standard normal noise term. 

\begin{figure}[th]
\center
	\includegraphics[width=0.49\textwidth]{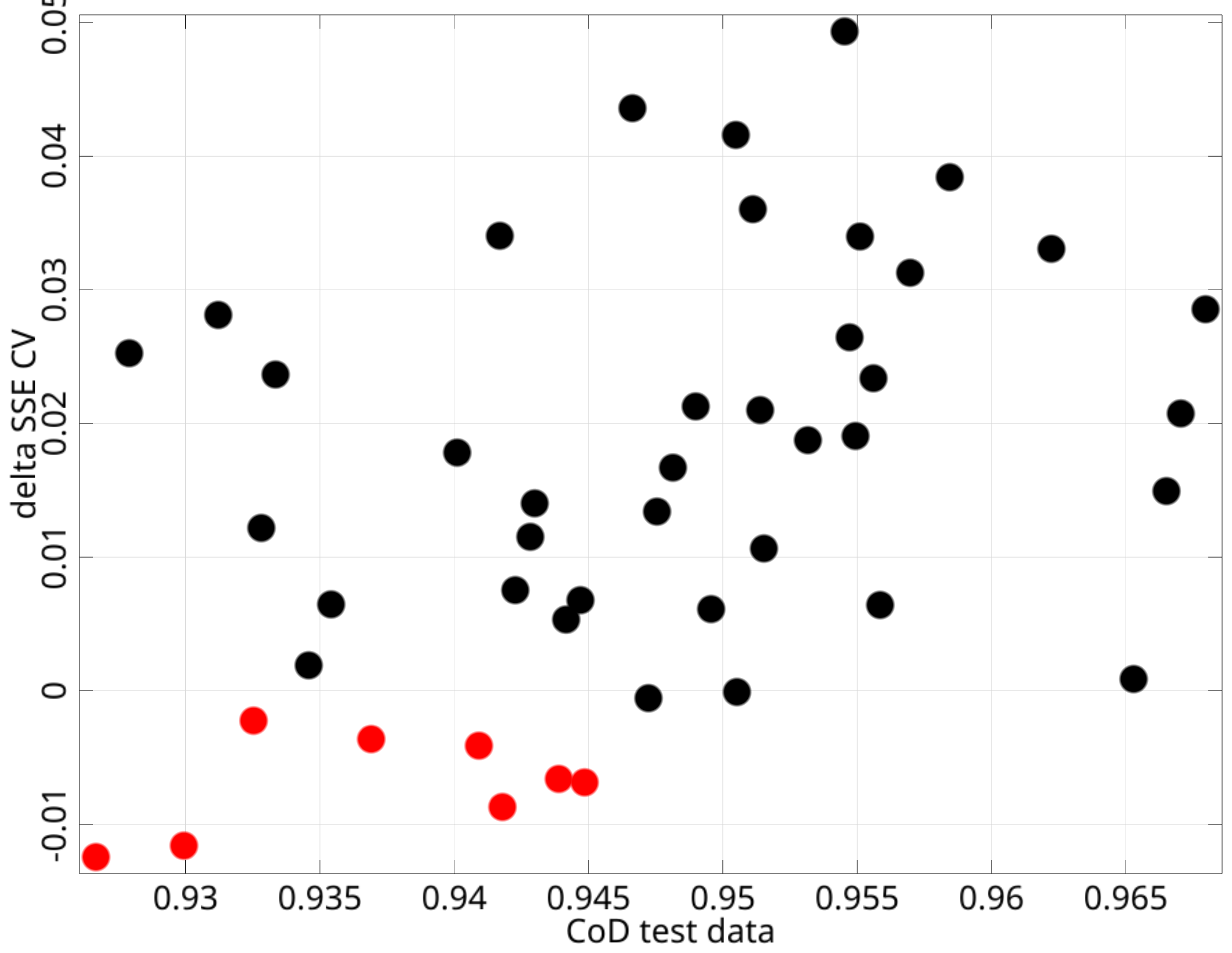}
	\hfill
	\includegraphics[width=0.49\textwidth]{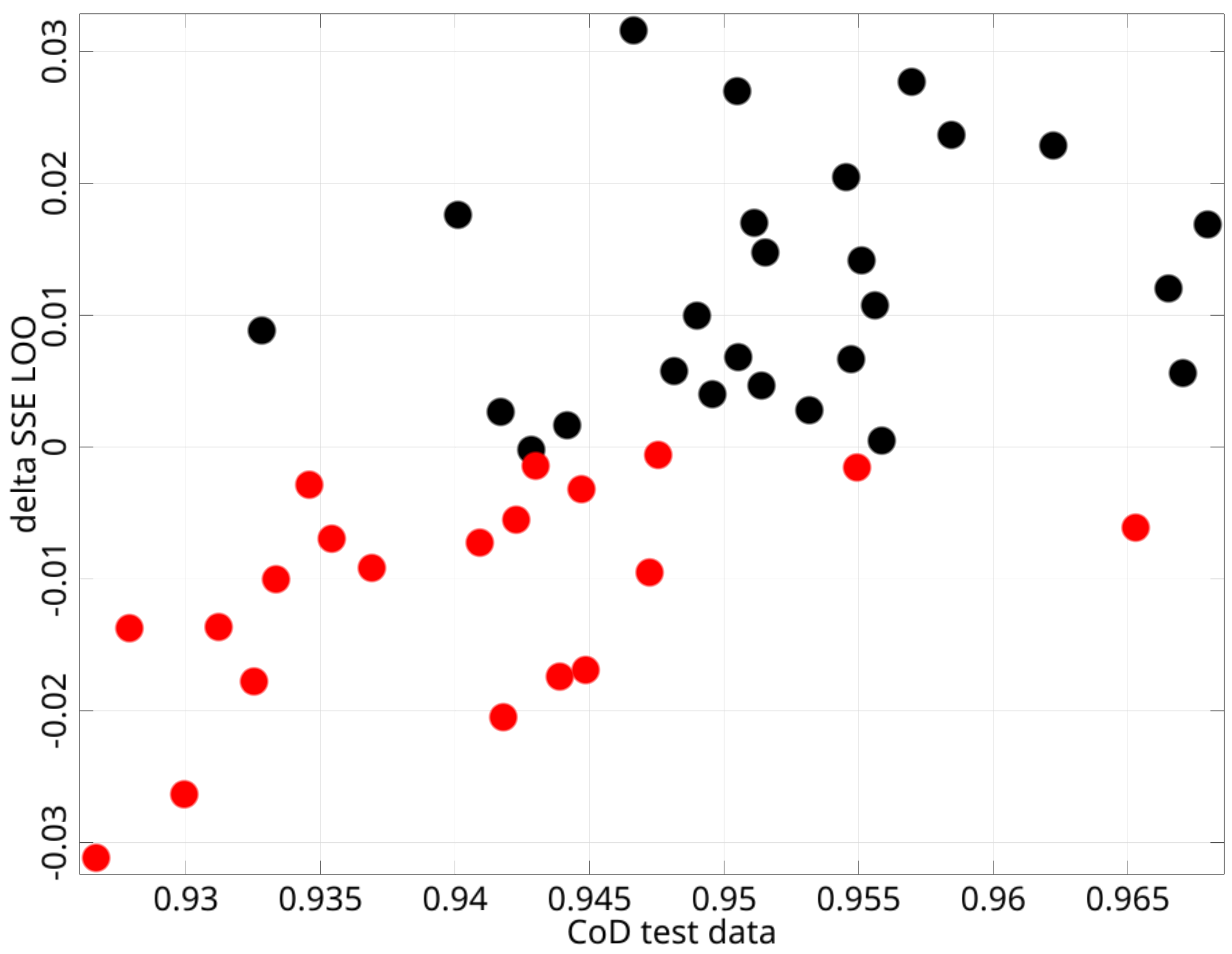}\\
	\hfill
 	\caption{Statistical evaluation of the prediction errors of the noisy test function by using 100 support points and 500 test points with k-fold-cross-validation (left) and  LOO-cross-validation (right)}
	\label{noisy_function_100}
\end{figure}

\begin{figure}[th]
\center
	\includegraphics[width=0.8\textwidth]{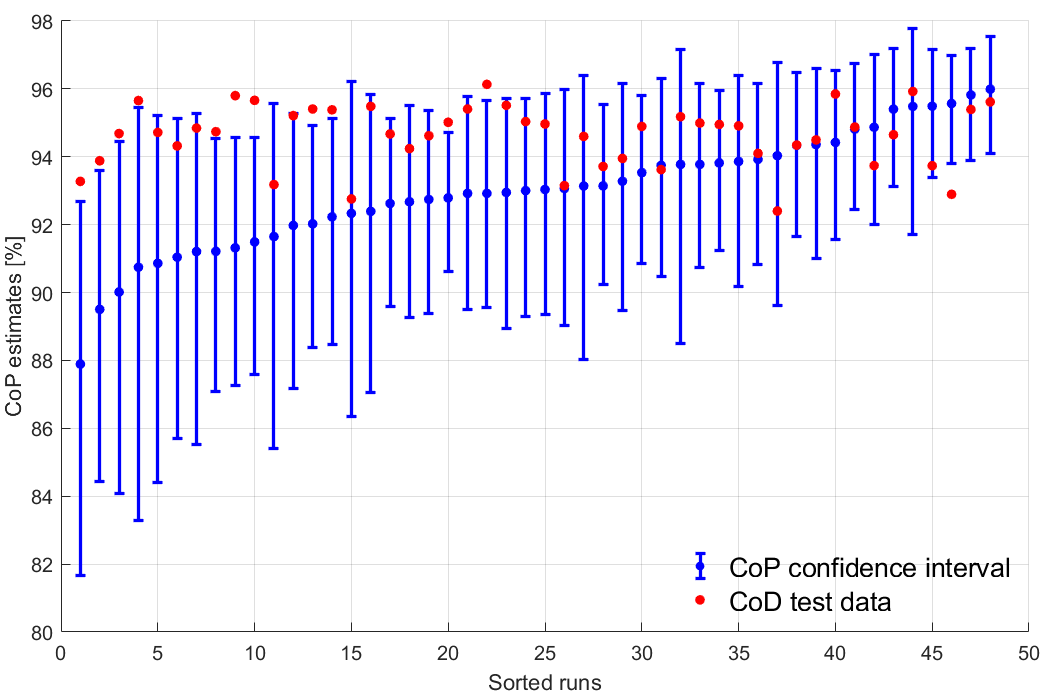}
 	\caption{CoP estimates and confidence bounds of the noisy test function by using k-fold-cross-validation compared to the CoD of the test data for 100 support points}
	\label{noisy_function_boot}
\end{figure}

We generate again 100 support points and 500 additional test samples by Latin-Hypercube Sampling (LHS) 
and apply an isotropic Kriging approximation model. The scatter of the statistical measures is analyzed again by evaluating 50 runs 
with LOO and k-fold-cross-validation.
In figure \ref{noisy_function_100} the calculated errors $\Delta SS_E^{cv}$ are compared for both cases. The figure indicates a similar behavior as in the first example, where the k-fold-cross-validation gives more conservative results and thus the CoP does not over-estimate the prediction quality.
In figure \ref{noisy_function_boot} the estimated confidence intervals are compared to the CoD of the additional test data. As in the previous example, the estimated confidence bounds of the CoP and the verified CoD agree very well.

\clearpage 


\subsection{Front crash example}

\begin{figure}[th]
\center
	\includegraphics[width=0.75\textwidth]{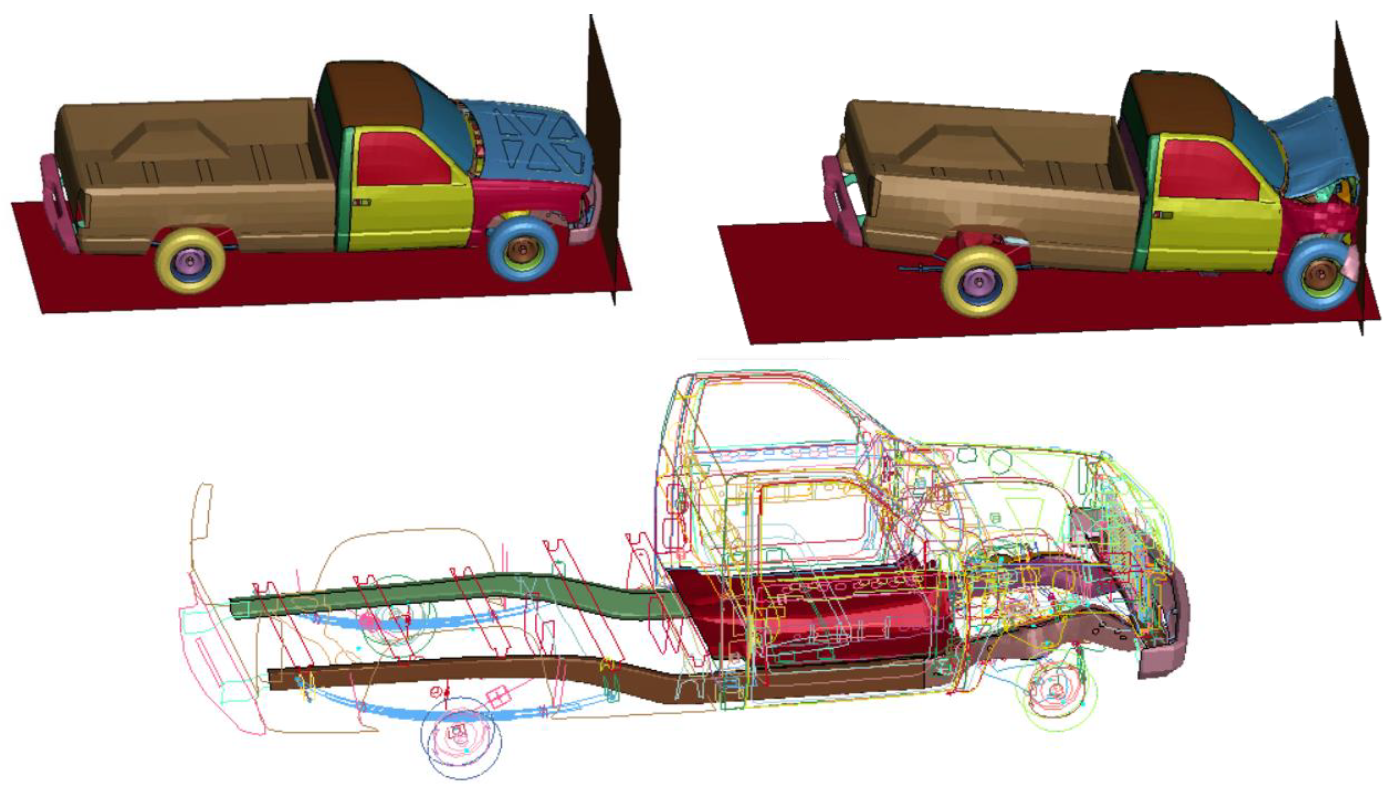}
 	\caption{Investigated front crash example according to \cite{Stander2021} considering 22 varying inputs 
	of specific parts of the car body in a LS-Dyna simulation model}
	\label{crash_01}
\end{figure}

In the third example we investigate the presented error measures on a highly non-linear application, where the intrusions and pulses of a truck impact example are analyzed with the LS-Dyna finite element solver as shown in figure \ref{crash_01}. The pulses are acceleration related quantities computed over two-time-intervals of the crash event. 22 input variables have been considered in the analysis which belong to the metal sheet thicknesses and the material properties of specific parts of the car body. Further details on this example can be found in \cite{Stander2021}.
For this example, different data sets of 100, 200 and 400 Latin Hypercube samples have been used for the model training and a single test data set of 1200 samples for the validation of the estimated prediction errors. 
Again we use the Metamodel of Optimal Prognosis \cite{Most_2011_WOST} to select the most suitable approximation model for each response automatically. As approximation models we consider polynomials and Moving Least Squares, each with linear and quadratic basis, as well as isotropic and anisotropic Kriging. Additionally to the best approximation model, the optimal subspace of important inputs is detected by using the maximum Coefficient of Prognosis as selection criterion. 

\begin{figure}[th]
\center
100 support points\\
	\includegraphics[width=0.95\textwidth]{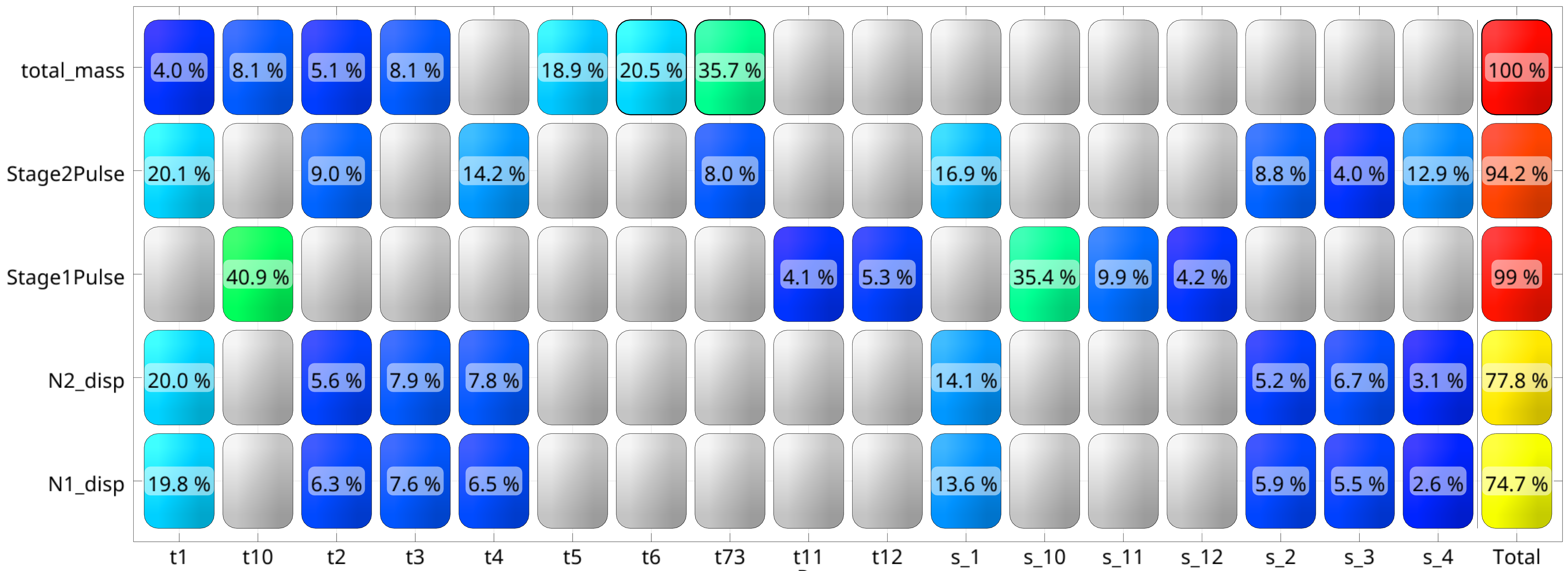}\\
$\quad$\\
200 support points\\
	\includegraphics[width=0.95\textwidth]{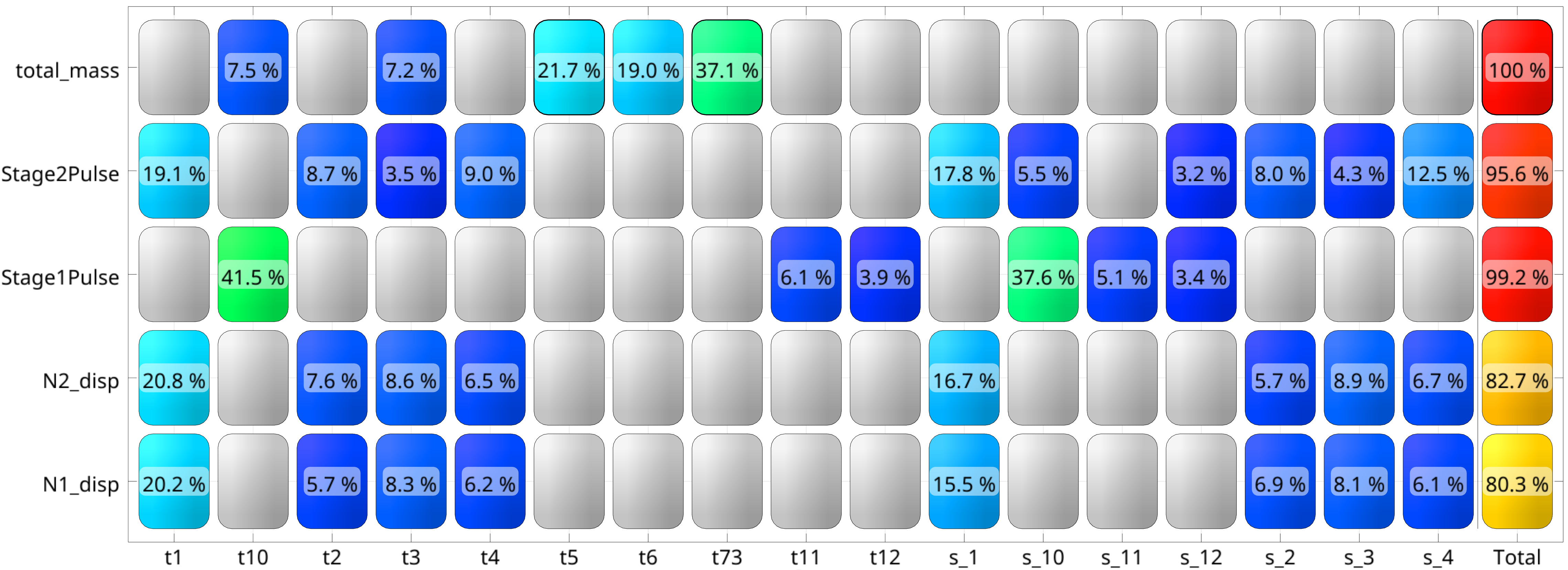}\\
$\quad$\\
400 support points\\
	\includegraphics[width=0.95\textwidth]{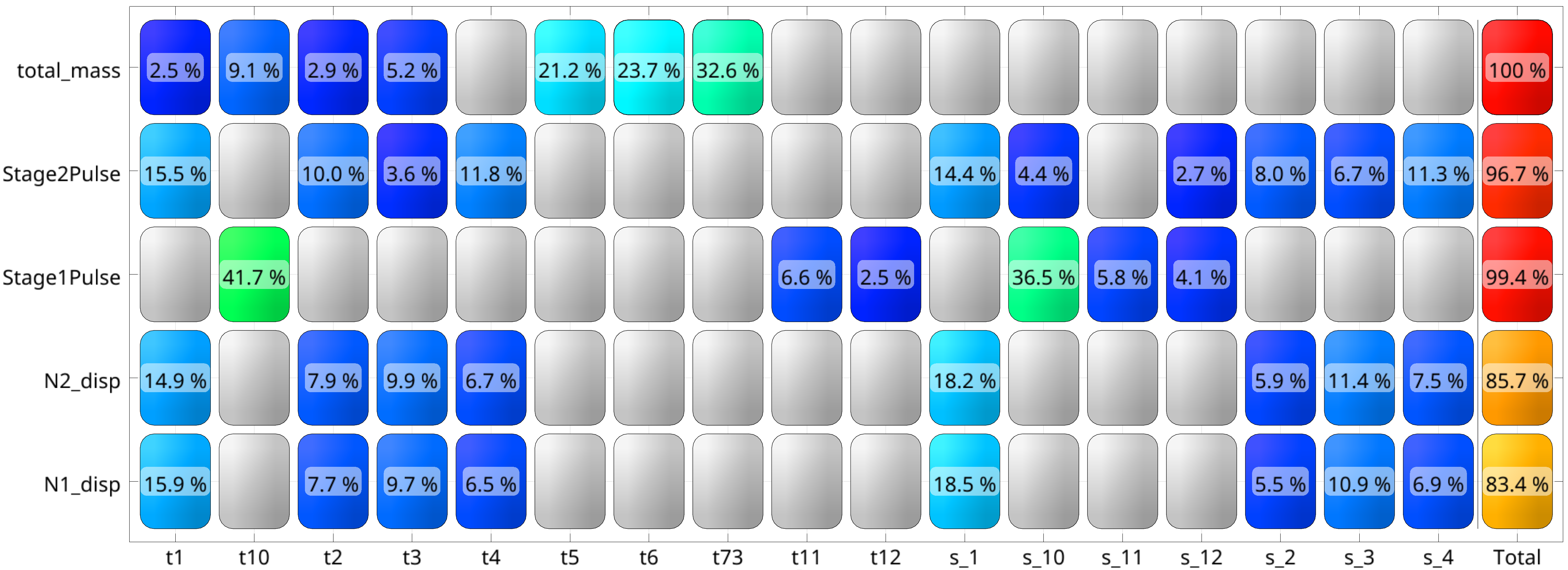}
 	\caption{Estimated CoPs and input sensitivities for the simulation model responses of the truck front crash example by using 100, 200 and 400 support points}
	\label{crash_cop_matrix}
\end{figure}

\begin{figure}[th]
\center
	100 support points \hspace{0.49\textwidth} $\quad$\\
  \includegraphics[width=0.49\textwidth]{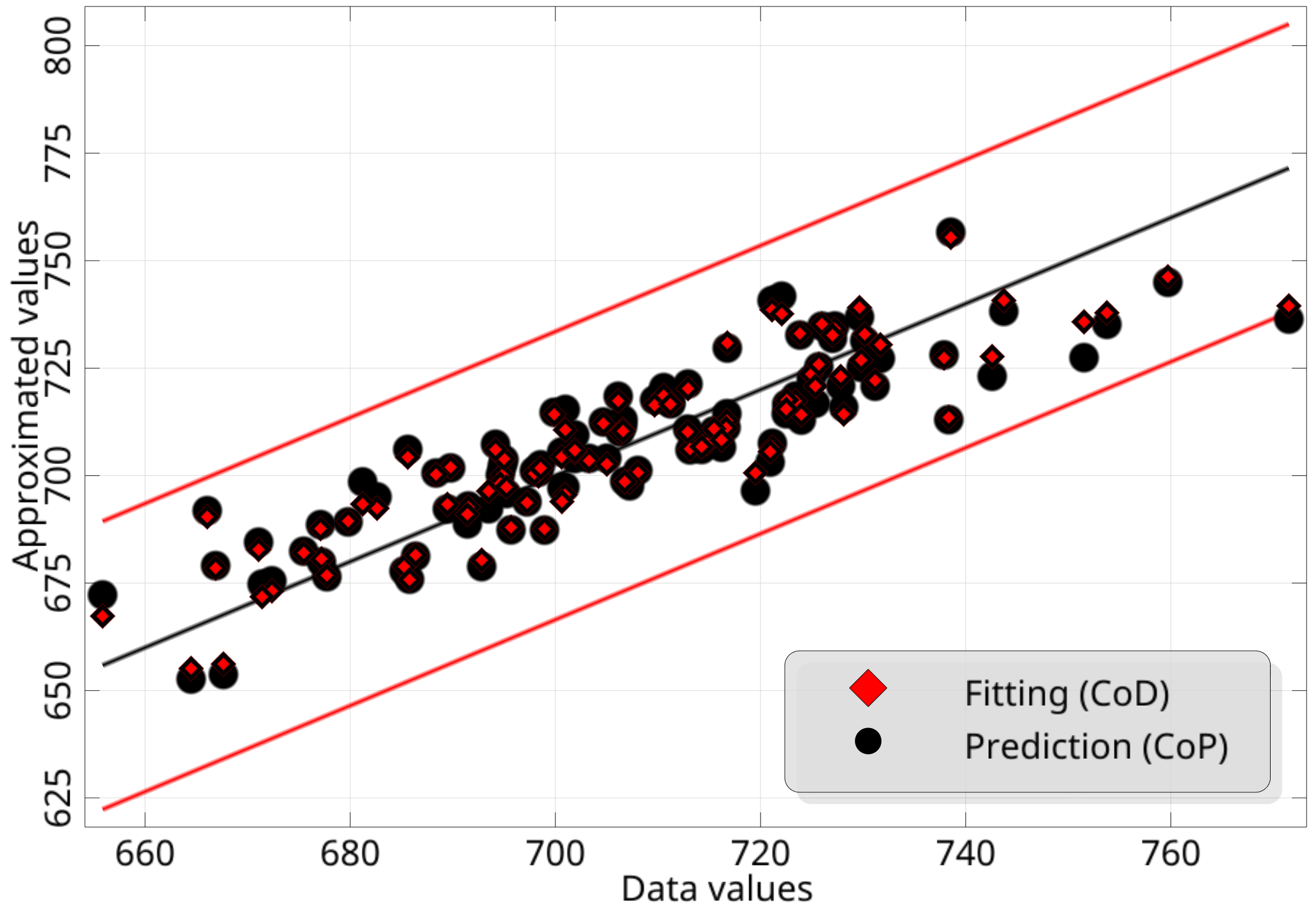}
	\hfill
	\includegraphics[width=0.49\textwidth]{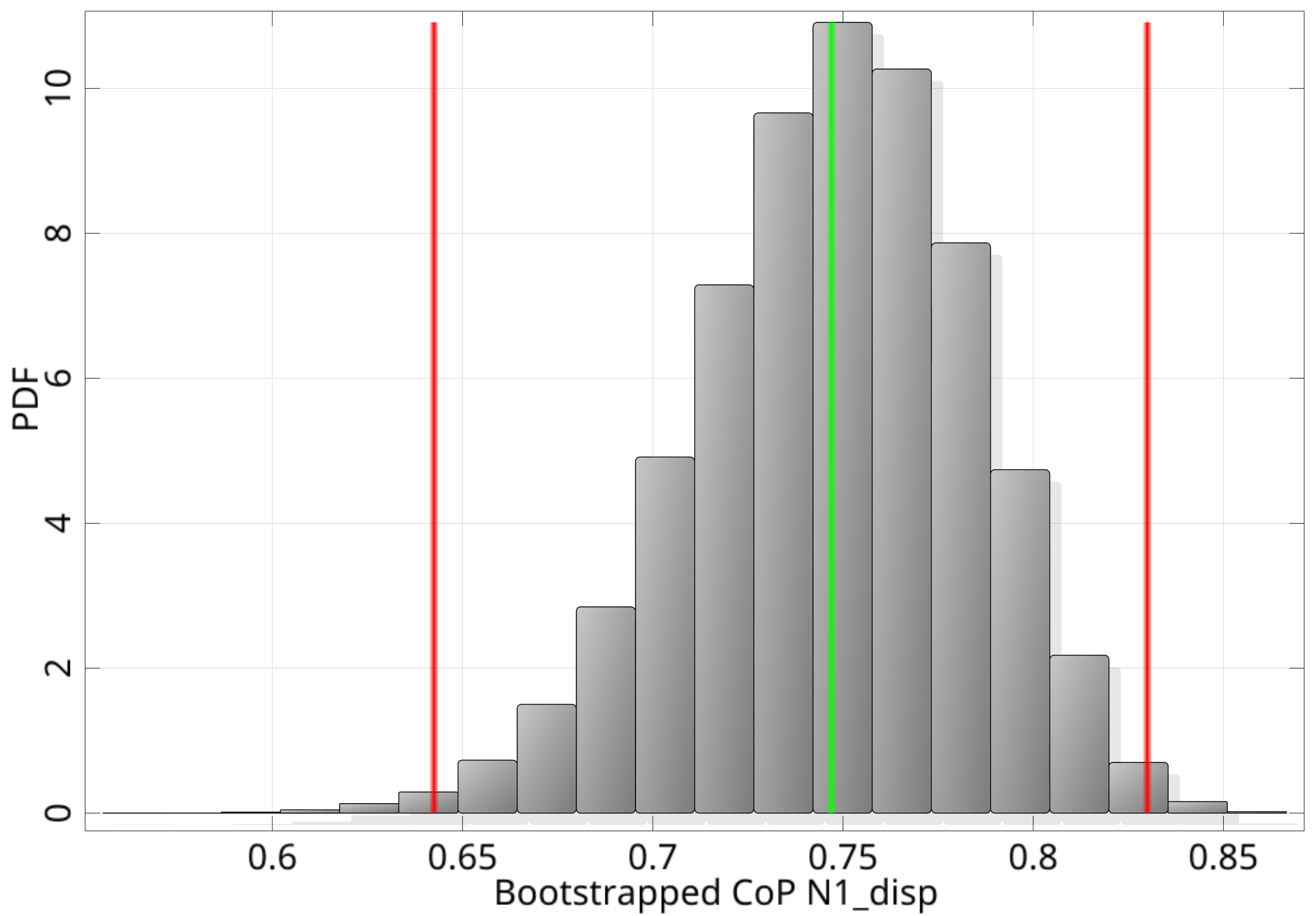}\\
$\quad$\\
200 support points \hspace{0.49\textwidth} $\quad$\\
	\includegraphics[width=0.49\textwidth]{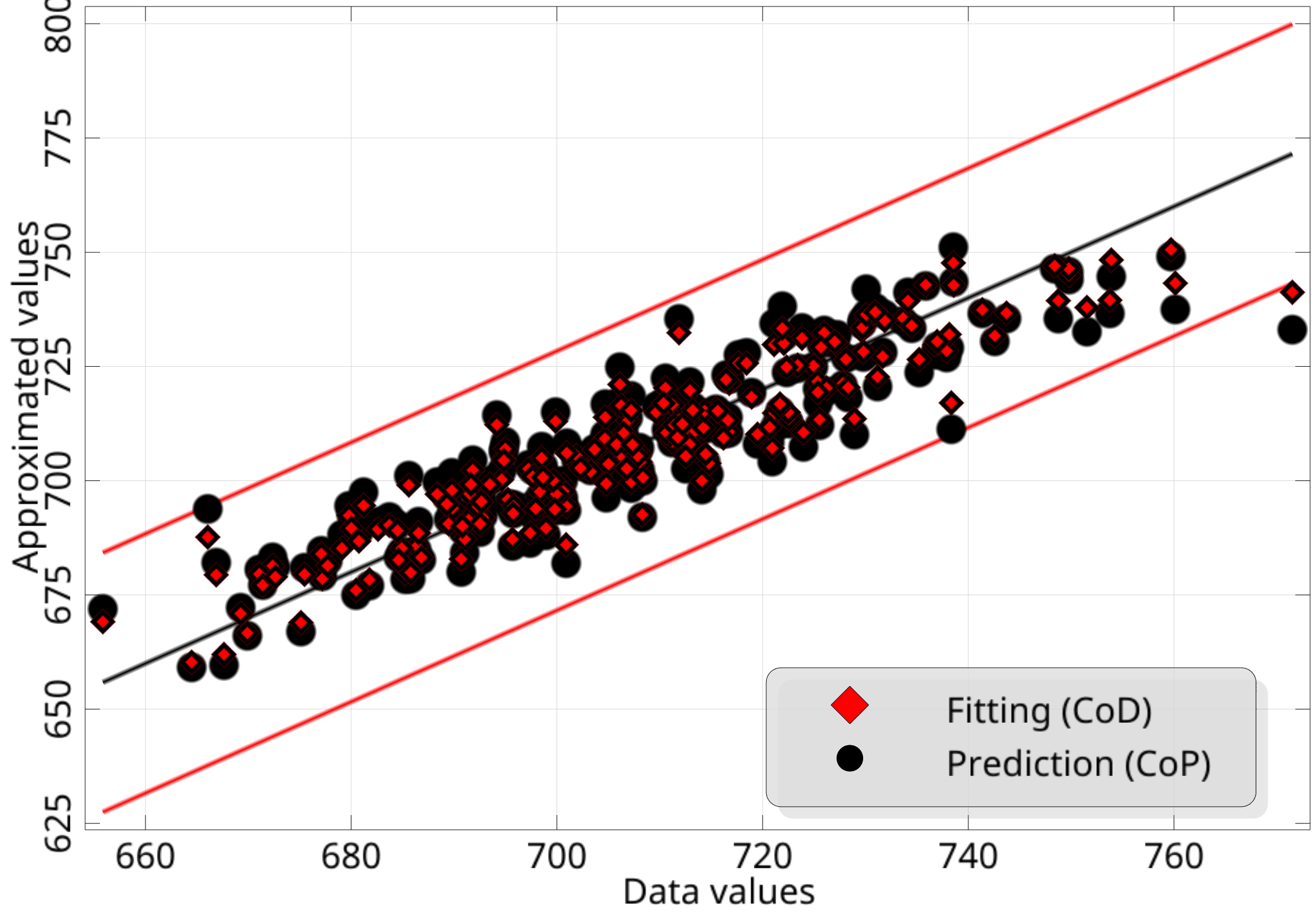}
	\hfill
	\includegraphics[width=0.49\textwidth]{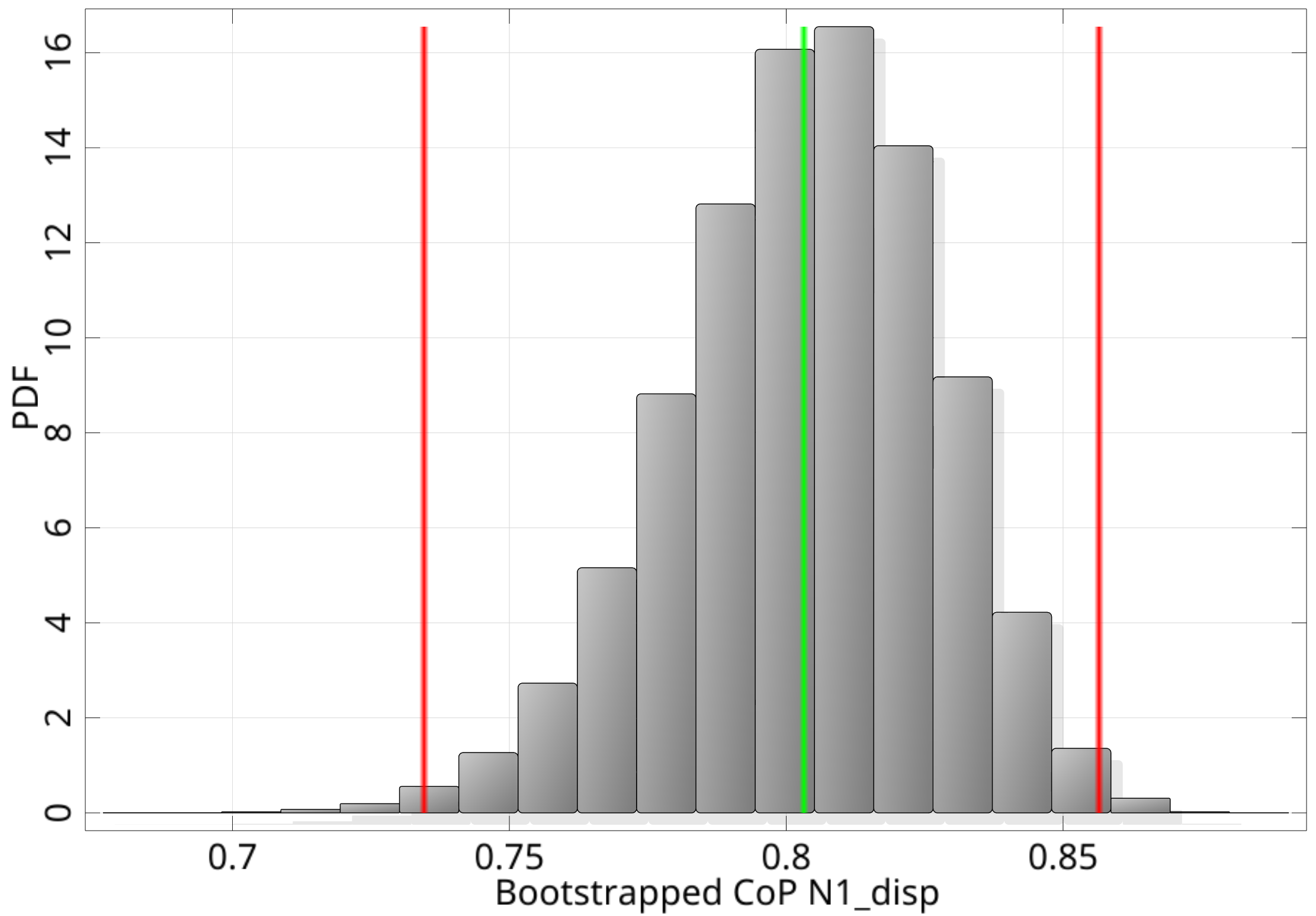}\\
$\quad$\\
400 support points \hspace{0.49\textwidth} $\quad$\\
	\includegraphics[width=0.49\textwidth]{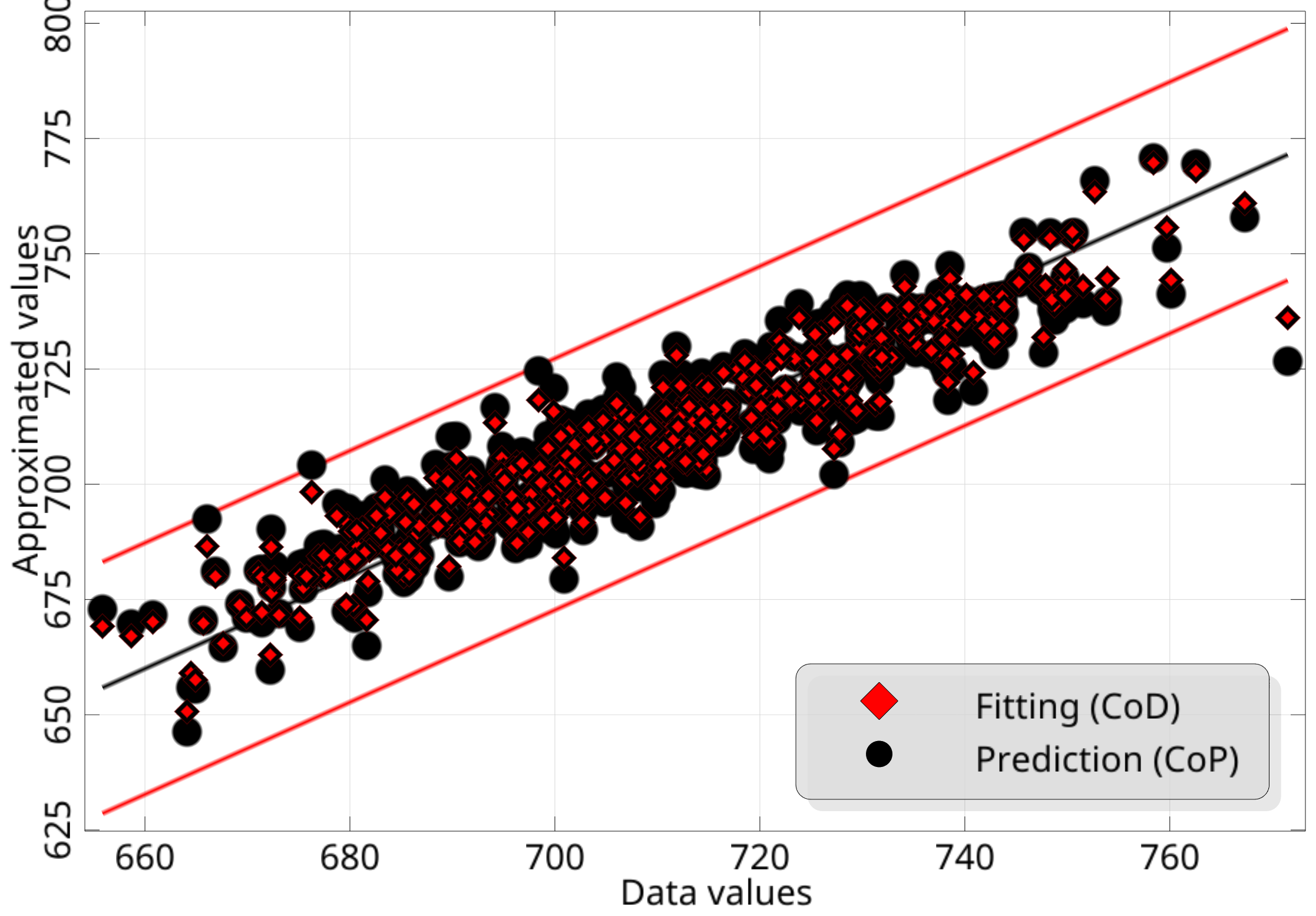}
	\hfill
	\includegraphics[width=0.49\textwidth]{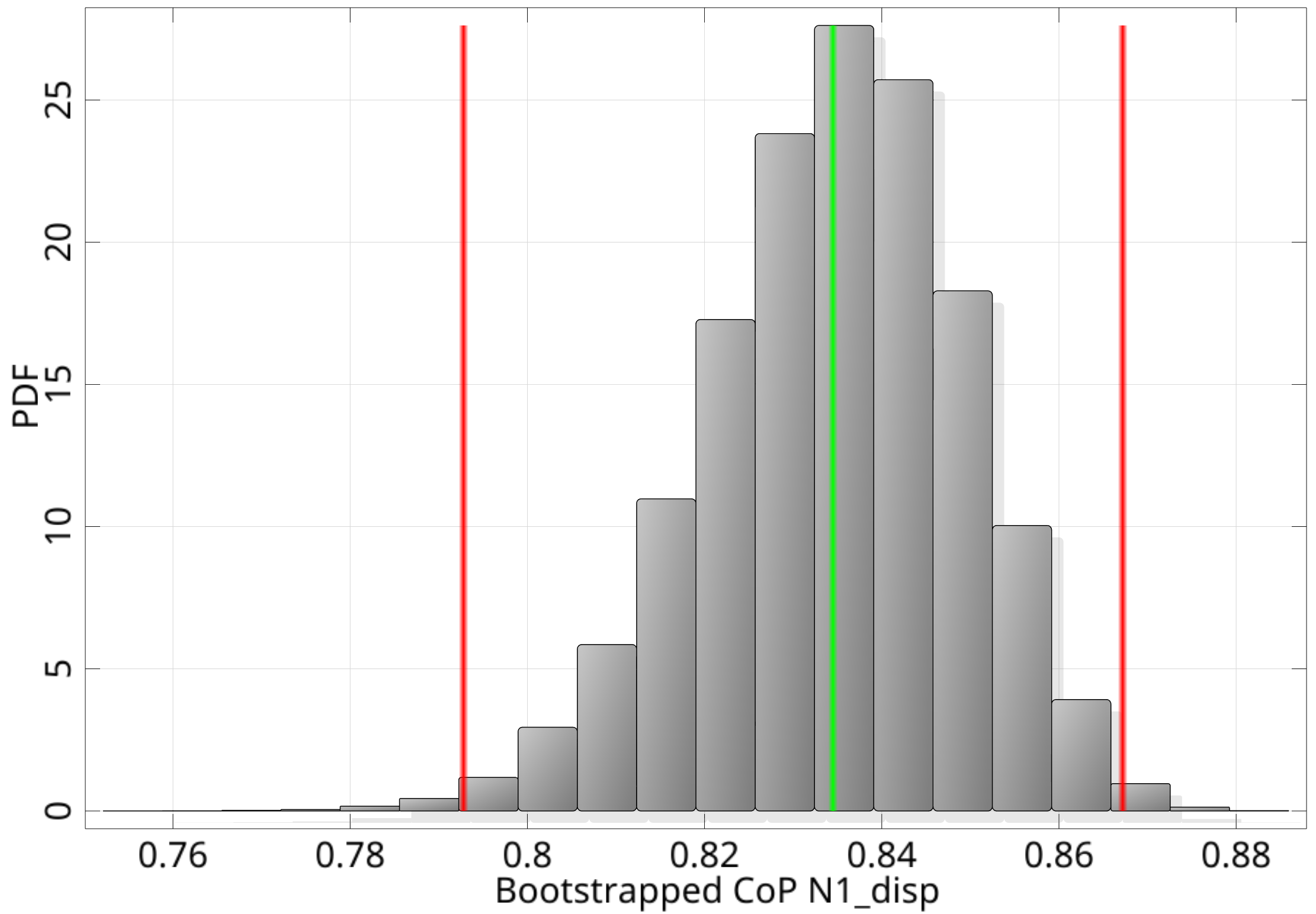}\\
 	\caption{Residual plots (left) and bootstrapped CoP's (right) of the output N1\_disp from the front crash example}
	\label{crash_residuals1}
\end{figure}

\begin{figure}[th]
\center
	100 support points \hspace{0.49\textwidth} $\quad$\\
  \includegraphics[width=0.49\textwidth]{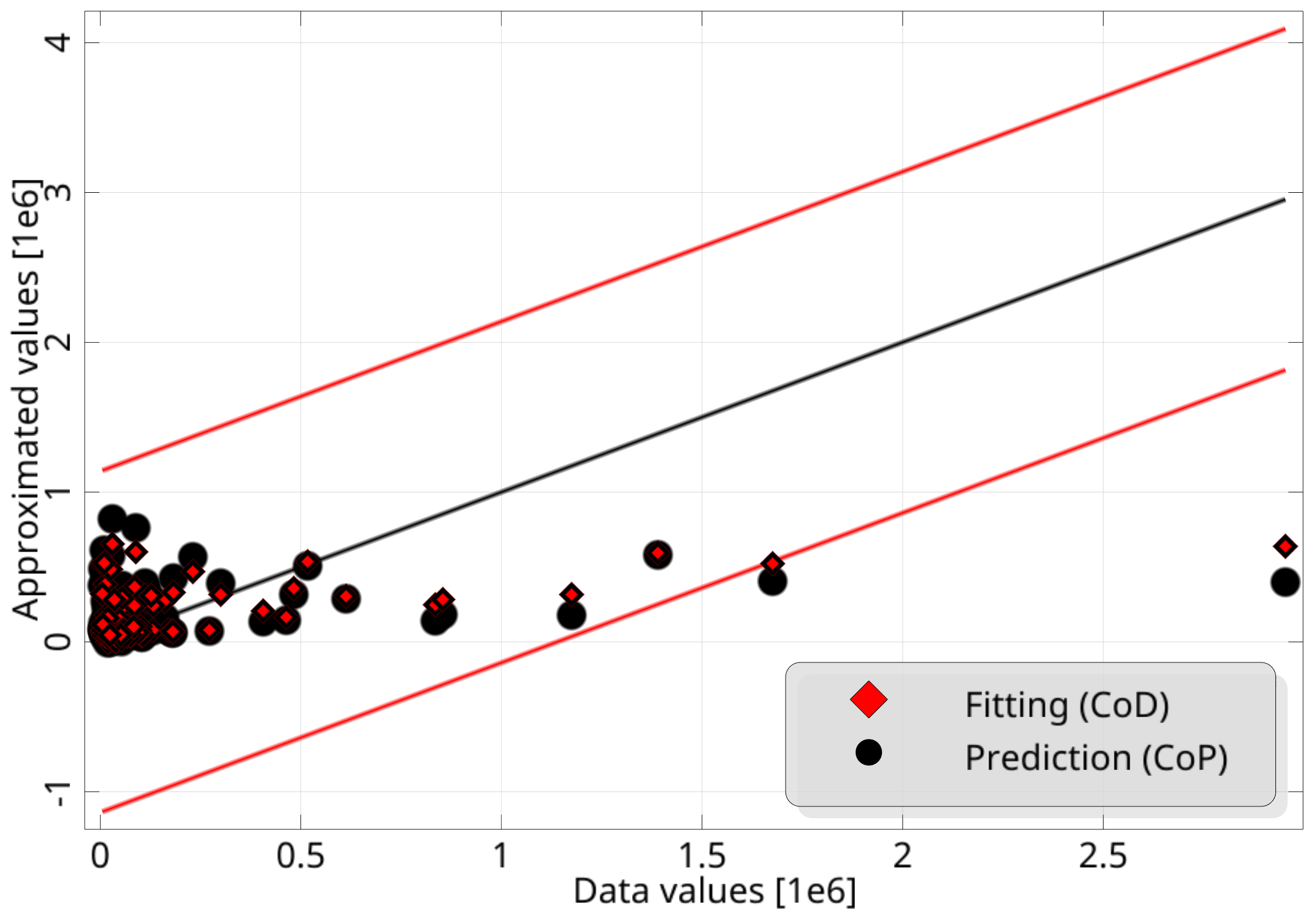}
	\hfill
	\includegraphics[width=0.49\textwidth]{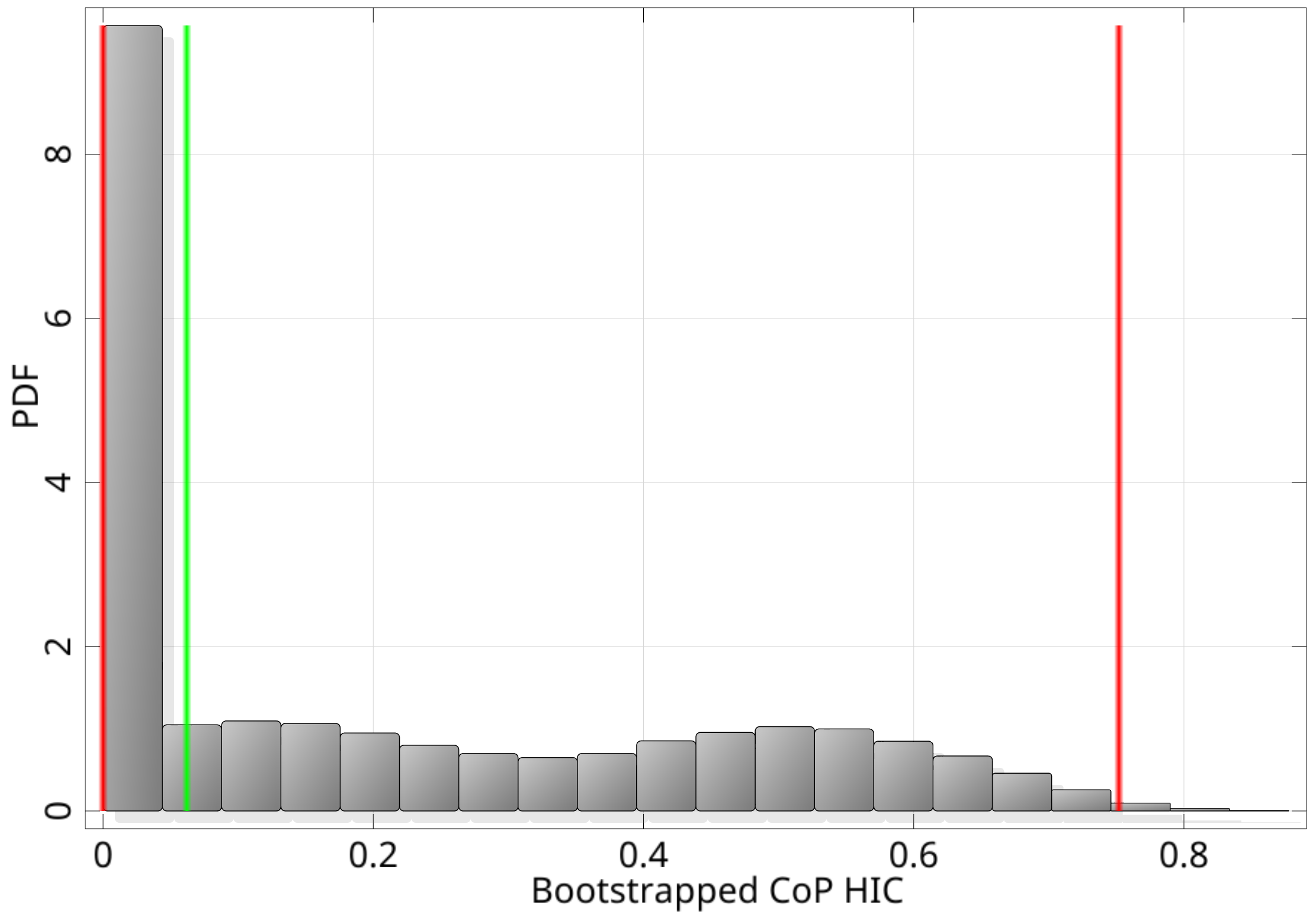}\\
$\quad$\\
200 support points \hspace{0.49\textwidth} $\quad$\\
	\includegraphics[width=0.49\textwidth]{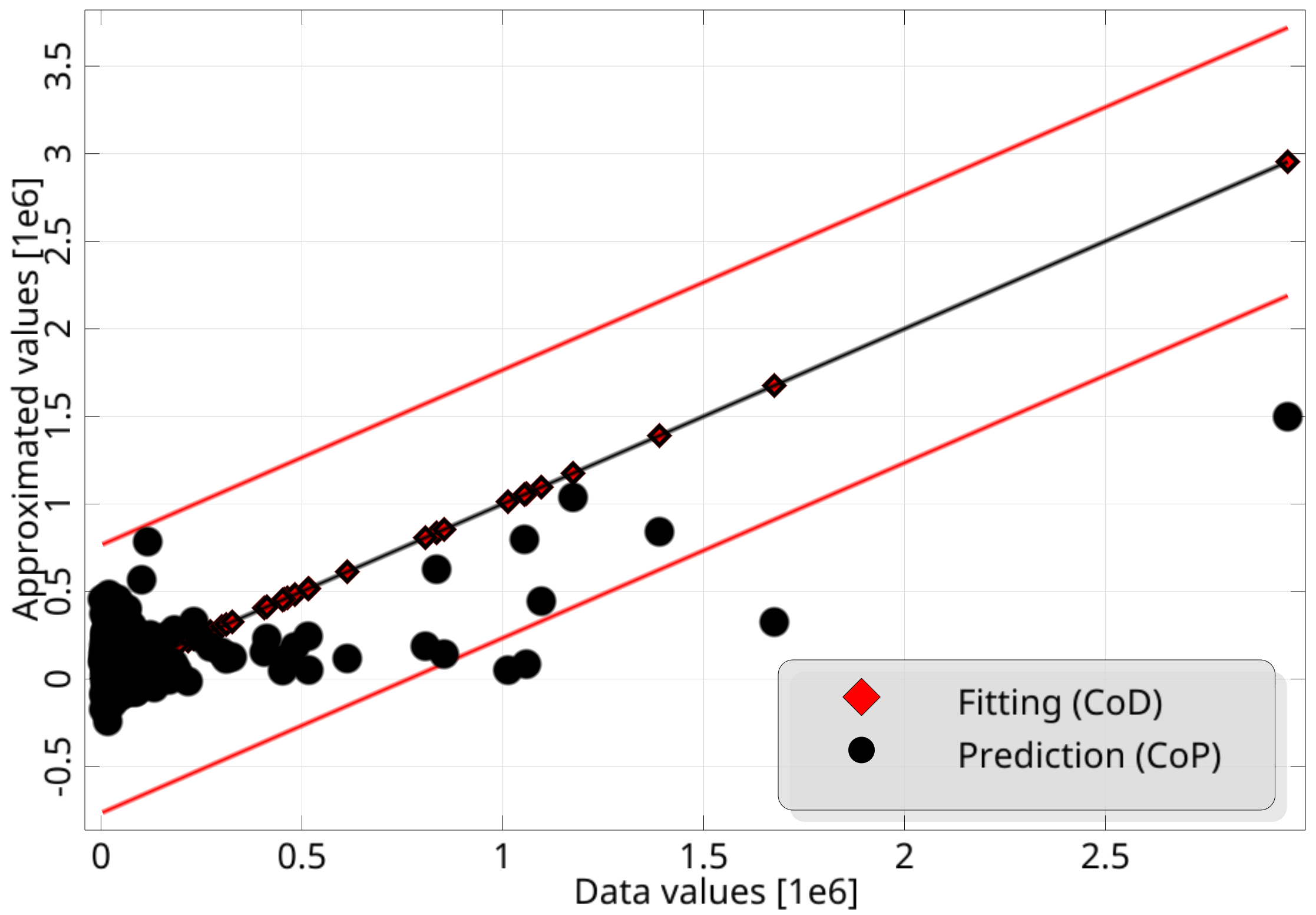}
	\hfill
	\includegraphics[width=0.49\textwidth]{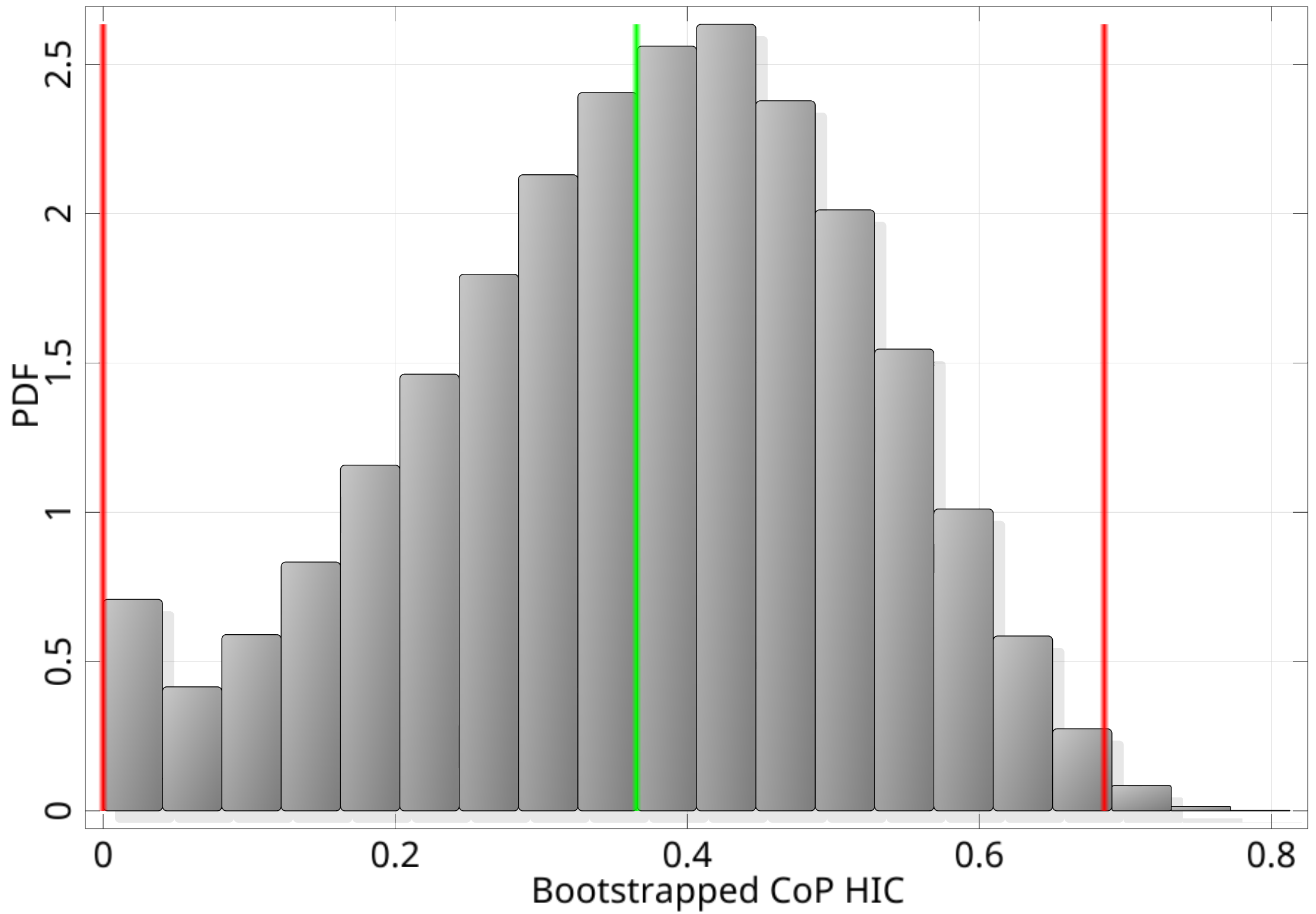}\\
$\quad$\\
400 support points \hspace{0.49\textwidth} $\quad$\\
	\includegraphics[width=0.49\textwidth]{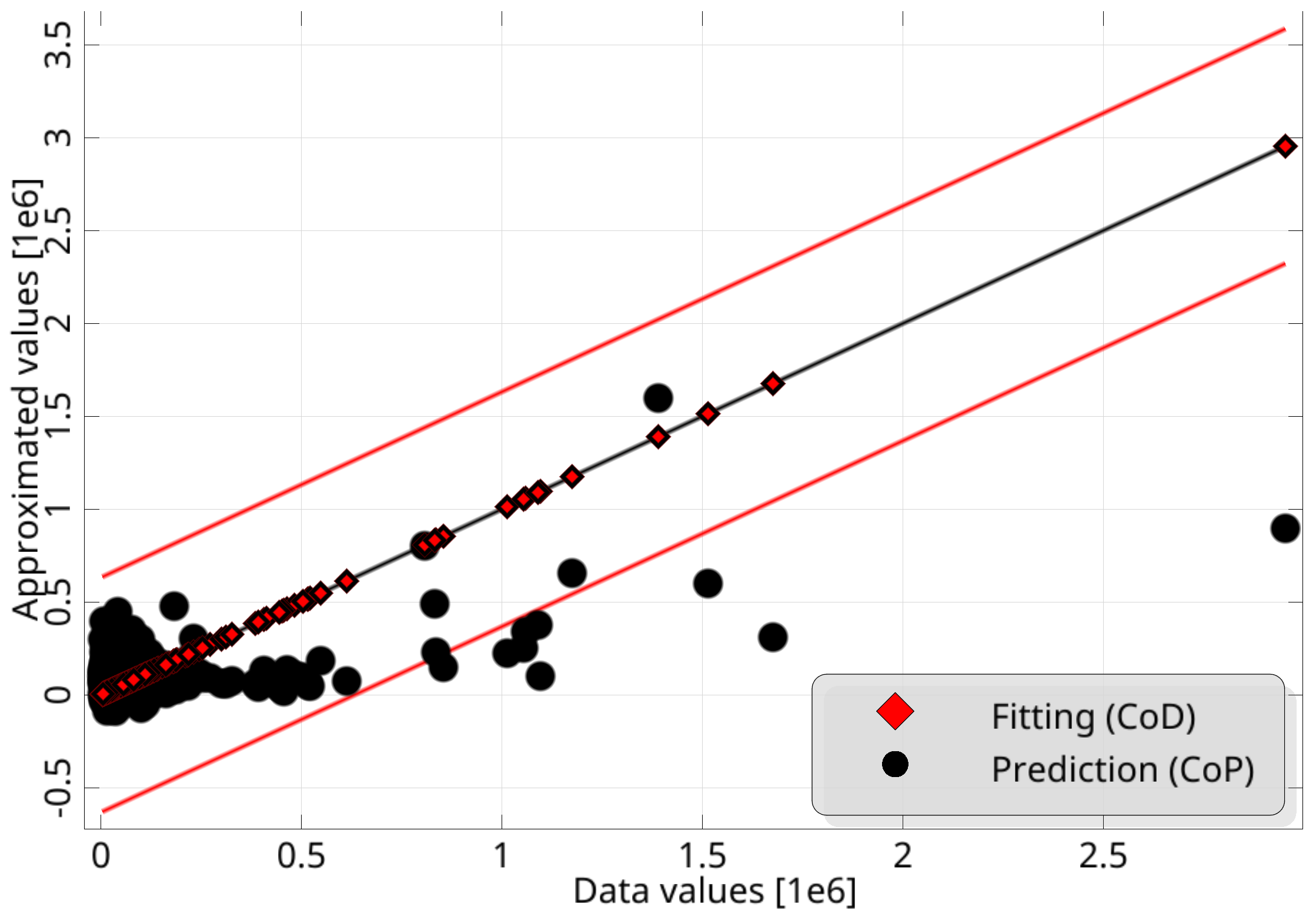}
	\hfill
	\includegraphics[width=0.49\textwidth]{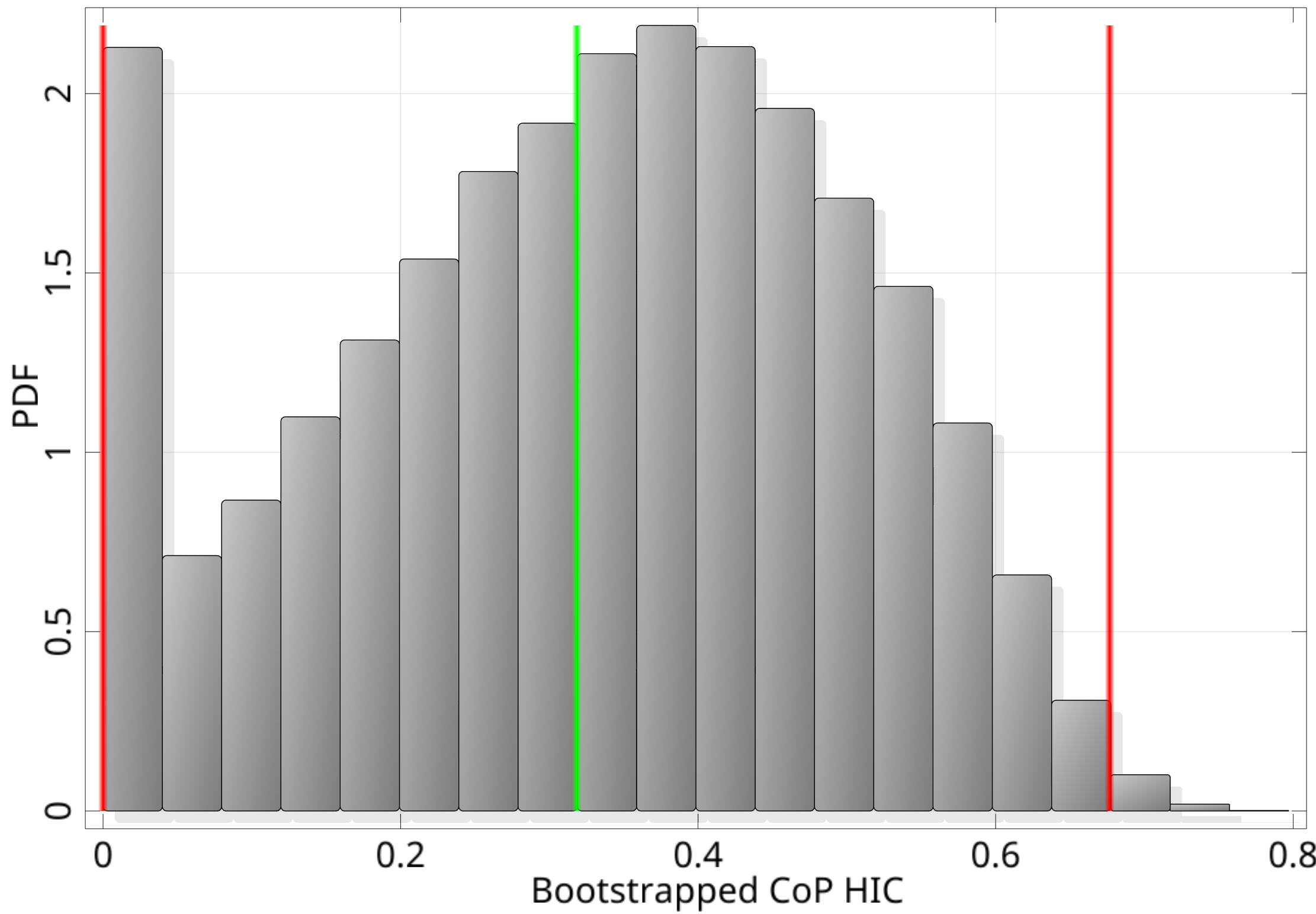}\\
 	\caption{Residual plots (left) and bootstrapped CoP's (right) of the HIC value from the front crash example}
	\label{crash_residuals2}
\end{figure}

\begin{table}[th]
\small
\centering
\begin{tabular}{ccccccc}
    \hline
Output & No. supports & Selected Model   & No. selected inputs& CoP & 99\% conf. interval & CoD test data \\
    \hline
             & 100 & Linear Polynomial   & 9 & 0.747 & 0.643 - 0.830 & 0.759 \\[-5pt]
N1\_disp     & 200 & Anisotropic Kriging & 14 & 0.803 & 0.735 - 0.856 & 0.809 \\[-5pt]
             & 400 & Anisotropic Kriging & 18 & 0.835 & 0.793 - 0.867 & 0.836 \\			 
    \hline
             & 100 & Linear Polynomial   & 9 & 0.778 & 0.676 - 0.856 & 0.787 \\[-5pt]
N2\_disp     & 200 & Anisotropic Kriging & 14 & 0.827 & 0.762 - 0.876 & 0.836 \\[-5pt]
             & 400 & Anisotropic Kriging & 15 & 0.857 & 0.823 - 0.885 & 0.853 \\			 
    \hline
             & 100 & Anisotropic Kriging & 11 & 0.990 & 0.986 - 0.993 & 0.989 \\[-5pt]
Stage1Pulse  & 200 & Anisotropic Kriging & 13 & 0.992 & 0.989 - 0.994 & 0.994 \\[-5pt]
             & 400 & Anisotropic Kriging & 13 & 0.994 & 0.992 - 0.995 & 0.994 \\			 
    \hline
             & 100 & Linear Polynomial   & 20 & 0.942 & 0.922 - 0.961 & 0.908 \\[-5pt]
Stage2Pulse  & 200 & Anisotropic Kriging & 18 & 0.956 & 0.946 - 0.965 & 0.932 \\[-5pt]
             & 400 & Anisotropic Kriging & 19 & 0.967 & 0.961 - 0.973 & 0.954 \\			 
    \hline
             & 100 & Linear Polynomial   & 9 & 1.000 & 1.000 - 1.000 & 1.000 \\[-5pt]
total\_mass  & 200 & Linear Polynomial   & 9 & 1.000 & 1.000 - 1.000 & 1.000 \\[-5pt]
             & 400 & Linear Polynomial   & 9 & 1.000 & 1.000 - 1.000 & 1.000 \\			 
    \hline
Head injury   & 100 & Anisotropic Kriging & 2 & 0.062 & 0.000 - 0.752 & 0.000 \\[-5pt]
criterion (HIC)& 200 & Anisotropic Kriging & 19 & 0.365 & 0.000 - 0.686 & 0.000 \\[-5pt]
             & 400 & Anisotropic Kriging & 21 & 0.318 & 0.000 - 0.677 & 0.000 \\			 
   \hline
 \end{tabular}
\caption{Computed quality estimates for the front crash example with 22 inputs and 6 investigated outputs by using k-fold cross validation and residual bootstrapping}
\label{crash_results}
\end{table}
In table \ref{crash_results} the results for the investigated six reponses are given. The table indicates, that with increasing number of support points, the prediction quality estimated with the CoP and verified with the test data set increases for almost all outputs. Furthermore, the estimated confidence interval of the CoP covers the verified test CoD very well. The table further indicates, that with increasing number of supports the number of selected important inputs increases, which is a typical phenomena in machine learning. In figure \ref{crash_cop_matrix} the estimated total effect sensitivity indices are shown for different support point sets. Already with 100 support points, the most important inputs could be detected for almost all outputs.
In contrast to the other outputs, only the Head injury criterion (HIC) could not be represented well with the investigated approximation models.
The CoP and its confidence interval indicate a very low prediction quality, which might be caused by numerical noise in the output or a high-dimensional non-linear relation between the inputs and the HIC value. In such a case, the estimate of the input sensitivity based on the approximation model as presented in section \ref{saltelli} is not reliable, since a high amount of the output variation can not be represented by the approximation model. In figure \ref{crash_residuals1} the residuals and the bootstrapped CoP's are shown exemplarily for the N1\_disp response. 
For this output a clear improvement of the prediction quality can be observed with increasing number of support points, which is  indicated by a narrower confidence interval. In the residual plots no significant outlier or systematic approximation errors could be recognized.
This is not the case for the HIC value residuals shown in figure \ref{crash_residuals2}. Here a clear systematic approximation error could be detected, which confirms the estimated poor approximation quality. The calculated confidence intervals cover almost the whole domain of possible CoP values.

\clearpage
\subsection{Cut-In scenario example}

\begin{figure}[th]
\vspace{-0.5cm}
\center
	\includegraphics[width=0.75\textwidth]{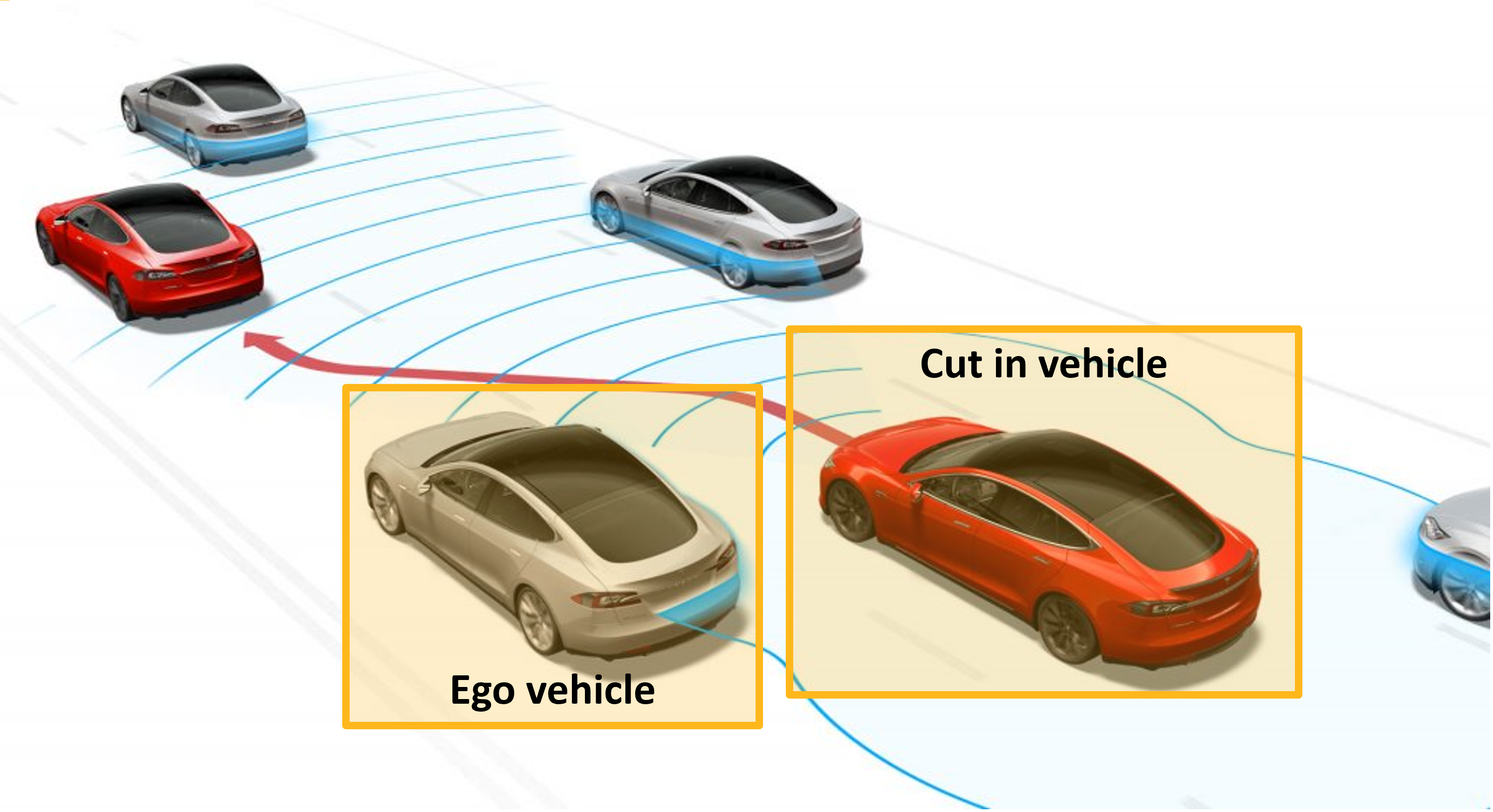}
 	\caption{Simulated Cut-Scenario of an autonomous vehicle}
	\label{cut_in}
\end{figure}

\vspace{-1.0cm}
\begin{table}[th]
\small
\centering
\begin{tabular}{ccccccc}
    \hline
Output & No. supports & Selected Model   & No. selected inputs& CoP & 99\% conf. interval & CoD test data \\
    \hline
Time		&280	&Anisotropic Kriging	&7	&0.666	&0.568 - 0.750	&0.745 \\[-5pt]
Headway	&560	&Anisotropic Kriging	&7	&0.722	&0.659 - 0.776 	&0.771 \\[-5pt]
(THW)		&1120	&Anisotropic Kriging	&7	&0.794	&0.756 - 0.830	&0.804 \\[-5pt]
				&1866	&Anisotropic Kriging	&8	&0.824	&0.795 - 0.850	&0.841 \\
    \hline
Time to		&280	&Anisotropic Kriging	&10	&0.417	&0.112 - 0.661 	&0.244 \\[-5pt]
collision	&560	&Anisotropic Kriging	&8	&0.459	&0.281 - 0.613	&0.490 \\[-5pt]
(TTC)			&1120	&Anisotropic Kriging	&10	&0.506	&0.385 - 0.613	&0.549 \\[-5pt]
					&1866	&Anisotropic Kriging	&9	&0.554	&0.464 - 0.634	&0.600 \\
    \hline
					&280	&Anisotropic Kriging	&5	&0.804	&0.735 - 0.863	&0.751 \\[-5pt]
Ego max 	&560	&Anisotropic Kriging	&8	&0.792	&0.721 - 0.845	&0.785 \\[-5pt]
speed			&1120	&Anisotropic Kriging	&9	&0.827	&0.791 - 0.858	&0.824 \\[-5pt]
					&1866	&Anisotropic Kriging	&8	&0.843	&0.812 - 0.869	&0.837 \\
    \hline
					&280	&Anisotropic Kriging	&7	&0.704	&0.583 - 0.800	&0.721 \\[-5pt]
Criticality	&560	&Anisotropic Kriging	&8	&0.713	&0.636 - 0.782	&0.758 \\[-5pt]
					&1120	&Anisotropic Kriging	&8	&0.773	&0.722 - 0.819	&0.797 \\[-5pt]
					&1866	&Anisotropic Kriging	&8	&0.808	&0.772 - 0.840	&0.830\\
    \hline
 \end{tabular}
\caption{Computed quality estimates for the Cut-In example with 10 inputs and 4 investigated outputs by using k-fold cross validation and residual bootstrapping}
\label{adas_results}
\end{table}

\vspace{-0.5cm}
In the current example, the simulation data of a Cut-In scenario of an autonomous vehicle are analyzed. Further details of the simulation analysis can be found in \cite{Most_2023_JAVS}. In this example 10 input parameters as ego and cut-in vehicle speeds, lead vehicle distance and breaking deceleration are considered. In the simulation the typical key performance indicators (KPIs) as critical time headway (THW), time to collision (TTC), collision speed and many others have been calculated. From these outputs a combined failure criterion was derived for each simulation run. 
For the analysis of the machine learning models, different data sets of 280, 560, 1120 and 1866 support points have been used for the training and 5600 data points are considered as verification data. Similar as in the previous example, different approximation models have been considered in the MOP competition and the most important inputs have been detected automatically. 

\begin{figure}[th]
\center
	\includegraphics[width=0.49\textwidth]{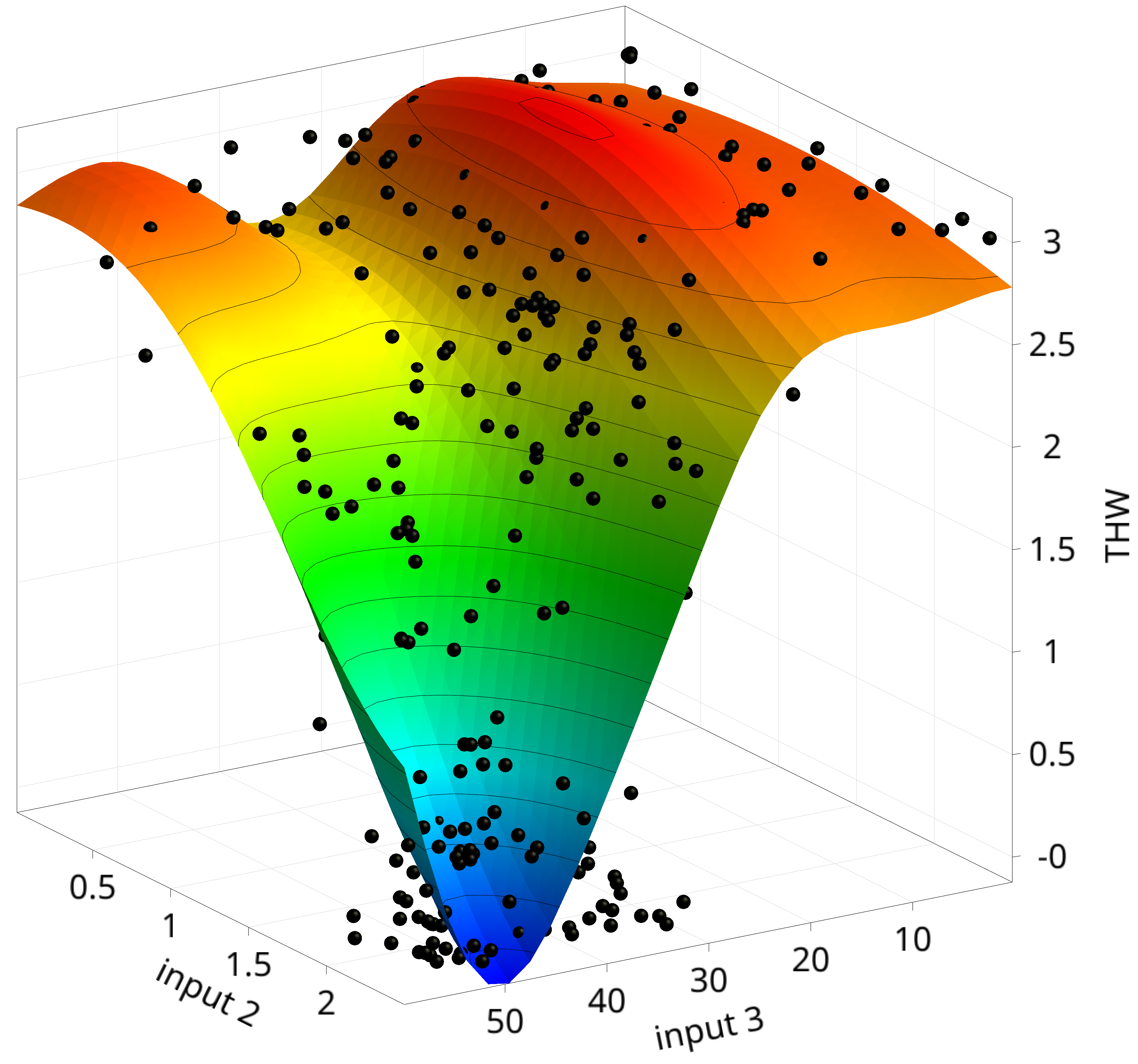}
	\hfill
	\includegraphics[width=0.49\textwidth]{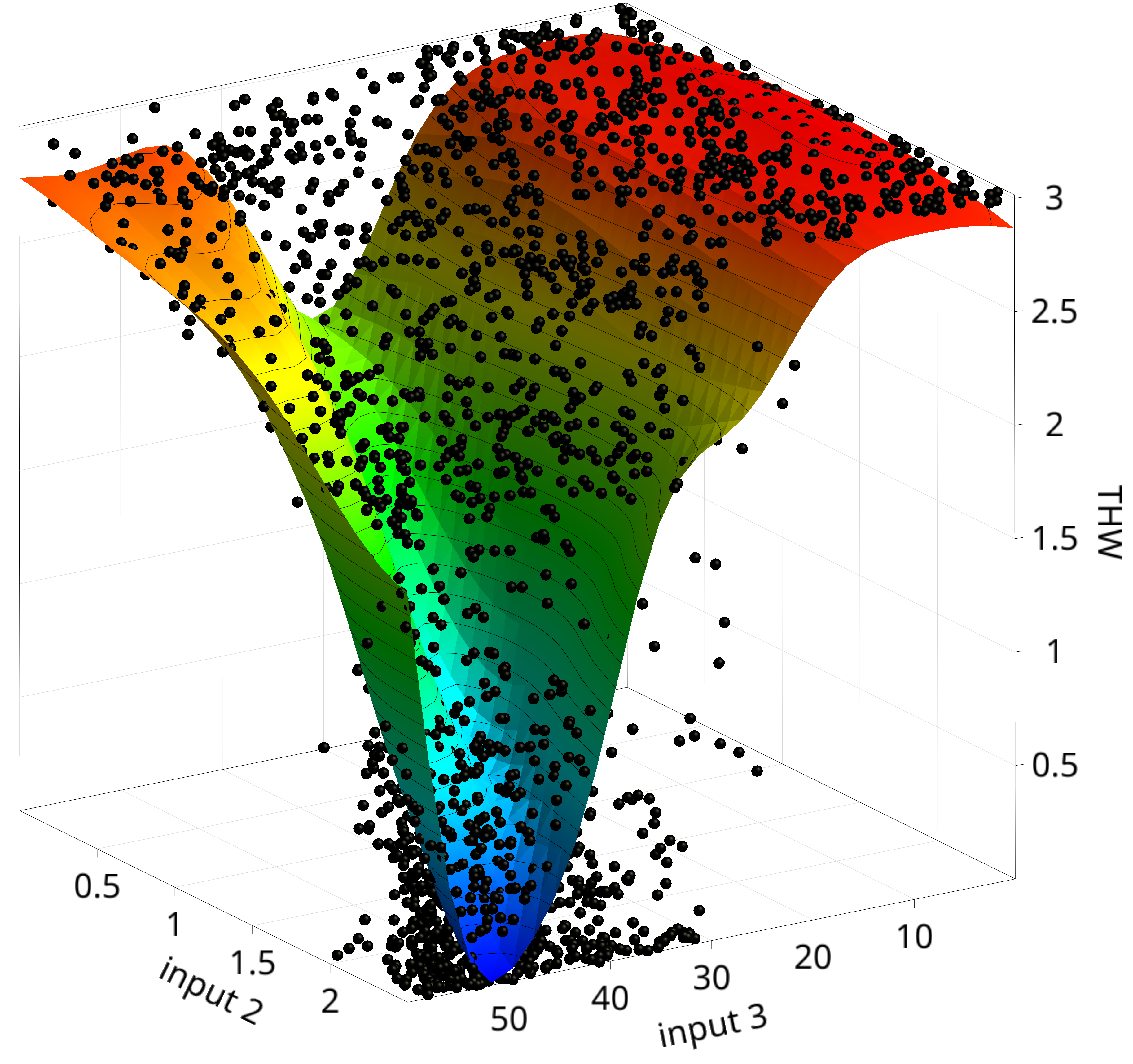}\\
 	\caption{Approximation model of the time headway (THW) for 280 training points (left) and 1866 training points (right) in the subspace of the two most important inputs of the Cut-In scenario example}
	\label{thw_plots}
\end{figure}

\begin{figure}[th]
\center
	280 support points, CoP = $66.6 \%$ \hspace{0.2\textwidth} 1866 support points, CoP = $82.4 \%$\\
  \includegraphics[width=0.49\textwidth]{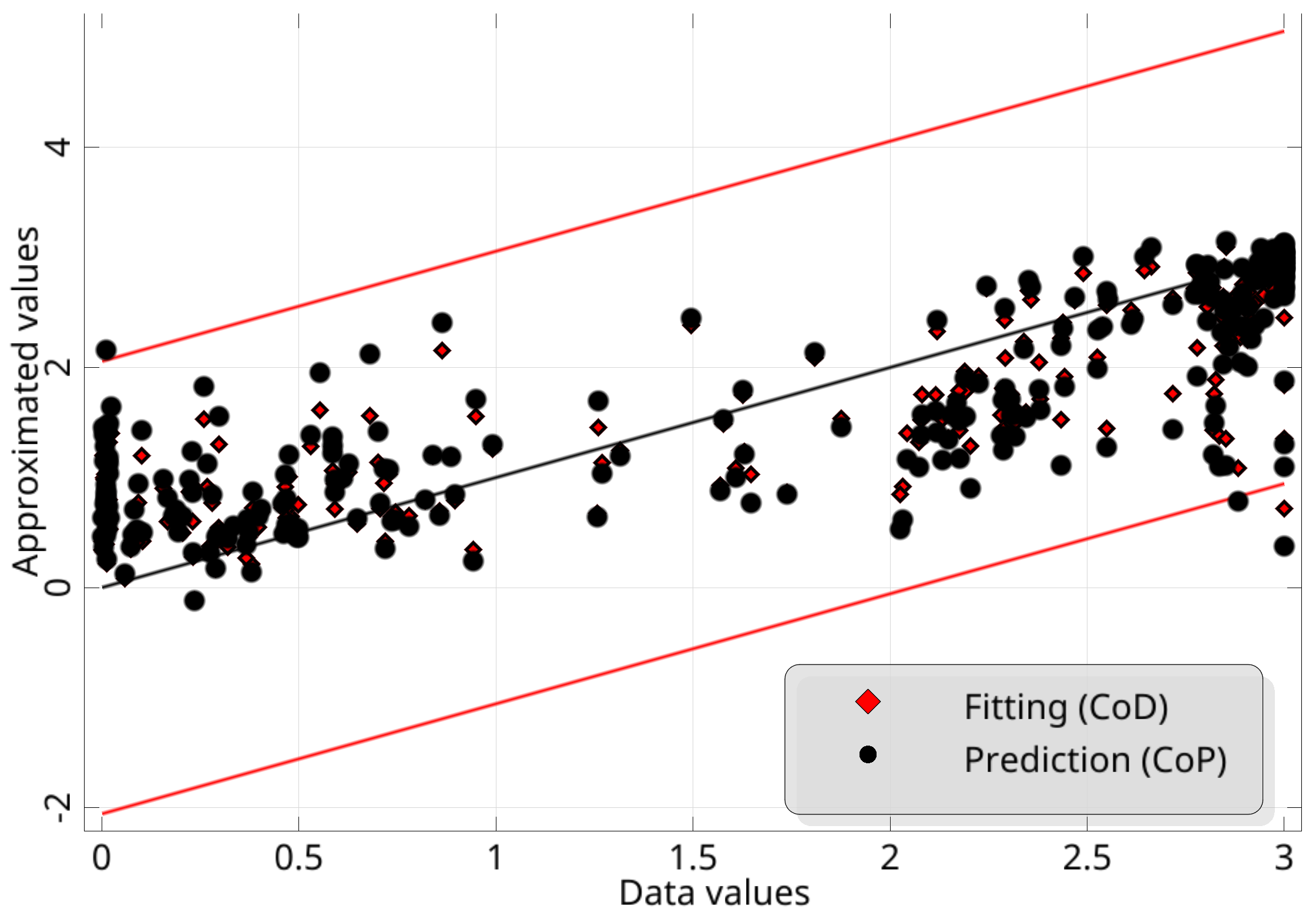}
	\hfill
	\includegraphics[width=0.49\textwidth]{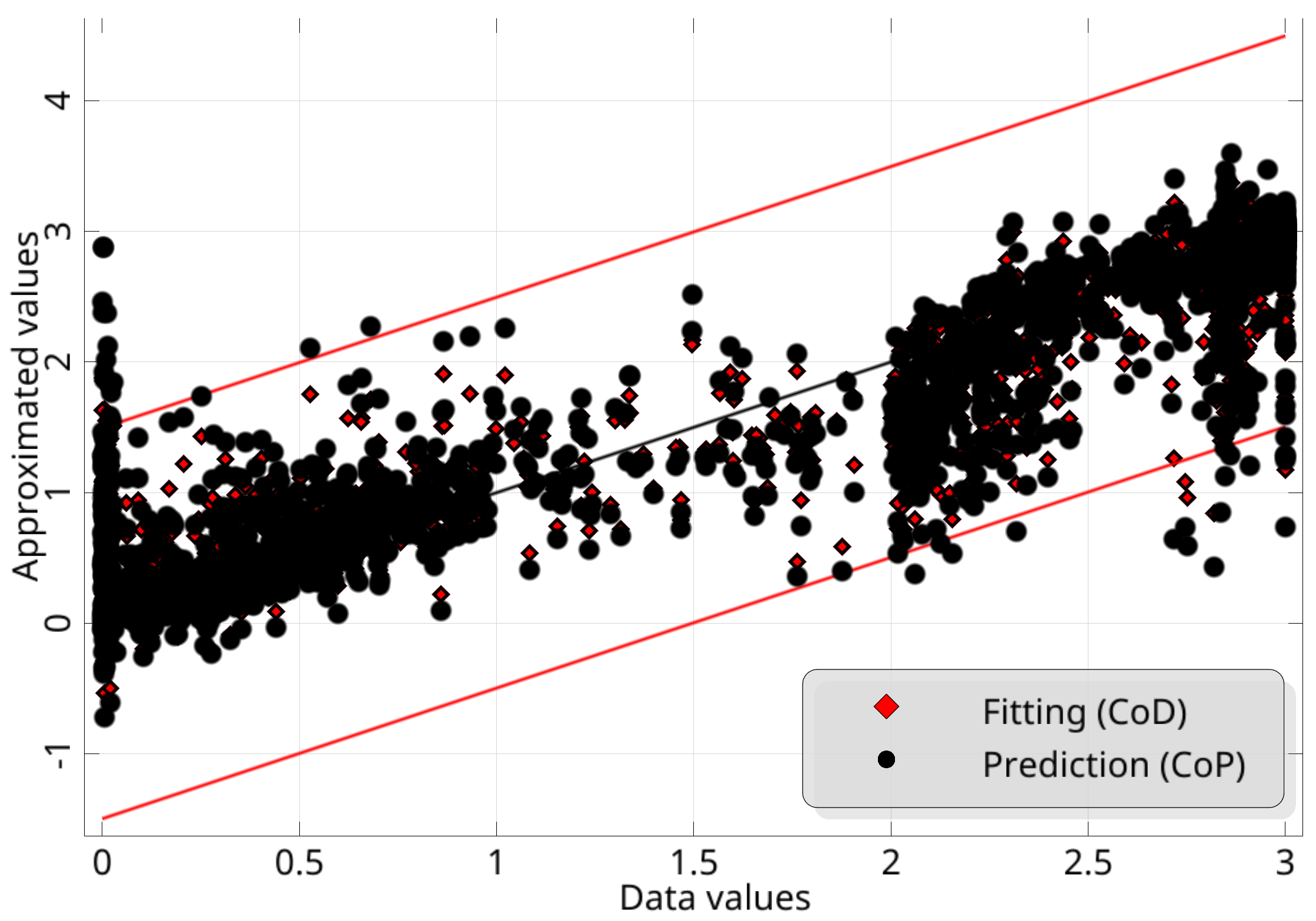}\\

	\includegraphics[width=0.49\textwidth]{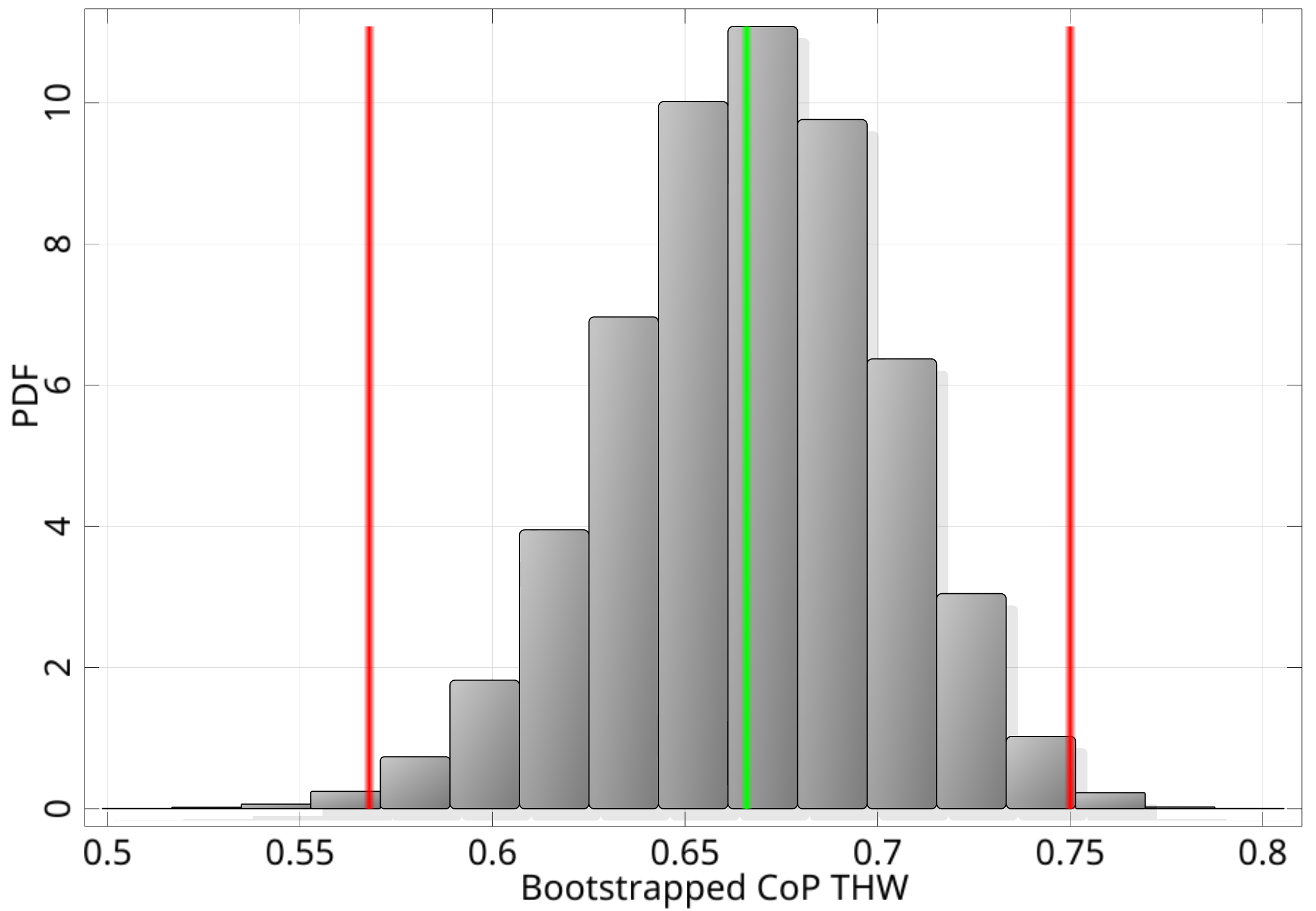}
	\hfill
	\includegraphics[width=0.49\textwidth]{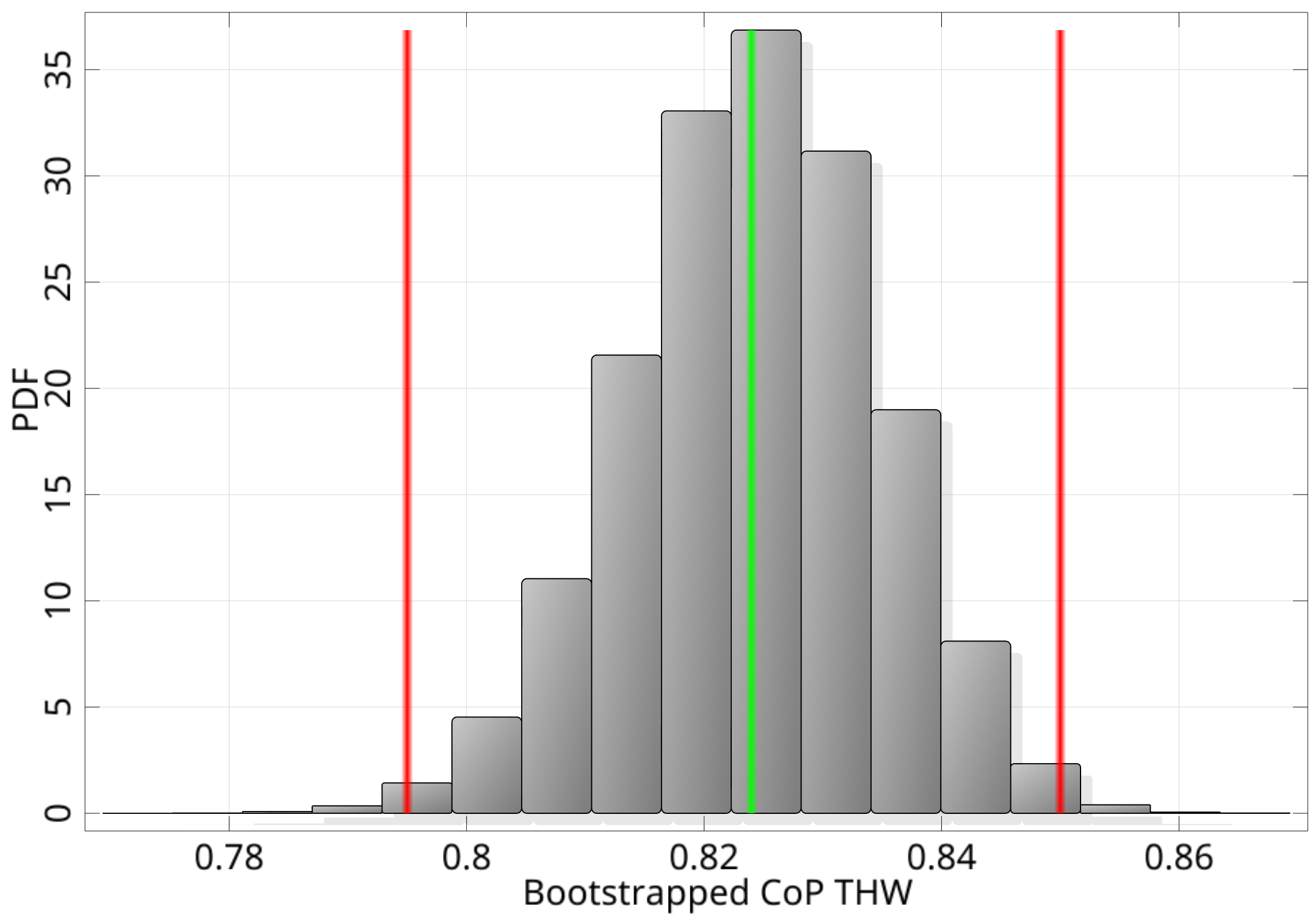}\\
 	\caption{Residual plots (top) and bootstrapped CoP's (bottom) of the time headway output of the Cut-In scenario example}
	\label{thw_residuals}
\end{figure}

In table~\ref{adas_results} the estimated CoPs for the training data and the CoDs of the verification data set are given for four selected outputs including the confidence interval from the bootstrapped residuals. 
The table shows, that for each investigated output, the estimated CoP increases with increasing number of training points. Furthermore, the CoD of the verification agrees very well with the CoP estimates and the corresponding confidence bounds.
In figure~\ref{thw_plots} the approximation model for the time-headway output is shown exemplarily for 280 and 1866 training points. In the first case the model already represents the global behaviour but local nonlinearities are filtered. For 1866 training points these local relations can be represented much more accurate. In figure~\ref{thw_residuals} the corresponding residual plots and the histograms of the bootstrapped CoP of both cases are shown. The figure indicates, that even with 1866 training points a perfect approximation of the simulated time headway is not possible. However, the estimated CoP was proven to be an accurate and reliable measure for the model prediction quality.

\clearpage
\subsection{Wedge splitting example with non-scalar outputs}
\begin{figure}[th]
\center
	\includegraphics[width=0.39\textwidth]{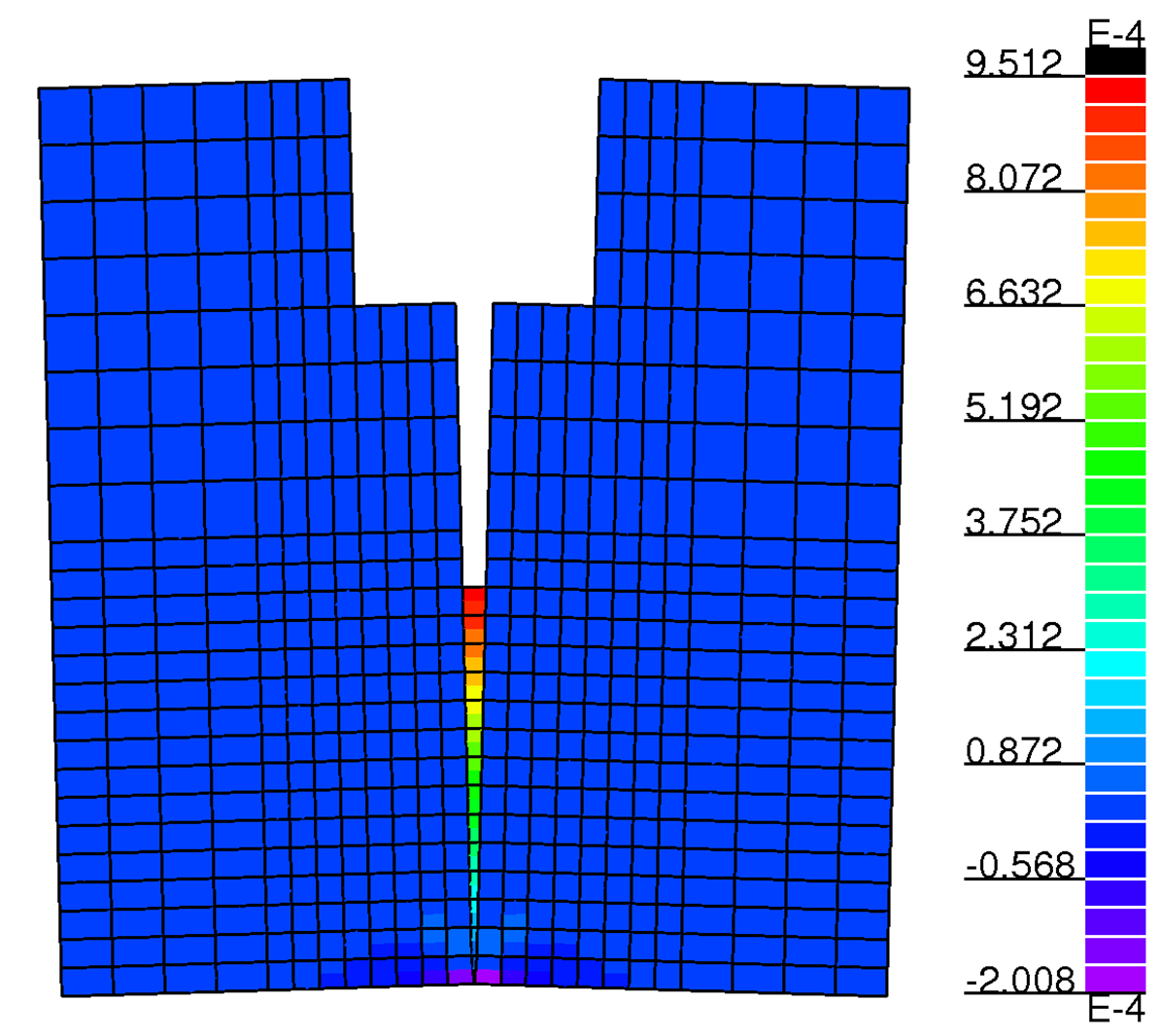}
	\hfill
	\includegraphics[width=0.59\textwidth]{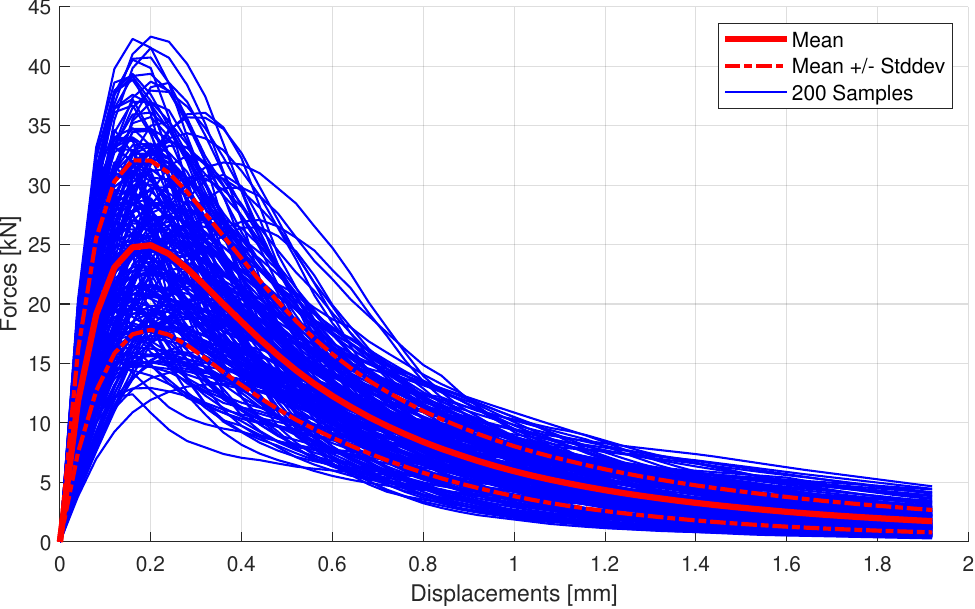}
 	\caption{Wedge splitting test example: finite element model (left) and simulated 200 Latin Hybercube samples of the load-displacement curve (right)}
	\label{wedge_splitting}
\end{figure}
In the final example, the CoP estimates are investigated for non-scalar outputs. 
We consider the load-displacement curve of a wedge splitting test as one-dimensional output.
The simulation model shown in figure~\ref{wedge_splitting} considers an elastic base material and a predefined crack with bi-linear softening law. The structure is discretized by 2D finite elements and the softening curve is obtained by a displacement-controlled simulation. The displacements are measured as the relative displacements between the load application points.
Further details on the simulation model can be found in \cite{Most_2005_PhD}.
Six material parameters are varied to generate the samples: the Young’s modulus, the Poisson’s ratio, the tensile strength, the Mode-I fracture energy and two shape parameters  of the bi-linear softening law.

\begin{figure}[th]
\center
	\includegraphics[width=0.49\textwidth]{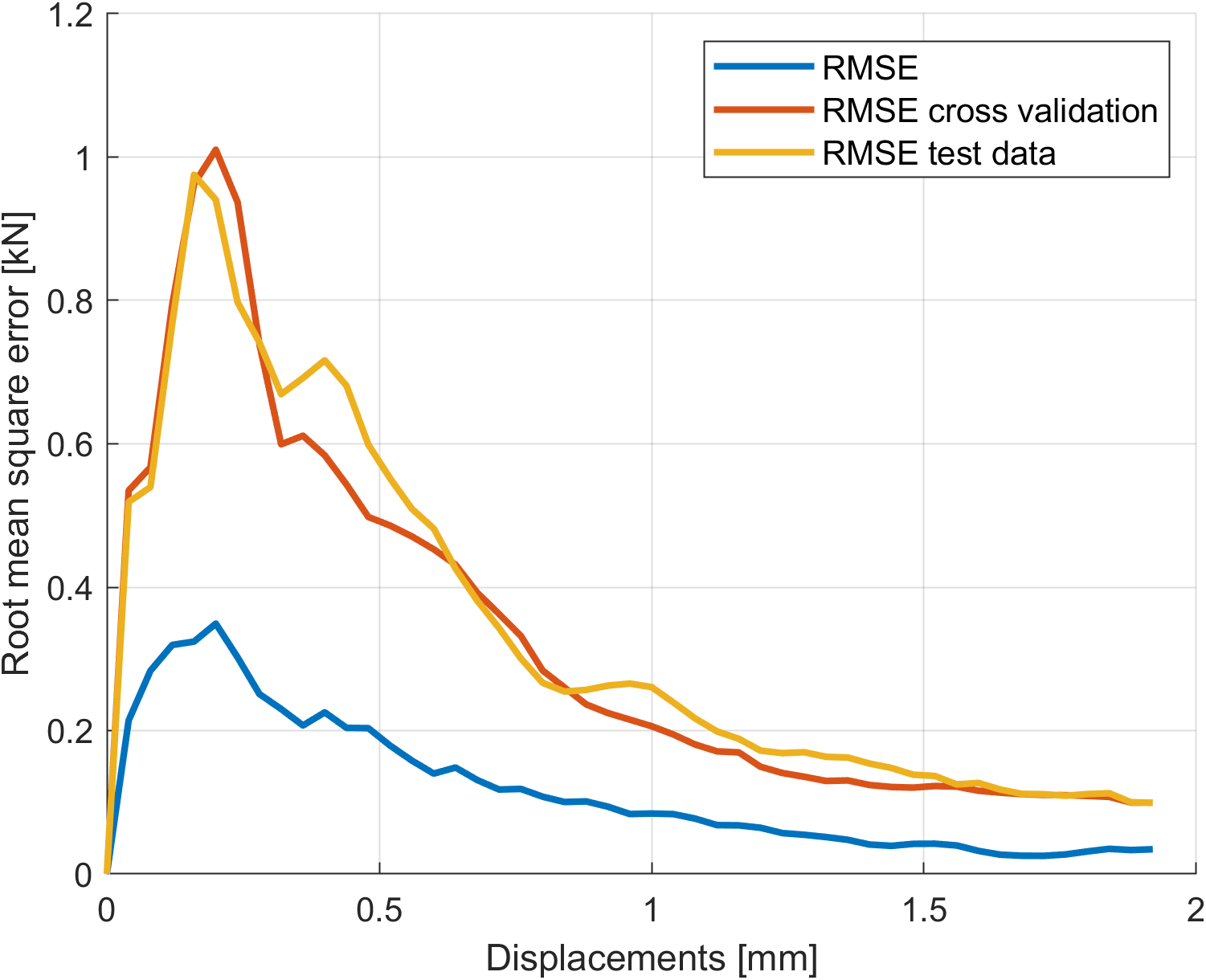}
	\hfill
	\includegraphics[width=0.49\textwidth]{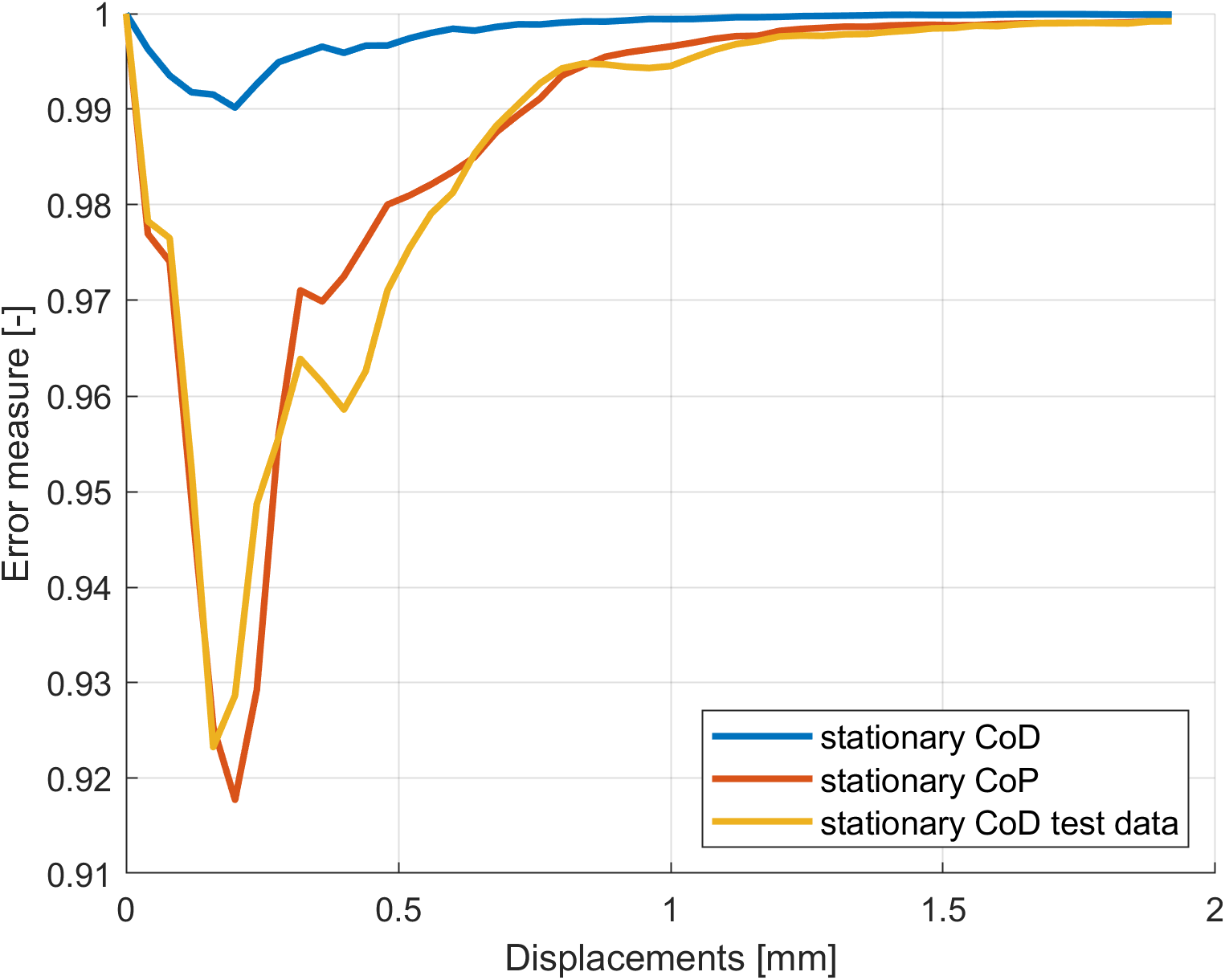}
 	\caption{Wedge splitting test example: RMSE of the model fit, from the cross-validation residuals and from the test samples (left) and corresponding stationary CoD and CoP values (right)}
	\label{wedge_results}
\end{figure}

The samples of the load-displacement curves are discretized at 49 equidistant displacement points. As approximation model we utilize the Deep Gaussian Covariance Network \cite{Cremanns_2021_Diss}, where an one-dimensional function is represented as a correlated Gaussian process model. A further application of this model for time-series approximation can be found in \cite{Most_2024_NAFEMS}. 
The estimated RMSE errors of the 200 training samples are shown for each displacement value in figure~\ref{wedge_results} together with the cross-validation results and the RMSE estimates from an independent test data set with 1000 Latin Hypercube samples. The figure indicates, that the cross-validation estimates agree very well with the test samples.
Since the standard deviation of the response decreases significantly with increasing displacements as shown in figure~\ref{wedge_splitting}, we consider the stationary CoD and CoP according to equation~\ref{CoD_stat} and \ref{CoP_stat}. In figure~\ref{wedge_results} both measures are shown together with the stationary CoD of the test data. As expected, all three quantities will approach to one for increasing displacements since the RMSE estimates decrease. If the ordinary CoD and CoP would be used instead of the stationary measures, the predicted approximation quality would 
decrease for larger displacements due to the reduced variation of the displacements in the samples. 

\section{Conclusions}
In this paper, statistical measures for the assessment of the prediction quality of machine learning models are investigated regarding their accuracy and robustness. Based on a cross-validation approach, the Coefficient of Prognosis was introduced as a model independent quality measure. However, the implementation of the cross-validation procedure is very important for a stable estimation of the prediction quality as shown in the numerical examples. From these findings, we would prefer the k-fold cross-validation towards the Leave-one-out approach since it gives more conservative estimates especially for a limited number of training data points.
Statistical confidence bounds of these global quality measures have been derived by using the bootstrap approach, whereas the resampling was evaluated directly on the cross-validation residuals. Therefore, this procedure can be applied without any additional model training. By means of several numerical examples, the value of the estimated confidence bounds could be demonstrated. This additional information helps to decide, how reliable the quality estimators are, if further data points are necessary, or if the prediction quality is affected by possible outliers.
Additionally, to the global quality measures, we introduced the local Root Mean Squared Error (RMSE) and the local CoP as local quality measures, which can be evaluated for each approximation point. They offer model independent error estimators of the local model prediction, which could be very valuable for Digital Twins applications.

The extension to non-scalar outputs requires the unique mapping of the discretization to a reference mesh, where the prediction error of each discretization point could be evaluated similarly to a scalar output. In order to obtain a unitless measure as the CoD and CoP, a normalization could be realized using the individual variation of each discretization point or assuming a stationary output variation. Both procedures require an efficient implementation as realized in Statistics on Structures \cite{Wolff_2016_RDO},\cite{sos2021} as part of the Ansys optiSLang software package \cite{optislang2023}.

\begin{acknowledgment}
This article is dedicated to Prof. Christian G. Bucher, former professor at the Bauhaus-University in Weimar, Germany, and Technical University in Vienna, Austria. Prof. Bucher supported the work of the former Dynardo GmbH over more than 20 years with his excellent expertise and knowledge.
Additionally, we want to thank Dr. Johannes Will, who is the founder of the former Dynardo GmbH, for his huge commitment to the success of the Metamodel of Optimal Prognosis approach within the Ansys optiSLang community.
\end{acknowledgment}




\end{document}